\documentclass[11pt]{article}
\usepackage{ifthen}
\usepackage{mdwlist}
\usepackage{amsmath,amssymb,amsfonts,amsthm}
\usepackage{bm}
\usepackage[colorlinks=true,linkcolor=blue,citecolor=blue,urlcolor=blue,hypertexnames=false]{hyperref}
\usepackage{enumitem}
\usepackage{graphicx}
\usepackage{xspace}
\usepackage{verbatim}
\usepackage{algorithm}
\usepackage{algpseudocode}
\usepackage[margin=1in]{geometry}
\usepackage{color}
\usepackage{thm-restate}
\usepackage{latexsym}
\usepackage{epsfig}
\usepackage{cleveref}
\usepackage[colorinlistoftodos,textsize=scriptsize]{todonotes}

\usepackage{lipsum}
\usepackage[numbers]{natbib}
\usepackage{mathrsfs}

\allowdisplaybreaks[1]

\newtheorem{corollary}{Corollary}

\newtheorem{lemma}{Lemma}[section]
\theoremstyle{definition}
\newtheorem{assumption}{Assumption}
\newcommand{\E}{\mathbb E}
\newcommand{\Prob}{\mathbb P}
\newcommand{\R}{\mathbb R}
\newcommand{\N}{\mathcal N}
\newcommand{\norm}[1]{\left\lVert#1\right\rVert}
\newcommand{\lp}[1]{\left\lVert#1\right\rVert_{L^p(\ell_\infty)}}
\newcommand{\op}{\mathrm{op}}
\newcommand{\tr}{\operatorname{tr}}
\newcommand{\elln}{\ell_n}
\newcommand{\cF}{\mathcal F}
\newcommand{\TV}{\mathrm{TV}}
\newcommand{\gap}{\mathrm{gap}}
\newcommand{\diag}{\operatorname{diag}}
\newcommand{\Cov}{\operatorname{Cov}}
\newcommand{\Var}{\operatorname{Var}}

\newcommand{\inorm}[1]{\left\|#1\right\|_\infty}

\numberwithin{theorem}{section} 
\numberwithin{assumption}{section} 
\numberwithin{lemma}{section} 
\numberwithin{proposition}{section} 
\numberwithin{corollary}{section} 
\theoremstyle{remark}
\newtheorem{remark}{Remark}[section]

\usepackage{tikz}
\usetikzlibrary{arrows.meta,calc,decorations.pathreplacing}

\title{Sharp Gaussian Approximation for Q-Learning}

\author {
Xingtu Liu\thanks{Department of Statistics and Data Science, University of California, Los Angeles. {\tt rltheory@outlook.com}.}
}

\begin{document}
\maketitle

\begin{abstract}
Gaussian approximation provides a foundation for online statistical inference in reinforcement learning. We establish sharp non-asymptotic Gaussian approximation rates in Wasserstein distance for asynchronous Q-learning with Polyak–Ruppert averaging. We prove an upper bound of $\widetilde O(n^{-1/4})$, where $n$ is the number of iterations, together with a matching lower bound up to logarithmic factors. Q-learning is more challenging to analyze than linear stochastic approximation due to the nonlinearity of the Bellman operator. Asynchronous updates introduce further challenges due to Markovian noise. Our analysis separates the errors from Bellman nonlinearity, Markovian noise, and the finite-sample covariance discrepancy. In particular, we show that Markovian noise induces an unavoidable bias even when the Bellman operator is linear. The trade-off between this bias and the covariance discrepancy determines the convergence rate. We further extend our analysis to the synchronous setting and establish a sharp rate of $\widetilde O(n^{-1/2})$.
\end{abstract}

\section{Introduction}\label{mb:introduction}

Q-learning \citep{watkins1992q} is one of the most fundamental algorithms in reinforcement learning (RL). It learns an optimal action-value function directly from sampled transitions and rewards. In this work, we study the convergence rate of Gaussian approximation for Q-learning as a function of the iteration number $n$, which is also the number of samples in the asynchronous setting. Considerable progress has been made in understanding the finite-sample guarantees for Q-learning \citep{even2003learning,beck2012error,li2020sample,li2024q}. However, this line of work only bounds the distance between
the estimate and the optimal Q-function, which does not determine the distribution of its error. Gaussian approximation addresses a different question. For the average $\bar Q_n=n^{-1}\sum_{k=1}^n Q_k$, we study how close the distribution of $\sqrt n(\bar Q_n-Q^\star)$ is to a centered Gaussian distribution. This provides a theoretical foundation for statistical inference, including the construction of confidence intervals and hypothesis tests \cite{shi2025statistical}. It also plays an important role in designing UCB- and LCB-based algorithms.

Stochastic approximation (SA) \citep{benveniste2012adaptive, borkar2008stochastic} is a general iterative framework to solve fixed-point equation problems. Since the Bellman operator in RL is a contraction map with a unique fixed point, many RL algorithms can be studied under this framework. A classic example is temporal-difference (TD) learning \citep{sutton1998reinforcement}, which is an instance of linear SA. In contrast, Q-learning involves maximizing over actions, making the underlying Bellman operator nonlinear. Q-learning is typically studied in two settings, synchronous and asynchronous, which correspond to SA with martingale difference noise and Markovian noise, respectively. In synchronous Q-learning, all state-action pairs are updated at each iteration using samples from a generative model. This requires the ability to sample transitions from arbitrary state-action pairs, which is typically only available in simulations. In contrast, asynchronous Q-learning updates only the state-action pair visited at each iteration, making it a realistic formulation for interaction with the real environment. However, this gives rise to Markovian noise. In this work, we focus mainly on the more challenging asynchronous setting and then extend our results to the synchronous setting.

Gaussian approximation for SA algorithms has been extensively studied \citep{mou2020linear,samsonov2025statistical,han2024decoupled,hu2024central,kong2025nonasymptotic,borkar2025ode,srikant2024rates,samsonov2024gaussian,li2026weak}. However, existing results may not directly apply to specific RL algorithms, which can present additional challenges. Moreover, existing SA results often rely on more restrictive assumptions and do not fully exploit problem-specific structure and parameters. The lower-bound constructions therein are also not applicable to RL algorithms. Non-asymptotic Gaussian approximations for TD learning have been established in \cite{srikant2024rates, samsonov2024gaussian}. For Q-learning, functional central limit theorems have been shown \citep{xie2022statistical,li2023statistical,zhang2024constant,panda2025asymptotic}. The first non-asymptotic Gaussian approximation bound for Q-learning was proved in \citep{pmlr-v331-liu26b}, which obtained a rate of $\widetilde{O}(n^{-1/6})$ in Wasserstein distance under the asynchronous setting. Subsequently, \citep{rubtsov2026gaussian} established a bound with the same rate on $n$ in a different distance. These results leave open the question of whether the convergence rate can be further improved and, more fundamentally, 

\begin{center}
    \textit{What is the sharpest achievable rate of Gaussian approximation for Q-learning in terms of $n$?}
\end{center}

\subsection{Overview of Main Results}

We study Q-learning with Polyak--Ruppert averaging and polynomial stepsizes $\alpha_t=c_0(t+b)^{-\beta}$. Our objective is to quantify how accurately a centered Gaussian approximates the normalized error $X_n=\sqrt n(\bar Q_n-Q^\star)$. We first consider asynchronous updates along a single Markovian trajectory. Theorem~\ref{mb:upper} gives the following bound in Wasserstein distance, up to logarithmic factors,
\[
 W_1(X_n,Z_\infty)\lesssim
 \underbrace{n^{-1/4}}_{\text{Gaussian approximation}}
 +\underbrace{n^{\beta-1}}_{\text{covariance discrepancy}}
 +\underbrace{n^{1/2-\beta}}_{\substack{\text{Bellman nonlinearity}\\\text{and Markovian noise}}},
\]
where $Z_\infty\sim\N(0,\Sigma_\infty)$, $\Sigma_\infty$ is the limiting covariance, and $\frac12<\beta<1$.

The three terms reflect different parts of the approximation. First, we approximate a weighted sum $M_n$ of fixed-point martingale noises by a Gaussian vector $Z_n\sim\N(0,C_n)$, where the deterministic covariance $C_n$ retains the finite horizon. Second, we compare $C_n$ with its limit $\Sigma_\infty$. Third, we analyze the errors from the nonlinear Bellman operator and the Markovian noises.

A key step in the analysis is to retain the exact averaging weights when constructing the intermediate Gaussian approximation. Replacing these weights by their limits and using a second-moment coupling gives an error bound of order $n^{(\beta-1)/2}$. We instead first approximate $M_n$ by $Z_n$ and then compare them through their covariance matrices. The latter error is proportional to $\|C_n-\Sigma_\infty\|_F=O(n^{\beta-1})$. Avoiding the additional square root allows a sharper balance with the $n^{1/2-\beta}$ term and improves the previously established polynomial rate from $n^{-1/6}$ to $n^{-1/4}$. We also carry out a numerical experiment (Figure \ref{fig:example}) to validate our theoretical results.

The stepsize creates a trade-off between the last two terms. Increasing $\beta$ makes the stepsizes smaller and reduces the accumulated nonlinear and Markovian errors of order $n^{1/2-\beta}$. However, it increases the covariance discrepancy of order $n^{\beta-1}$. Smaller updates give samples collected near the end less time to influence the average, increasing the difference between the finite-sample covariance and its limit. The two orders balance at $\beta=3/4$, giving the upper bound $\widetilde O(n^{-1/4})$.

To determine whether this rate can be improved, we construct a fixed two-state, one-action MDP and establish matching lower bounds for every fixed $\beta\in(1/2,1)$ in Theorem~\ref{lb:theorem}. On this example, the mean bias has order $\sqrt n\,\alpha_n$, and the covariance has the expansion $C_n=\Sigma_\infty+n^{\beta-1}G+o(n^{\beta-1})$ with $G\succ0$. These two results give a lower bound of order $\max\{n^{1/2-\beta},n^{\beta-1}\}$. Since there is only one action, the Bellman operator is linear. Thus, the mean-bias lower bound isolates the effect of Markovian sampling. Together, the upper and lower bounds establish the optimal polynomial rate $n^{-1/4}$ up to logarithmic factors.

We then study synchronous Q-learning, which uses independent transition samples for every state-action pair at each iteration. Under certain conditions and up to logarithmic factors,
\[
 W_1(X_n,Z_\infty^{\mathrm{syn}})
 \lesssim n^{-1/2}+n^{-\beta}+n^{\beta-1},
\]
for $\frac12\le\beta<1$. Independent sampling removes the Markovian mean bias. This allows $\beta=1/2$ and yields a better rate $\widetilde O(n^{-1/2})$. Switching the same lower-bound MDP to the synchronous setting gives an $\Omega(n^{-1/2})$ lower bound at $\beta=1/2$ and an $\Omega(n^{\beta-1})$ lower bound for every fixed $\beta>1/2$. Thus the synchronous rate is also sharp up to logarithmic factors. 

\begin{remark}
In asynchronous Q-learning, each iteration uses one sample, and therefore $n$ denotes both the number of iterations and the number of samples. In the synchronous setting, we use $n$ to count iterations, each of which requires one sample per state-action pair. The total sample size is therefore $N=n|\mathcal S||\mathcal A|$. Thus, faster convergence in iterations does not by itself imply better sample complexity. A comparison at the same sample budget must also account for the dimension dependence hidden in our bounds, which we leave for future work.
\end{remark}

\section{Preliminaries}\label{mb:preliminaries}

An infinite-horizon discounted Markov decision process (MDP) is denoted by $\mathcal{M}$, and is defined by the tuple $\langle \mathcal S, \mathcal A, P, r, \gamma \rangle$ where $\mathcal S$ is the set of states, $\mathcal A$ is the set of actions, $P: \mathcal S \times \mathcal A \rightarrow \Delta_{\mathcal S}$ is the transition probability function, $r(s,a)\in[0,1]$ is the reward, and $\gamma \in (0, 1)$ is the discount factor. Let $d=|\mathcal S||\mathcal A|$ and $H=(1-\gamma)^{-1}$. Let $\Delta_{\mathcal A}$ denotes the simplex over the action space. The action-value function (Q-function) of a stationary and stochastic policy $\pi: \mathcal S \rightarrow \Delta_{\mathcal A}$ is defined as $Q^{\pi}(s,a) = \mathbb{E} \Big[\sum_{t=0}^\infty \gamma^t r(s_t, a_t)|s_0=s,a_0=a\Big]$, where $a_t \sim \pi( \cdot | s_t)$ and $s_{t+1} \sim P( \cdot | s_t, a_t)$. The optimal Q-function is defined as $Q^{\star}:=\max_{\pi}Q^{\pi}$.  We identify a Q-function with a vector in $\mathbb R^d$.
For $q\in\mathbb R^d$, let $V(q)\in \mathbb R^{|\mathcal S|}$ be its associated value vector and let
$\mathcal T:\mathbb R^d\to\mathbb R^d$ be the Bellman operator:
\begin{equation}
 V(q)(s)=\max_{a\in\mathcal A}q(s,a),\qquad
 (\mathcal Tq)(s,a)
 =r(s,a)+\gamma\sum_{s'\in\mathcal S}P(s'\mid s,a)V(q)(s').
 \label{mb:bellman}
\end{equation}
The operator $\mathcal T$ is a $\gamma$-contraction in the maximum norm.
It therefore has a unique fixed point $Q^\star$.
The reward bound gives $Q^\star\in[0,H]^d$. We write
$V^\star=V(Q^\star)$ for the optimal value function.

Samples are collected using a fixed behavior policy
$\pi_{\mathrm b}$. Given a state $s_t$, an action is sampled
from $\pi_{\mathrm b}(\cdot\mid s_t)$, and a next-state
$s_{t+1}$ is sampled from $ P(\cdot\mid s_t,a_t)$. The transition triples
$Y_t=(s_t,a_t,s_{t+1})$ form a Markov chain with kernel $K$.
For $y=(s,a,s')$ and $\widetilde y=(\widetilde s,\widetilde a,
\widetilde s')$, this kernel is
\[
 K(y,\widetilde y)
 =\mathbf 1\{\widetilde s=s'\}
   \pi_{\mathrm b}(\widetilde a\mid\widetilde s)
   P(\widetilde s'\mid\widetilde s,\widetilde a).
\]
Let $\bar\mu$ be its stationary distribution and let $\mu$ be the
stationary distribution of the state-action pair. Thus
\[
 \mu(s,a)=\sum_{s'\in\mathcal S}\bar\mu(s,a,s'),\qquad
 \bar\mu(s,a,s')=\mu(s,a)P(s'\mid s,a).
\]

\begin{assumption}[Mixing and exploration]\label{ass:mix}
There exist constants $C_{\rm mix},c_{\rm mix}>0$ such that
\begin{equation}
 \sup_y\bigl\|K^j(y,\cdot)-\bar\mu\bigr\|_{\rm TV}
 \le C_{\rm mix}\exp(-c_{\rm mix}j),\qquad j\ge0.
 \label{eq:mix}
\end{equation}
Moreover, $\rho:=\min_{s,a}\mu(s,a)>0$.
\end{assumption}

Here $\|\nu-\nu'\|_{\rm TV}=\sup_E|\nu(E)-\nu'(E)|$, and the
supremum in \eqref{eq:mix} ranges over admissible triples
$(s,a,s')$ with $P(s'\mid s,a)>0$. We will use the following notations throughout the paper,
\[
 D=\operatorname{diag}\bigl(\mu(s,a):(s,a)\in\mathcal S\times\mathcal A\bigr),
 \qquad a_\star=\rho(1-\gamma)>0.
\]
Assumption~\ref{ass:mix} serves two
purposes. The mixing condition ensures the existence of bounded solutions to Poisson equations, supporting martingale approximations of the Markovian noise.
The condition $\rho>0$ ensures that every state-action
pair is visited with positive probability and that $D$ is nonsingular. Neither condition requires every transition probability $P(s'\mid s,a)$ to be positive. Assumption~\ref{ass:mix} is commonly used in the analysis of asynchronous Q-learning \citep{li2020sample, qu2020finite,chen2021lyapunov,zhang2022global,zhang2024constant}.

\paragraph{Notation.}
For vectors, $\|\cdot\|_\infty$ and $\|\cdot\|_2$ denote the maximum
and Euclidean norms. For matrices, $\|\cdot\|_\infty$ is the induced
maximum absolute row-sum norm, while $\|\cdot\|_F$ and
$\|\cdot\|_{\rm op}$ are the Frobenius and Euclidean operator norms.
For $p\ge1$, we write $
 \|Z\|_{L^p(\ell_\infty)}
 =\bigl(\mathbb E\|Z\|_\infty^p\bigr)^{1/p}.$
For random vectors with finite first moments, $W_1$ is the
Wasserstein distance between their laws with Euclidean cost;
equivalently,
\[
 W_1(X,Z)=\sup_{\operatorname{Lip}_2(h)\le1}
          \bigl|\mathbb E h(X)-\mathbb E h(Z)\bigr|,
\]
where $\operatorname{Lip}_2$ is the Lipschitz constant relative to
$\|\cdot\|_2$. For the sake of simplicity, we use $C$ to hide constants and some parameters. Throughout, $C$ may change from line to line and may depend on many factors, but not on $n,t,p$.

\section{Main Results}\label{mb:upper-section}

For each $(s,a)$, let $\mathbf u_{s,a}\in\mathbb R^d$ be the standard
basis vector with a $1$ in coordinate $(s,a)$ and zeros elsewhere.
For $y=(s,a,s')$, define the sampled update
\[
 U(q,y)=\mathbf u_{s,a}
 \bigl\{r(s,a)+\gamma V(q)(s')-q(s,a)\bigr\}.
\]
Starting from an initial $Q_0\in[0,H]^d$, we run
\begin{equation}
 Q_{t+1}=Q_t+\alpha_tU(Q_t,Y_t),\qquad
 \alpha_t=c_0(t+b)^{-\beta},\qquad \frac12<\beta<1,
 \label{eq:algorithm}
\end{equation}
where $c_0>0$, $\alpha_0\le1$, and the fixed shift $b\ge1$ satisfies $\frac{\beta}{c_0}b^{\beta-1}\le\frac{a_\star}{4}$.
Note that $b$ does not depend on $n$. The range $\beta\in(1/2,1)$ ensures $\sum_t\alpha_t=\infty$ and $\sum_t\alpha_t^2<\infty$. The polynomial stepsize and the restriction $\beta\in(1/2,1)$ are chosen to satisfy the standard convergence conditions for SA with Polyak–Ruppert averaging \cite{polyak1992acceleration,li2023statistical}. Polyak-Ruppert averaging is a classical technique for reducing variance and improving the statistical efficiency of SA. We study Q-learning with Polyak-Ruppert averaging and define the iterate error, averaged estimate, and normalized error as
\begin{align}
\Delta_t=Q_t-Q^\star,\qquad
 \bar Q_n=\frac1n\sum_{k=1}^nQ_k,\qquad
 X_n=\sqrt n\,(\bar Q_n-Q^\star). \label{eqn_main}
\end{align}
We use the filtration $\mathcal F_t=\sigma(Q_0,Y_0,Y_1,\ldots,Y_t)$.

\subsection{Assumptions} \label{sec_asm}

We discuss Assumption \ref{ass:quadratic} and Assumption \ref{ass:cov}, which are needed to establish our main results. Meanwhile, we introduce some key quantities, the nonlinear remainder $\varphi$, the martingale noise $\epsilon$, and the limiting covariance matrix $\Sigma_\infty$, that are important throughout our analysis.
We first discuss Assumption \ref{ass:quadratic}, which controls a remainder term $\varphi$ arising from the nonlinearity of the Bellman operator. Recall that $U$ denotes the update and $D$ accounts for the different update frequencies of the coordinates. For a fixed $q\in\mathbb R^d$, we define the stationary mean update as
\[
 B(q)=\mathbb E_{Y\sim\bar\mu}[U(q,Y)]
     =D(\mathcal Tq-q).
\]
Note that $B(Q^\star)=0$ and $\mathbb E[U(q,Y_t)\mid\mathcal F_{t-1}]
\ne B(q).$
Fix a deterministic optimal policy
$\pi^\star(s)\in\arg\max_{a\in\mathcal A}Q^\star(s,a)$, we define
\begin{equation}
 (P^{\pi^\star}q)(s,a)
 =\sum_{s'\in\mathcal S}P(s'\mid s,a)
   q(s',\pi^\star(s')),\qquad
 A=D(I-\gamma P^{\pi^\star}).
 \label{mb:drift}
\end{equation}
Since $Q^\star=r+\gamma P^{\pi^\star}Q^\star$, if the update rule $U$ does not take maximization over actions in each iteration, we have $B(Q^\star+\delta) =D\left[r+\gamma P^{\pi^\star}(Q^\star+\delta)-Q^\star-\delta\right]=-D(I-\gamma P^{\pi^\star})\delta=-A\delta$ for a perturbation $\delta\in\mathbb R^d$. In this case, the problem becomes linear SA. Now we use $\varphi(\delta)$ to denote the remainder term due to the nonlinearity of Q-learning. Since $B(Q^\star+\delta)=-A\delta+\varphi(\delta)$ for some $\varphi(\delta)$, we write
\[
\varphi(\delta):=B(Q^\star+\delta)+A\delta.
\]
The remainder has the explicit form
\[
 \varphi(\delta)(s,a)
 =\gamma\mu(s,a)\sum_{s'\in\mathcal S}P(s'\mid s,a)
 \bigl\{V(Q^\star+\delta)(s')-V^\star(s')
        -\delta(s',\pi^\star(s'))\bigr\}.
\]
Now we state our Assumption \ref{ass:quadratic}, which controls the nonsmoothness caused by maximizing over actions in the stationary mean update.
\begin{assumption}\label{ass:quadratic}
There is a finite constant $L$ such that$
 \|\varphi(\delta)\|_\infty\le L\|\delta\|_\infty^2$ for every $\delta$ with $Q^\star+\delta\in[0,H]^d.$
\end{assumption}

\begin{remark}
Similar assumptions have been made for studying nonlinear SA \cite{mokkadem2006convergence} and Q-learning \cite{khamaru2021instance,li2023statistical,pmlr-v331-liu26b,rubtsov2026gaussian} in previous works. For example, \cite{li2023statistical,pmlr-v331-liu26b} assume that $\| (P^{\pi_Q}-P^{\pi^{\star}})(Q-Q^{\star}) \|_{\infty} \leq L\| Q-Q^{\star} \|_{\infty}^2$, while \cite{rubtsov2026gaussian} assumes a strictly positive optimal action gap $\kappa=\min_{s\in\mathcal S}\min_{a\ne\pi^\star(s)}
 \bigl\{Q^\star(s,\pi^\star(s))-Q^\star(s,a)\bigr\}>0$. We show in the appendix that either condition is sufficient for Assumption \ref{ass:quadratic}.
\end{remark}
Next, we discuss the limiting covariance in our Gaussian approximation target. We first note that the noise in the update $U$ does not vanish at $Q^\star$. We denote this noise by
\begin{equation}
 \varepsilon_t=U(Q^\star,Y_t)=\gamma\mathbf u_{s_t,a_t}
 \left\{V^\star(s_{t+1})-
        \sum_{s'\in\mathcal S}P(s'\mid s_t,a_t)V^\star(s')\right\}.
 \label{eq:noise}
\end{equation}
Note that this is a martingale noise. To see this, let
$\mathcal G_t=\sigma(Q_0,s_0,a_0,\ldots,s_t,a_t)$ denote the
history before observing $s_{t+1}$. Conditional on
$\mathcal G_t$, the pair $(s_t,a_t)$ is known and the next state follows $P(\cdot\mid s_t,a_t)$. Thus, we have $\mathbb E[\varepsilon_t\mid\mathcal G_t]=0$. Since
$\mathcal F_{t-1}\subseteq\mathcal G_t$ for $t\ge1$, this implies $\mathbb E\varepsilon_0=0$ and $\mathbb E[\varepsilon_t\mid\mathcal F_{t-1}]=0$ under any initial state distribution. Thus the fixed-point noise is a martingale difference sequence even though the transitions $\{Y_t\}_{t\in[T]}$ are temporally dependent. Moreover, the centered update $U(Q_t,Y_t)-B(Q_t)$ need not be a martingale difference.

Now we denote $S=\mathbb E_{\bar\mu}[\varepsilon_0\varepsilon_0^\top]$. The martingale difference property gives
$\mathbb E[\varepsilon_t\varepsilon_u^\top]=0$ for $t\ne u$.
Consequently, under stationarity the covariance of a sum of these
increments has no additional lag-covariance terms. Moreover,
$\varepsilon_t$ has only one potentially nonzero coordinate. Therefore, $S$ is diagonal with $S_{(s,a),(s,a)} =\mu(s,a)\gamma^2 \operatorname{Var}_{s'\sim P(\cdot\mid s,a)}\bigl(V^\star(s')\bigr).$ The covariance of our Gaussian approximation target $\mathcal N(0,\Sigma_\infty)$ is
\[
 \Sigma_\infty=A^{-1}SA^{-\top},
\]
where $A^{-\top}=(A^{-1})^\top$. We make the following assumption.

\begin{assumption}\label{ass:cov}
The matrix $\Sigma_\infty$ is positive definite.
\end{assumption}

\begin{remark}
Although this assumption is not explicitly stated in \cite{pmlr-v331-liu26b,rubtsov2026gaussian}, it is implicitly required in their applications of the martingale CLT (see Lemma 10 of \cite{pmlr-v331-liu26b} and Lemma 8 of \cite{rubtsov2026gaussian}).
\end{remark}

\subsection{Upper Bound} \label{sec_ub}

We now state our main result.

\begin{restatable}{theorem}{maintheorem}
\label{mb:upper}
Under Assumptions~\ref{ass:mix}, \ref{ass:quadratic}, and \ref{ass:cov}, let $Z_\infty\sim\N(0,\Sigma_\infty)$.
For every $n\ge2$,
\begin{equation}
 W_1(X_n,Z_\infty)
 \le C\{n^{-1/4}+n^{\beta-1}+n^{1/2-\beta}\}.
 \label{mb:upper-bound}
\end{equation}
In particular, $\beta=3/4$ gives $Cn^{-1/4}$.
\end{restatable}

We next discuss the sources of error in the bound and present the proof sketch of the theorem. 
For $i\le j$, define $\Phi_{i:j}=\prod_{u=i}^j(I-\alpha_uA)$. We use the convention that $\Phi_{i:j}=I$ when $i>j$. We define
\begin{equation}
 W_{n,t}=\alpha_t\sum_{k=t+1}^n\Phi_{t+1:k-1},\qquad
 M_n=\frac1{\sqrt n}\sum_{t<n}W_{n,t}\varepsilon_t,\qquad
 C_n=\frac1n\sum_{t<n}W_{n,t}SW_{n,t}^\top.
 \label{eq:weights}
\end{equation}
We recall that $\Delta_t=Q_t-Q^\star$ and $X_n=\sqrt n\,(\bar Q_n-Q^\star).$ The first step of the proof is to decompose $X_n$ as 
$$X_n = M_n + R_n,$$ where $R_n$ is a remainder term that accounts for several sources of error. Denote $Z_n\sim\N(0,C_n)$. Then, we have
\begin{align}
W_1(X_n,Z_\infty) =W_1(M_n+R_n,Z_\infty) \le W_1(M_n+R_n,M_n)
+W_1(M_n,Z_n)+W_1(Z_n,Z_\infty). \label{eqn_dec}
\end{align}
We can see from \eqref{eqn_dec} that we apply Gaussian approximation to $M_n$, which is a weighted sum of martingale difference noises. The error due to the reduction from Markovian noise to martingale difference noise is captured by the remainder term \(R_n\).
We now discuss each term on the right-hand side of \eqref{eqn_dec} separately.

\begin{figure}[t]
    \centering
    \includegraphics[width=0.8\textwidth]{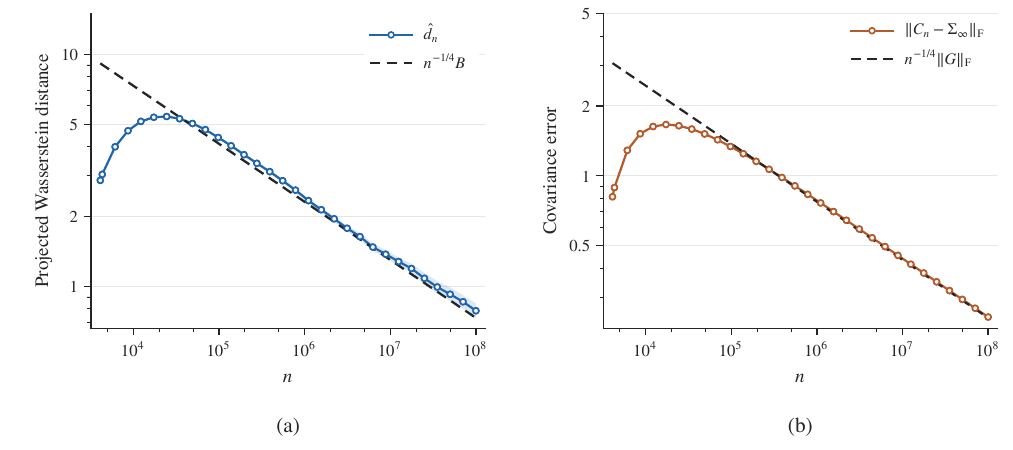}
    \caption{
Gaussian approximation for asynchronous Q-learning on a four-state, three-action MDP with $\beta=3/4$. (a) Empirical projected Wasserstein distance $\hat{d}_n$
and its theoretical rate $n^{-1/4}B$, where
$B\approx73.13670$ is computed from the given MDP and stepsize. (b) Covariance error $\|C_n-\Sigma_\infty\|_{\mathrm F}$
and its theoretical rate $n^{-1/4}\|G\|_{\mathrm F}$,
where $\|G\|_{\mathrm F}\approx24.44716$.
}
    \label{fig:example}
\end{figure}

\paragraph{The Error Remainder $R_n$}  For the first term, we have $W_1(M_n+R_n,M_n) \le \mathbb E\|(M_n+R_n)-M_n\|_2 =\mathbb E\|R_n\|_2$. Thus, we only need to upper bound the norm of $R_n$. We begin by observing that 
\begin{equation}
\Delta_{t+1}
=(I-\alpha_tA)\Delta_t
+\alpha_t\{\varepsilon_t+f_{\Delta_t}(Y_t)+\varphi(\Delta_t)\}
\label{eq:recursion}
\end{equation}
where $\varphi$ and $\varepsilon$ are discussed in Section \ref{sec_asm}, and
\begin{align*}
f_\delta(y) =U(Q^\star+\delta,y)-U(Q^\star,y)-\mathbb E_{Y\sim\bar\mu} \bigl[U(Q^\star+\delta,Y)-U(Q^\star,Y)\bigr].
\end{align*}
Therefore, $f_{\Delta_t}(Y_t)$ is the fluctuation of $U(Q_t,Y_t)-U(Q^\star,Y_t)$ around its stationary average $B(Q_t)$, which is a Markovian noise term. Iterating the recursion for $\Delta_t$, and using $X_n=n^{-1/2}\sum_{k=1}^n\Delta_k$ and $R_n = X_n - M_n$, we obtain 
\begin{align*}
R_n&=R_n^{\rm init}+R_n^{\rm nonlin}+R_n^{\rm Markov}
\end{align*}
where
\begin{align*}
R_n^{\rm init}=\frac1{\sqrt n}\sum_{k=1}^n\Phi_{0:k-1}\Delta_0, \quad
R_n^{\rm nonlin}=\frac1{\sqrt n}\sum_{t<n}W_{n,t}\varphi(\Delta_t), \quad
R_n^{\rm Markov}=\frac1{\sqrt n}\sum_{t<n}W_{n,t}f_{\Delta_t}(Y_t).
\end{align*}
 Thus, the term $R_n$ can be decomposed into three sources of errors, arising from initialization, the nonlinearity of the Bellman operator, and Markovian noise, respectively.  We next discuss each term separately.

\paragraph{The Initialization Error $R_n^{\rm init}$} The analysis of $\Phi_{i:j}$ and $W_{n,t}$ uses deterministic
estimates for polynomial stepsizes, building on standard techniques in stochastic approximation (see, e.g.,
\cite{polyak1992acceleration,li2023statistical,srikant2024rates,pmlr-v331-liu26b}).
The decay of the propagators bounds the initialization term $$\|R_n^{\rm init}\|_2 \le \frac{\sqrt d\,\|\Delta_0\|_\infty}{\sqrt n} \sum_{k=1}^n\|\Phi_{0:k-1}\|_\infty \le Cn^{-1/2}.$$

\paragraph{The Nonlinearity Error $R_n^{\rm nonlin}$} For the nonlinearity term, Assumption~\ref{ass:quadratic} gives
$\|\varphi(\delta)\|_\infty\le L\|\delta\|_\infty^2$.
The last-iterate moment bound proved in the appendix gives $\mathbb E\|\Delta_t\|_\infty^2\le C\alpha_t$. Combining these estimates with the uniform bound on $W_{n,t}$, we obtain
\[
\mathbb E\|R_n^{\rm nonlin}\|_2
\le
\frac C{\sqrt n}\sum_{t<n}
\mathbb E\|\Delta_t\|_\infty^2
\le
\frac C{\sqrt n}\sum_{t<n}\alpha_t
\le Cn^{1/2-\beta}.
\]

\paragraph{The Markovian Error $R_n^{\rm Markov}$} We now apply the Poisson equation technique \citep{glynn1996liapounov,makowski2002poisson,douc2018markov,chen2020explicit} to handle the Markovian term $R_n^{\rm Markov}$. Recall that $K$ denotes the transition kernel of the observation chain $Y_t$. The technique provides a solution to the Poisson equation $h_\delta-Kh_\delta=f_\delta$ with $\sup_y\inorm{h_\delta(y)} +\sup_y\inorm{Kh_\delta(y)} \le C\inorm\delta$. We denote $v_\delta:=Kh_\delta$. Adding and subtracting $v_{\Delta_t}(Y_{t-1})$ gives
\[
f_{\Delta_t}(Y_t)
=
\underbrace{
h_{\Delta_t}(Y_t)-v_{\Delta_t}(Y_{t-1})
}_{m_t}
+
v_{\Delta_t}(Y_{t-1})-v_{\Delta_t}(Y_t).
\]
Since $\Delta_t$ is $\mathcal F_{t-1}$-measurable, the Markov
property implies $\mathbb E[h_{\Delta_t}(Y_t)\mid\mathcal F_{t-1}] = (Kh_{\Delta_t})(Y_{t-1}) = v_{\Delta_t}(Y_{t-1}).$
Thus $m_t$ is a martingale difference, and the Poisson bounds give
$\inorm{m_t}\le C\inorm{\Delta_t}$.
Since the weights are uniformly bounded and $\mathbb E\|\Delta_t\|_2^2\le C\alpha_t$, we have
\[
\mathbb E\left\|
\frac1{\sqrt n}\sum_{t=1}^{n-1}W_{n,t}m_t
\right\|_2
\le
C\left(
\frac1n\sum_{t<n}\mathbb E\|\Delta_t\|_2^2
\right)^{1/2}
\le Cn^{-\beta/2}.
\]
For the remaining difference $T_n:= \frac1{\sqrt n}\sum_{t=1}^{n-1} W_{n,t}\{v_{\Delta_t}(Y_{t-1})-v_{\Delta_t}(Y_t)\}$, shifting the index in the first sum and separating the endpoints gives
\begin{align*}
T_n
={}
\frac1{\sqrt n}
\left\{
W_{n,1}v_{\Delta_1}(Y_0)
-
W_{n,n-1}v_{\Delta_{n-1}}(Y_{n-1})
\right\} +\frac1{\sqrt n}\sum_{t=1}^{n-2}
\left\{
W_{n,t+1}v_{\Delta_{t+1}}(Y_t)
-
W_{n,t}v_{\Delta_t}(Y_t)
\right\}.
\end{align*}
By the uniform boundedness of the weights and $v_{\Delta_t}$, the first part on the right-hand side can be bounded by $Cn^{-1/2}$. For the second part, adding and subtracting $W_{n,t+1}v_{\Delta_t}(Y_t)$ in each
interior term yields
\begin{align*}
&W_{n,t+1}v_{\Delta_{t+1}}(Y_t)
-
W_{n,t}v_{\Delta_t}(Y_t) =
(W_{n,t+1}-W_{n,t})v_{\Delta_t}(Y_t)
+
W_{n,t+1}
\{v_{\Delta_{t+1}}(Y_t)-v_{\Delta_t}(Y_t)\}.
\end{align*}
The first term accounts for changes in the weights, while the
second accounts for changes in the iterate. The total variation bound on the weights gives $C n^{-1/2}$ for the weight-variation sum.
For the other sum, Lipschitz continuity gives
\[
\|v_{\Delta_{t+1}}(Y_t)-v_{\Delta_t}(Y_t)\|_2
\le C\|\Delta_{t+1}-\Delta_t\|_2
\le C\alpha_t.
\]
Consequently,
\[
\|T_n\|_2
\le
\frac C{\sqrt n}
\left(1+1+\sum_{t<n}\alpha_t\right)
\le
C\{ n^{-1/2}+n^{1/2-\beta}\}.
\]
Combining all the above, we have
\[
\mathbb E\|R_n^{\rm Markov}\|_2
\le
C\{n^{-\beta/2}+n^{1/2-\beta}+ n^{-1/2}\}.
\]

\paragraph{Gaussian Approximation for $M_n$} We now turn to the remaining two terms on the right-hand side of \eqref{eqn_dec}. Recall that $M_n=\frac1{\sqrt n}\sum_{t<n}W_{n,t}\varepsilon_t$ is a weighted sum of martingale noises. A natural approach is to replace the time-varying weights $W_{n,t}$ by $A^{-1}$ and introduce $\widetilde M_n=\frac1{\sqrt n}\sum_{t<n}A^{-1}\varepsilon_t$. This allows us to apply existing Gaussian approximation results for sums of martingale differences (e.g., \cite{srikant2024rates}) by separately bounding $W_1(M_n,\widetilde M_n)$ and $W_1(\widetilde M_n,Z_\infty)$. However, this is why previous work \cite{pmlr-v331-liu26b,rubtsov2026gaussian} obtained a loose rate of $Cn^{-1/6}$. To obtain a sharper rate, we instead retain the exact weights $W_{n,t}$ and first approximate $M_n$ by the Gaussian vector $Z_n$ with covariance $C_n$. We then bound the error from replacing $C_n$ by its limit $\Sigma_\infty$. This leads us to bound $W_1(M_n,Z_n)$ and $W_1(Z_n,Z_\infty)$ separately.
We first explain why the previous approach leads to a slower convergence rate. By the coupling inequality for Wasserstein distance, we have $W_1(M_n,\widetilde M_n) \le \left(\mathbb E\|M_n-\widetilde M_n\|_2^2\right)^{1/2}.$
Since 
$$\mathbb E\|M_n-\widetilde M_n\|_2^2
\le \frac{C}{n}\sum_{t<n}\|W_{n,t}-A^{-1}\|_\infty
\le Cn^{\beta-1},$$
we obtain $W_1(M_n,\widetilde M_n)\le Cn^{(\beta-1)/2}$.
Balancing this term with another source of error of order $n^{1/2-\beta}$ requires choosing $\beta=2/3$, leading to the rate $Cn^{-1/6}$.
Our approach avoids this by retaining the exact weights $W_{n,t}$ and directly approximating $M_n$ by a centered Gaussian vector $Z_n \sim \mathcal N(0, C_n)$, where \(C_n\) approximates the limiting covariance \(\Sigma_\infty\). We first establish the Gaussian approximation for $M_n$, $W_1(M_n,Z_n)\le C n^{-1/4}.$ We then bound the distance between the two Gaussian distributions through their covariance matrices,
\[
W_1(Z_n,Z_\infty)
\le C\|C_n-\Sigma_\infty\|_{\mathrm F} \le \frac Cn\sum_{t<n}\|W_{n,t}-A^{-1}\|_\infty \le Cn^{\beta-1}.
\]
This bound does not have an additional square root on the power. Consequently, balancing $n^{\beta-1}$ with $n^{1/2-\beta}$ allows us to choose $\beta=3/4$, resulting in the improved rate $Cn^{-1/4}$.
The key insight from this proof technique is that bounding {the second moment of the difference between two random vectors} can yield a much looser estimate than directly bounding {the difference between their covariance matrices}.

\paragraph{Putting everything together} Combining all of the above, we obtain
\begin{align*}
W_1(X_n,Z_\infty) \le C\{n^{-\beta/2}+n^{1/2-\beta}+ n^{-1/2}
+ n^{-1/4}+n^{\beta-1}\}.
\end{align*}
Since $\beta>1/2$, the terms $n^{-\beta/2}$ and $n^{-1/2}$ are absorbed by $n^{-1/4}$, proving Theorem \ref{mb:upper}. 
The three terms left in the theorem have separate sources,
\[
\underbrace{n^{-1/4}}_{\text{Gaussian approximation of }M_n},\qquad
\underbrace{n^{\beta-1}}_{\text{finite-sample covariance}},\qquad
\underbrace{n^{1/2-\beta}}_{\text{Markovian noises}}.
\]
In the current analysis, the last term arises from both Markovian noises and Bellman nonlinearity. In the next section, we show that Markovian noise is the fundamental source of this error.

\subsection{Lower Bound}\label{lb:section}

In this section, we establish matching lower bounds over the full stepsize range $\beta\in(1/2,1)$.  We use an MDP with one action and two states, for which the Bellman operator is linear. The updates remain asynchronous and the noises are Markovian. This construction shows that Markovian sampling can contribute to the Gaussian approximation error even in the absence of Bellman nonlinearity. We construct the following hard instance,
\begin{equation}\label{lb:example}
 \gamma=\frac45,\qquad
 P=\begin{pmatrix}4/5&1/5\\3/10&7/10\end{pmatrix},\qquad
 r=\begin{pmatrix}0\\1\end{pmatrix},\qquad
 \mu=\begin{pmatrix}3/5\\2/5\end{pmatrix}.
\end{equation}
Since there is only one action, we have $P^{\pi^\star}=P$, $V^\star(i)=Q_i^\star$, $\mathbf u_i=\mathbf u_{i,a}$, and $\mu$ is the stationary state distribution.
Initialize $s_0\sim\mu$ and $Q_0=0$. Direct calculation gives
\begin{equation}\label{lb:matrices}
 Q^\star=\begin{pmatrix}4/3\\3\end{pmatrix},\qquad
 A=\frac1{125}\begin{pmatrix}27&-12\\-12&22\end{pmatrix},\qquad
 S=\diag(64/375,56/375).
\end{equation}
The chain is geometrically mixing, both states are visited, and
$A,S\succ0$. Thus all assumptions of the upper bound hold.
The same problem will be used for every choice of stepsize parameters.
Here $a_\star=2/25$, and the admissible shift satisfies $ b\ge1,\alpha_0\le1,\frac\beta{c_0}b^{\beta-1}\le\frac1{50}$.

\begin{restatable}{theorem}{lbthm}
\label{lb:theorem}
For the problem in \eqref{lb:example}, every fixed
$\beta\in(1/2,1)$, and $c_0>0$, there exist $c>0$ and $n_0<\infty$ such that, for $n\ge n_0$,
\begin{equation}\label{lb:lower}
 W_1(X_n,Z_\infty)\ge c
 \begin{cases}
 n^{1/2-\beta},&1/2<\beta\le3/4,\\
 n^{\beta-1},&3/4<\beta<1.
 \end{cases}
\end{equation}
The problem and initialization are independent of $\beta,c_0,b$.
The constants may depend on these parameters.
\end{restatable}

The lower-bound argument consists of two parts, which we analyze separately below. The first part comes from the mean bias.

\begin{restatable}{proposition}{llemmaa}
\label{lb:prop-bias}
On \eqref{lb:example}, with stationary state initialization,
\begin{equation}\label{lb:mean-iterate}
 \E\Delta_t=\alpha_t v+O(\alpha_t^2+(t+b)^{-1}),\qquad
 v=\begin{pmatrix}248/1125\\-34/375\end{pmatrix}\ne0.
\end{equation}
For any fixed bounded deterministic $Q_0$, this implies
\begin{equation}\label{lb:mean-average}
 \E X_n=\frac{c_0}{1-\beta}v\,n^{1/2-\beta}
                 +o(n^{1/2-\beta}).
\end{equation}
\end{restatable}

Recall that the Markovian remainder $R_n^{\mathrm{Markov}}
=\frac1{\sqrt n}\sum_{t<n}W_{n,t}f_{\Delta_t}(Y_t)$
contributes an $n^{1/2-\beta}$ term to the upper bound.
Although $\E_{\bar\mu}f_\delta(Y)=0$ for each fixed $\delta$,
the dependence between $\Delta_t$ and $Y_t$ can make
$\E f_{\Delta_t}(Y_t)$ nonzero.
Proposition~\ref{lb:prop-bias} shows that this dependence
produces a systematic mean bias on our fixed example.
There is no Bellman nonlinearity, $\E M_n=0$, and the
initialization error term is $O(n^{-1/2})$, so the leading bias
comes entirely from $R_n^{\mathrm{Markov}}$.
Equation~\eqref{lb:mean-iterate} identifies its order
$\alpha_t$ and fixed nonzero direction $v$.
These biases accumulate under averaging to give
\eqref{lb:mean-average}. Since $Z_\infty$ is centered,
\[
W_1(X_n,Z_\infty)\ge \|\E X_n\|_2
\gtrsim n^{1/2-\beta}.
\]
Thus, the $n^{1/2-\beta}$ contribution in the upper bound
can be unavoidable even when the Bellman operator is affine.

The second part comes from the finite-sample covariance approximation error, contributing the same order as in the upper bound. Previous work \cite{sheshukova2025gaussian} showed that the discrepancy between the finite-sample and limiting variances yields an \(\Omega(n^{\beta-1})\) lower bound in convex distance for Polyak–Ruppert averaged SGD on a scalar quadratic problem. \cite{butyrin2025improved} used the same example for linear stochastic approximation. On our two-state MDP, it identifies the leading covariance correction for the weighted martingale term.  This correction has the same Lyapunov form as in \cite{sheshukova2025gaussian, butyrin2025improved}. Let $\mathcal X=\int_0^\infty e^{-As}S e^{-A^\top s}\,ds$ denote the unique positive-definite solution of the Lyapunov equation $A\mathcal X+\mathcal X A^\top=S$ \cite{khalil2002nonlinear}.

\begin{restatable}{proposition}{llemmab}
\label{lb:prop-cov}
For the weights in \eqref{eq:weights} on \eqref{lb:example},
\begin{equation}\label{lb:cov-expansion}
 C_n=\Sigma_\infty+n^{\beta-1}G+o(n^{\beta-1}),\qquad
 G=\frac1{c_0}A^{-1}\mathcal X A^{-\top}\succ0.
\end{equation}
\end{restatable}

This expansion shows that the covariance error is genuinely of
order $n^{\beta-1}$. Recall that $Z_n\sim N(0,C_n)$. Since $G\succ0$, every diagonal entry
satisfies $G_{jj}>0$. Projection onto coordinate $j$ and using the formula $W_1\!\left(N(0,\sigma^2),N(0,\tau^2)\right) =\sqrt{\frac2\pi}\,|\sigma-\tau|$ give
\[
W_1(Z_n,Z_\infty)
\ge \sqrt{\frac2\pi}
\left|\sqrt{(C_n)_{jj}}-\sqrt{(\Sigma_\infty)_{jj}}\right|
\gtrsim n^{\beta-1}.
\]
Thus, the covariance comparison used in the upper bound has
the correct order on this example.
For $\beta>3/4$, the finite-covariance approximation established
in the upper-bound proof gives
$W_1(X_n,Z_n)=o(n^{\beta-1})$.
The reverse triangle inequality then yields
\[
W_1(X_n,Z_\infty)
\ge W_1(Z_n,Z_\infty)-W_1(X_n,Z_n)
\gtrsim n^{\beta-1}.
\]
Combining both the upper and lower bounds, and letting $\beta=3/4$, we have the following corollary.

\begin{corollary}\label{lb:corollary}
When $\beta=3/4$ and $n\geq n_0$, we have
\[
 c n^{-1/4}\le W_1(X_n,Z_\infty)\le C n^{-1/4}\log n.
\]
\end{corollary}
The stepsize creates a trade-off between the accumulated Markovian mean bias and the finite-sample covariance discrepancy. Smaller stepsizes reduce the former, of order $\sqrt n\,\alpha_n$, but increase the latter, of order $(n\alpha_n)^{-1}$. This happens because smaller stepsizes make the updates slower, so samples collected near the end have less time to influence the average. This increases the discrepancy between the finite-sample covariance and its limit. Two orders $n^{1/2-\beta}$ and $n^{\beta-1}$ balance at $\beta=3/4$, which gives the optimal polynomial Gaussian approximation rate $n^{-1/4}$.

\begin{remark}
Our lower bound also applies to Polyak–Ruppert averaged tabular TD(0), since asynchronous Q-learning coincides with tabular TD(0) on a one-action MDP. Specifically, it yields the same Wasserstein lower bound for TD(0) with arithmetic averaging and polynomial stepsizes \(\alpha_t=c_0(t+b)^{-\beta}\), where \(\beta\in(1/2,1)\). On our fixed two-state instance, no fixed choice of \(\beta\) in this range gives a Gaussian approximation rate faster than \(n^{-1/4}\). This complements existing \(O(n^{-1/4})\) upper bounds for Markovian linear stochastic approximation in Kolmogorov distance \cite{samsonov2025statistical}.
\end{remark}

\section{Synchronous Q-Learning} \label{sec_syn}

We now study Gaussian approximation for synchronous Q-learning. In this setting, every state-action pair is updated at each iteration using independent samples. Unlike in the asynchronous setting, here $n$ denotes only the number of iterations, not the sample size. We retain the MDP and the notation $\Delta_t$, $\bar Q_n$, and $X_n$ in \eqref{eqn_main}, now defined using the synchronous iterates. We use the superscript $\mathrm{syn}$ to distinguish the drift and covariance matrices from their asynchronous counterparts.

At each iteration $t$, independently sample $S'_t(s,a)\sim P(\cdot\mid s,a)$ for every $(s,a)$. The sample blocks $\omega_t=(S'_t(s,a))_{s,a}$ are independent across iterations, and the samples within each block are independent. Starting from a deterministic $Q_0\in[0,H]^d$, we run
\begin{equation}\label{syn:algorithm}
 Q_{t+1}(s,a)=(1-\alpha_t)Q_t(s,a)
 +\alpha_t\{r(s,a)+\gamma V(Q_t)(S'_t(s,a))\},
 \qquad \alpha_t=c_0(t+b)^{-\beta}.
\end{equation}
As before, $c_0>0$ and $b\ge1$ are fixed independently of $n$. Write $a_{\mathrm{syn}}=1-\gamma$ and choose parameters to satisfy $ \frac12\le\beta<1, \alpha_0\le1,
 \frac\beta{c_0}b^{\beta-1}\le\frac{a_{\mathrm{syn}}}{4}.$
Unlike the asynchronous setting, the endpoint $\beta=1/2$ is included. The iterates remain in $[0,H]^d$. Fix a deterministic optimal policy $\pi^\star$ as in \eqref{mb:drift}, and define $ (\widehat P_t^{\pi^\star}q)(s,a)
 =q(S'_t(s,a),\pi^\star(S'_t(s,a))).$
The synchronous drift, fixed-point noise, and covariance matrices are
\begin{align}
 A^{\mathrm{syn}}&=I-\gamma P^{\pi^\star},
 &\varepsilon_t^{\mathrm{syn}}
 &=\gamma(\widehat P_t^{\pi^\star}-P^{\pi^\star})Q^\star,
 \notag\\
 S^{\mathrm{syn}}&=\E[\varepsilon_t^{\mathrm{syn}}
                      (\varepsilon_t^{\mathrm{syn}})^\top],
 &\Sigma_\infty^{\mathrm{syn}}
 &=(A^{\mathrm{syn}})^{-1}S^{\mathrm{syn}}(A^{\mathrm{syn}})^{-\top}.
 \label{syn:matrices}
\end{align}
Since all coordinates are updated at every iteration, the visitation matrix $D$ does not appear in $A^{\mathrm{syn}}$. Row independence gives
\[
 S^{\mathrm{syn}}=\diag\!\left(
 \gamma^2\Var_{s'\sim P(\cdot\mid s,a)}(V^\star(s')):
 (s,a)\in\mathcal S\times\mathcal A\right).
\]
We impose the following action-gap and covariance conditions.
\begin{assumption}\label{syn:assumption}
There is one action, or every state has a unique optimal action with
\[
 \kappa=\min_{s\in\mathcal S}\min_{a\ne\pi^\star(s)}
 \{Q^\star(s,\pi^\star(s))-Q^\star(s,a)\}>0.
\]
Moreover, $\Sigma_\infty^{\mathrm{syn}}\succ0$.
\end{assumption}

We will discuss this assumption in the appendix. This assumption allows us to express each sampled Bellman update as an exact linear recursion near \(Q^\star\), a key step in the proof for synchronous Q-learning. Assumption~\ref{ass:quadratic} bounds the remainder in the stationary mean update and applying it would give an error of order $n^{1/2-\beta}$, which does not vanish at $\beta=1/2$.
The covariance condition is the synchronous counterpart of Assumption~\ref{ass:cov} and is equivalent to requiring every diagonal entry of $S^{\mathrm{syn}}$ to be positive. No mixing, exploration, or stationary-initialization assumption is needed.

\subsection{Upper Bound}\label{syn:upper-section}

\begin{restatable}{theorem}{synupperthm}\label{syn:upper}
Under Assumption~\ref{syn:assumption}, let
$Z_\infty^{\mathrm{syn}}\sim\N(0,\Sigma_\infty^{\mathrm{syn}})$.
For every $n\ge2$,
\begin{equation}\label{syn:upper-bound}
 W_1(X_n,Z_\infty^{\mathrm{syn}})
 \le C\{n^{-1/2}\log(en)+n^{-\beta}+n^{\beta-1}\}.
\end{equation}
In particular, $\beta=1/2$ gives $Cn^{-1/2}\log(en)$.
For every fixed $\beta\in(1/2,1)$, the bound is $O(n^{\beta-1})$.
\end{restatable}

We next describe the sources of error in this bound. To handle the Bellman nonlinearity, we compare $Q_t$ with a fixed-policy recursion driven by the same samples:
\begin{equation}\label{syn:linear-recursion}
 L_0=Q_0,\qquad
 L_{t+1}=(1-\alpha_t)L_t
       +\alpha_t(r+\gamma\widehat P_t^{\pi^\star}L_t),\qquad
 X_n^{\mathrm{lin}}=\frac1{\sqrt n}\sum_{k=1}^n(L_k-Q^\star).
\end{equation}
The sampled Bellman map at $Q_t$ agrees with its fixed-policy counterpart whenever $\|Q_t-Q^\star\|_\infty<\kappa/2$.
For the covariance comparison, define the synchronous versions of the weights and covariance in \eqref{eq:weights}:
\begin{equation}\label{syn:weights}
\begin{aligned}
 \Phi^{\mathrm{syn}}_{i:j}=\prod_{u=i}^j(I-\alpha_uA^{\mathrm{syn}}),\qquad
 W^{\mathrm{syn}}_{n,t}=\alpha_t\sum_{k=t+1}^n\Phi^{\mathrm{syn}}_{t+1:k-1},\qquad
 C_n^{\mathrm{syn}}=\frac1n\sum_{t<n}W^{\mathrm{syn}}_{n,t}
 S^{\mathrm{syn}}(W^{\mathrm{syn}}_{n,t})^\top,
\end{aligned}
\end{equation}
with the same empty-product convention. Write $Z_n^{\mathrm{syn}}\sim\N(0,C_n^{\mathrm{syn}})$.

\begin{restatable}{proposition}{synproxyprop}\label{syn:proxy}
Under the assumptions of Theorem~\ref{syn:upper}, for every $n\ge2$, we have
\begin{align}
 \E\|X_n-X_n^{\mathrm{lin}}\|_2&\le Cn^{-1/2},
 \label{syn:localization-bound}\\
 W_1(X_n^{\mathrm{lin}},Z_n^{\mathrm{syn}})
 &\le C\{n^{-1/2}\log(en)+n^{-\beta}\},
 \label{syn:finite-bound}\\
 \|C_n^{\mathrm{syn}}-\Sigma_\infty^{\mathrm{syn}}\|_F
 &\le Cn^{\beta-1}.
 \label{syn:covariance-bound}
\end{align}
\end{restatable}

The first term bounds the difference between $X_n$ and $X_n^{\mathrm{lin}}$. The difference is zero when there is one action. When $\|Q_t-Q^\star\|_\infty<\kappa/2$, the policy $\pi^\star$ is greedy with respect to $Q_t$, so the maximization introduces no additional error. Such errors arise only when $Q_t$ lies outside this neighborhood. These exceptional updates become rare enough that their expected errors have a finite sum, giving an $O(n^{-1/2})$ contribution to the normalized average. This gives a sharper bound than using the quadratic estimate of the Bellman remainder.

The second term is the Gaussian approximation error for $X_n^{\mathrm{lin}}$. We use a resampling argument based on the independent sample blocks to obtain the $n^{-1/2}\log(en)$ term. The additional $O(n^{-\beta})$ term accounts for the difference between the true covariance of $X_n^{\mathrm{lin}}$ and the deterministic matrix $C_n^{\mathrm{syn}}$ used in the Gaussian approximation.

The last term is the finite-sample covariance discrepancy, which also appears in the asynchronous bound. Independent sampling removes the accumulated Markovian mean bias of order $n^{1/2-\beta}$. Together with the sharper treatment of the sampled recursion and Bellman nonlinearity, this allows the choice $\beta=1/2$ and improves the optimal polynomial rate from $n^{-1/4}$ to $n^{-1/2}$. 

\subsection{Lower Bound}\label{syn:lower-section}

We use the same two-state, one-action MDP in \eqref{lb:example}, now drawing independent samples from both transition rows at every iteration and initializing $Q_0=0$. Thus, the MDP and Q-function initialization are unchanged, while the sampling scheme is synchronous.

\begin{restatable}{theorem}{synlowerthm}\label{syn:lower}
For the MDP in \eqref{lb:example} with synchronous sampling and $Q_0=0$, and for fixed choice of $c_0,b,\beta$, constants $c>0$ and $n_0<\infty$ such that, for $n\ge n_0$,
\begin{equation}\label{syn:lower-bound}
 W_1(X_n,Z_\infty^{\mathrm{syn}})\ge c
 \begin{cases}
 n^{-1/2},&\beta=1/2,\\
 n^{\beta-1},&1/2<\beta<1.
 \end{cases}
\end{equation}
The MDP and initialization are independent of the stepsize parameters.
\end{restatable}

The difference from the asynchronous lower bound is the absence of the Markovian mean bias. Under asynchronous sampling, the dependence between the current iterate and the visited state produces a bias of order $n^{1/2-\beta}$ on this same one-action MDP. Under synchronous sampling, each new sample block is independent of the current iterate, so the mean error on this linear instance follows a homogeneous deterministic recursion. Thus, the lower bound no longer contains the $n^{1/2-\beta}$ term that forces the asynchronous choice $\beta=3/4$. For $\beta>1/2$, the lower bound comes from the finite-sample covariance discrepancy. The deterministic covariance expansion used in Proposition~\ref{lb:prop-cov} also applies to the synchronous matrices on this example. Letting $\beta=1/2$, we have the following corollary.

\begin{corollary}\label{syn:corollary}
For all sufficiently large $n$, we have
\[
 c n^{-1/2}\le W_1(X_n,Z_\infty^{\mathrm{syn}})
             \le Cn^{-1/2}\log(en).
\]
\end{corollary}

\section{Conclusion and Future Work}

In this paper, we established sharp non-asymptotic Gaussian approximation rates for Polyak-Ruppert averaged Q-learning. For polynomial stepsizes $\alpha_t=c_0(t+b)^{-\beta}$, we proved an $\widetilde O(n^{-1/4}+n^{\beta-1}+n^{1/2-\beta})$ upper bound for asynchronous Q-learning and an $\widetilde O(n^{-1/2}+n^{\beta-1}+n^{-\beta})$ upper bound for synchronous Q-learning, together with matching lower bounds up to logarithmic factors.
Our analysis identifies the main sources of Gaussian approximation error, including Bellman nonlinearity, Markovian noise, and the finite-sample covariance discrepancy. In particular, we show that Markovian sampling can produce an unavoidable mean bias even when the Bellman operator is linear. The trade-off between this bias and the covariance discrepancy explains the optimal choice $\beta=3/4$ and the resulting $n^{-1/4}$ rate in the asynchronous setting. These results establish the fundamental convergence limits of Gaussian approximation for Q-learning and provide a theoretical foundation for finite-sample statistical inference in reinforcement learning. 

We believe that the analysis and techniques developed in this work can be extended to other metrics beyond $W_1$ without major modifications. Nevertheless, there are several important and interesting questions that remain for future work. 

First, this work only focuses on $n$, while the optimal dependence on other problem-specific parameters, such as the state-action space size, the discount factor $\gamma$, and the exploration parameter $\rho$, remains unknown.

Second, although our assumptions are standard in the literature, relaxing them remains important, as they exclude certain classes of MDPs. Assumption \ref{ass:quadratic} is slightly weaker than the corresponding assumptions in previous work \cite{li2023statistical,pmlr-v331-liu26b,rubtsov2026gaussian}, but it still excludes MDPs with multiple optimal actions. Assumption \ref{ass:cov}, which requires the limiting covariance matrix to be positive definite, is also commonly used when there are Markovian noises. However, it excludes MDPs with deterministic transitions, since
$\Sigma_\infty\succ0 \iff \operatorname{Var}_{s'\sim P(\cdot\mid s,a)} [V^\star(s')]>0, \forall(s,a)$.
Relaxing these assumptions while preserving sharp Gaussian approximation rates is an important but challenging direction for future research.

Third, an interesting problem is to study non-asymptotic Gaussian approximation under other stepsize schedules, particularly constant and linearly decaying stepsizes. Following the seminal work of \cite{polyak1992acceleration}, classical convergence results for Polyak-Ruppert averaging typically rely on the following sufficient conditions: (i) $0\leq\sup_t\alpha_t\leq1$, $\alpha_t\downarrow0$, and $t\alpha_t\to\infty$; (ii) $\frac{\alpha_{t-1}-\alpha_t}{\alpha_{t-1}}=o(\alpha_{t-1})$; (iii) $\frac{1}{\sqrt T}\sum_{t=0}^T\alpha_t\to0$; and (iv) $\frac{\sum_{t=0}^T\alpha_t}{T\alpha_T}\leq C$. Polynomial stepsizes $\alpha_t=c_0(t+b)^{-\beta}$ with $\beta\in(1/2,1)$ satisfy these conditions, whereas constant and linearly decaying stepsizes do not. Nevertheless, these conditions are not necessary for Gaussian approximation, and both alternative stepsize schedules may admit Gaussian limits under alternative assumptions. Previous works \citep{li2023statistical,zhang2024constant} have studied Gaussian approximation for asynchronous and synchronous Q-learning beyond polynomial stepsize, but the results are asymptotic. Establishing non-asymptotic Gaussian approximation rates for Q-learning under different stepsize schedules remains open.

Last but not least, Gaussian approximation for the last iterate, beyond the Polyak--Ruppert averaging scheme, is an important direction and has recently attracted increasing attention \citep{chen2025concentration,kong2026finite,butyrin2026gaussian,wei2026gaussian}.

\paragraph{AI Disclosure} The proof was developed with assistance from GPT-5.6 Sol and GPT-6 Astra, which was verified and rewritten by the author. The author takes full responsibility for the correctness, presentation, and communication of the results. The author also made an effort to identify and cite, to the best of the author's knowledge, prior work whose results were used in the proof but were not cited by the model.

\bibliographystyle{alpha}
\bibliography{ref}

@article{butyrin2025improved,
  title={Improved Central Limit Theorem and Bootstrap Approximations for Linear Stochastic Approximation},
  author={Butyrin, Bogdan and Moulines, Eric and Naumov, Alexey and Samsonov, Sergey and Shao, Qi-Man and Zhang, Zhuo-Song},
  journal={arXiv preprint arXiv:2510.12375},
  year={2025}
}

@article{sheshukova2025gaussian,
  title={Gaussian approximation and multiplier bootstrap for stochastic gradient descent},
  author={Sheshukova, Marina and Samsonov, Sergey and Belomestny, Denis and Moulines, Eric and Shao, Qi-Man and Zhang, Zhuo-Song and Naumov, Alexey},
  journal={arXiv preprint arXiv:2502.06719},
  year={2025}
}

@article{chatterjee2008new,
  title={A new method of normal approximation},
  author={Chatterjee, Sourav},
  volume = {36},
  journal = {The Annals of Probability},
  number = {4},
  publisher = {Institute of Mathematical Statistics},
  pages = {1584 -- 1610},
  year={2008}
}

@article{shi2025statistical,
  title={Statistical inference in reinforcement learning: A selective survey},
  author={Shi, Chengchun},
  journal={arXiv preprint arXiv:2502.16195},
  year={2025}
}

@article{wei2026gaussian,
  title={Gaussian approximation and concentration of constant learning-rate stochastic gradient descent},
  author={Wei, Ziyang and Li, Jiaqi and Lou, Zhipeng and Wu, Wei Biao},
  journal={Advances in Neural Information Processing Systems},
  volume={38},
  pages={85003--85033},
  year={2025}
}

@article{butyrin2026gaussian,
  title={Gaussian approximation for two-timescale linear stochastic approximation},
  author={Butyrin, Bogdan and Rubtsov, Artemy and Naumov, Alexey and Ulyanov, Vladimir V and Samsonov, Sergey},
  journal={Proceedings of the AAAI Conference on Artificial Intelligence},
  volume={40},
  number={43},
  pages={36627--36635},
  year={2026},
  doi={10.1609/aaai.v40i43.40986}
}

@article{kong2026finite,
  title={Finite-Sample Wasserstein Error Bounds and Concentration Inequalities for Nonlinear Stochastic Approximation},
  author={Kong, Seo Taek and Srikant, R},
  journal={arXiv preprint arXiv:2602.02445},
  year={2026}
}

@article{chen2025concentration,
  title={Concentration of contractive stochastic approximation: Additive and multiplicative noise},
  author={Chen, Zaiwei and Maguluri, Siva Theja and Zubeldia, Martin},
  journal={The Annals of Applied Probability},
  volume={35},
  number={2},
  pages={1298--1352},
  year={2025},
  publisher={Institute of Mathematical Statistics}
}

@article{bhatia1997,
  title   = {How and Why to Solve the Operator Equation {$AX-XB=Y$}},
  author  = {Bhatia, Rajendra and Rosenthal, Peter},
  journal = {Bulletin of the London Mathematical Society},
  volume  = {29},
  number  = {1},
  pages   = {1--21},
  year    = {1997},
  doi     = {10.1112/S0024609396001828}
}

@book{folland1999,
  title     = {Real Analysis: Modern Techniques and Their Applications},
  author    = {Folland, Gerald B.},
  edition   = {2},
  publisher = {John Wiley \& Sons},
  year      = {1999}
}

@article{jiang2024generalized,
  title   = {A Generalized Version of {Chung}'s Lemma and its Applications},
  author  = {Jiang, Li and Li, Xiao and Milzarek, Andre and Qiu, Junwen},
  journal = {arXiv preprint arXiv:2406.05637},
  year    = {2024},
  note    = {The cited Proposition A.1 and Lemma 2.4(a) are in version 2, revised February 2026},
  url     = {https://arxiv.org/abs/2406.05637v2}
}

@book{levin2017,
  title     = {Markov Chains and Mixing Times},
  author    = {Levin, David A. and Peres, Yuval},
  edition   = {2},
  publisher = {American Mathematical Society},
  year      = {2017},
  note      = {With contributions by Elizabeth L. Wilmer},
  url       = {https://pages.uoregon.edu/dlevin/MARKOV/mcmt2e.pdf}
}

@article{pinelis1994,
  title   = {Optimum Bounds for the Distributions of Martingales in {Banach} Spaces},
  author  = {Pinelis, Iosif},
  journal = {The Annals of Probability},
  volume  = {22},
  number  = {4},
  pages   = {1679--1706},
  year    = {1994},
  url     = {https://arxiv.org/abs/1208.2200}
}

@article{raic2019,
  title   = {A Multivariate Central Limit Theorem for {Lipschitz} and Smooth Test Functions},
  author  = {Rai\v{c}, Martin},
  journal = {arXiv preprint arXiv:1812.08268},
  year    = {2019},
  note    = {Version 2; the bound used here is equation (3.5) in Section 3},
  url     = {https://arxiv.org/abs/1812.08268v2}
}

@book{villani2009,
  title     = {Optimal Transport: Old and New},
  author    = {Villani, C\'edric},
  series    = {Grundlehren der mathematischen Wissenschaften},
  volume    = {338},
  publisher = {Springer},
  year      = {2009},
  doi       = {10.1007/978-3-540-71050-9}
}

@book{khalil2002nonlinear,
  title={Nonlinear systems},
  author={Khalil, Hassan K},
  edition={3},
  year={2002},
  publisher={Prentice hall, Upper Saddle River, NJ}
}

@article{mokkadem2006convergence,
  title={Convergence rate and averaging of nonlinear two-time-scale stochastic approximation algorithms},
  author={Mokkadem, Abdelkader and Pelletier, Mariane},
  journal={The Annals of Applied Probability},
  volume={16},
  number={3},
  pages={1671--1702},
  year={2006},
  doi={10.1214/105051606000000448}
}

@article{khamaru2021instance,
  title={Instance-optimality in optimal value estimation: Adaptivity via variance-reduced Q-learning},
  author={Khamaru, Koulik and Xia, Eric and Wainwright, Martin J and Jordan, Michael I},
  journal={arXiv preprint arXiv:2106.14352},
  year={2021}
}

@article{rubtsov2026gaussian,
  title={Gaussian Approximation for Asynchronous Q-learning},
  author={Rubtsov, Artemy and Samsonov, Sergey and Ulyanov, Vladimir and Naumov, Alexey},
  journal={arXiv preprint arXiv:2604.07323},
  year={2026}
}

@InProceedings{pmlr-v331-liu26b,
  title = 	 {Central Limit Theorems for Asynchronous Averaged Q-Learning},
  author =       {Liu, Xingtu},
  booktitle = 	 {Proceedings of The 8th Annual Learning for Dynamics and Control Conference},
  pages = 	 {2207--2230},
  year = 	 {2026},
  volume = 	 {331},
  series = 	 {Proceedings of Machine Learning Research},
  month = 	 {17--19 Jun},
  publisher =    {PMLR}
}

@article{li2026weak,
  title={Weak Convergence Rates for Partial-Sum Processes of Nonlinear Stochastic Approximation},
  author={Li, Xiang and Liang, Jiadong and Zhang, Zhihua},
  journal={arXiv preprint arXiv:2609.40338},
  year={2026}
}

@article{even2003learning,
  title={Learning rates for Q-learning},
  author={Even-Dar, Eyal and Mansour, Yishay},
  journal={Journal of machine learning Research},
  volume={5},
  number={Dec},
  pages={1--25},
  year={2003}
}

@article{beck2012error,
  title={Error bounds for constant step-size Q-learning},
  author={Beck, Carolyn L and Srikant, Rayadurgam},
  journal={Systems \& control letters},
  volume={61},
  number={12},
  pages={1203--1208},
  year={2012},
  publisher={Elsevier}
}

@article{li2024q,
  title={Is Q-learning minimax optimal? a tight sample complexity analysis},
  author={Li, Gen and Cai, Changxiao and Chen, Yuxin and Wei, Yuting and Chi, Yuejie},
  journal={Operations Research},
  volume={72},
  number={1},
  pages={222--236},
  year={2024},
  publisher={INFORMS}
}

@article{samsonov2025statistical,
  title={Statistical inference for Linear Stochastic Approximation with Markovian Noise},
  author={Samsonov, Sergey and Sheshukova, Marina and Moulines, Eric and Naumov, Alexey},
  journal={arXiv preprint arXiv:2505.19102},
  year={2025}
}

@article{panda2025asymptotic,
  title={Asymptotic Analysis of Sample-averaged Q-learning},
  author={Panda, Saunak Kumar and Liu, Ruiqi and Xiang, Yisha},
  journal={IEEE Transactions on Information Theory},
  year={2025},
  publisher={IEEE}
}

@article{xie2022statistical,
  title={A statistical online inference approach in averaged stochastic approximation},
  author={Xie, Chuhan and Zhang, Zhihua},
  journal={Advances in Neural Information Processing Systems},
  volume={35},
  pages={8998--9009},
  year={2022}
}

@inproceedings{chen2020explicit,
  title={Explicit mean-square error bounds for monte-carlo and linear stochastic approximation},
  author={Chen, Shuhang and Devraj, Adithya and Busic, Ana and Meyn, Sean},
  booktitle={International Conference on Artificial Intelligence and Statistics},
  pages={4173--4183},
  year={2020},
  organization={PMLR}
}

@article{glynn1996liapounov,
  title={A Liapounov bound for solutions of the Poisson equation},
  author={Glynn, Peter W and Meyn, Sean P},
  journal={The Annals of Probability},
  pages={916--931},
  year={1996},
  publisher={JSTOR}
}

@incollection{makowski2002poisson,
  title={The Poisson equation for countable Markov chains: probabilistic methods and interpretations},
  author={Makowski, Armand M and Shwartz, Adam},
  booktitle={Handbook of Markov Decision Processes: Methods and Applications},
  pages={269--303},
  year={2002},
  publisher={Springer}
}

@book{douc2018markov,
  title={Markov chains},
  author={Douc, Randal and Moulines, Eric and Priouret, Pierre and Soulier, Philippe},
  year={2018},
  publisher={Springer}
}

@inproceedings{qu2020finite,
  title={Finite-time analysis of asynchronous stochastic approximation and $ Q $-learning},
  author={Qu, Guannan and Wierman, Adam},
  booktitle={Conference on Learning Theory},
  pages={3185--3205},
  year={2020},
  organization={PMLR}
}

@article{li2020sample,
  title={Sample complexity of asynchronous Q-learning: Sharper analysis and variance reduction},
  author={Li, Gen and Wei, Yuting and Chi, Yuejie and Gu, Yuantao and Chen, Yuxin},
  journal={Advances in neural information processing systems},
  volume={33},
  pages={7031--7043},
  year={2020}
}

@article{zhang2022global,
  title={Global optimality and finite sample analysis of softmax off-policy actor critic under state distribution mismatch},
  author={Zhang, Shangtong and Des Combes, Remi Tachet and Laroche, Romain},
  journal={Journal of Machine Learning Research},
  volume={23},
  number={343},
  pages={1--91},
  year={2022}
}

@article{zhang2024constant,
  title={Constant stepsize q-learning: Distributional convergence, bias and extrapolation},
  author={Zhang, Yixuan and Xie, Qiaomin},
  journal={arXiv preprint arXiv:2401.13884},
  year={2024}
}

@inproceedings{mou2020linear,
  title={On linear stochastic approximation: Fine-grained Polyak-Ruppert and non-asymptotic concentration},
  author={Mou, Wenlong and Li, Chris Junchi and Wainwright, Martin J and Bartlett, Peter L and Jordan, Michael I},
  booktitle={Conference on Learning Theory},
  pages={2947--2997},
  year={2020},
  organization={PMLR}
}

@article{borkar2025ode,
  title={The ODE method for asymptotic statistics in stochastic approximation and reinforcement learning},
  author={Borkar, Vivek and Chen, Shuhang and Devraj, Adithya and Kontoyiannis, Ioannis and Meyn, Sean},
  journal={The Annals of Applied Probability},
  volume={35},
  number={2},
  pages={936--982},
  year={2025},
  publisher={Institute of Mathematical Statistics}
}

@article{han2024decoupled,
  title={Decoupled Functional Central Limit Theorems for Two-Time-Scale Stochastic Approximation},
  author={Han, Yuze and Li, Xiang and Liang, Jiadong and Zhang, Zhihua},
  journal={arXiv preprint arXiv:2412.17070},
  year={2024}
}

@inproceedings{hu2024central,
  title={Central limit theorem for two-timescale stochastic approximation with markovian noise: Theory and applications},
  author={Hu, Jie and Doshi, Vishwaraj and Eun, Do Young },
  booktitle={International Conference on Artificial Intelligence and Statistics},
  pages={1477--1485},
  year={2024},
  organization={PMLR}
}

@article{kong2025nonasymptotic,
  title={Nonasymptotic CLT and Error Bounds for Two-Time-Scale Stochastic Approximation},
  author={Kong, Seo Taek and Zeng, Sihan and Doan, Thinh T and Srikant, R},
  journal={arXiv preprint arXiv:2502.09884},
  year={2025}
}

@inproceedings{samsonov2024gaussian,
  title={Gaussian approximation and multiplier bootstrap for polyak-ruppert averaged linear stochastic approximation with applications to td learning},
  author={Samsonov, Sergey and Moulines, Eric and Shao, Qi-Man and Zhang, Zhuo-Song and Naumov, Alexey},
  booktitle={The Thirty-eighth Annual Conference on Neural Information Processing Systems},
  year={2024}
}

@article{polyak1992acceleration,
  title={Acceleration of stochastic approximation by averaging},
  author={Polyak, Boris T and Juditsky, Anatoli B},
  journal={SIAM journal on control and optimization},
  volume={30},
  number={4},
  pages={838--855},
  year={1992},
  publisher={SIAM}
}

@article{srikant2024rates,
  title={Rates of convergence in the central limit theorem for markov chains, with an application to TD learning},
  author={Srikant, R},
  journal={arXiv preprint arXiv:2401.15719},
  year={2024}
}

@inproceedings{li2023statistical,
  title={A statistical analysis of polyak-ruppert averaged q-learning},
  author={Li, Xiang and Yang, Wenhao and Liang, Jiadong and Zhang, Zhihua and Jordan, Michael I},
  booktitle={International Conference on Artificial Intelligence and Statistics},
  pages={2207--2261},
  year={2023},
  organization={PMLR}
}

@article{chen2021lyapunov,
  title={A Lyapunov theory for finite-sample guarantees of asynchronous Q-learning and TD-learning variants},
  author={Chen, Zaiwei and Maguluri, Siva Theja and Shakkottai, Sanjay and Shanmugam, Karthikeyan},
  journal={arXiv preprint arXiv:2102.01567},
  year={2021}
}

@article{watkins1992q,
  title={Q-learning},
  author={Watkins, Christopher JCH and Dayan, Peter},
  journal={Machine learning},
  volume={8},
  pages={279--292},
  year={1992},
  publisher={Springer}
}

@book{sutton1998reinforcement,
  title={Reinforcement learning: An introduction},
  author={Sutton, Richard S and Barto, Andrew G},
  year={1998},
  publisher={MIT press Cambridge}
}

@book{benveniste2012adaptive,
  title={Adaptive algorithms and stochastic approximations},
  author={Benveniste, Albert and M{\'e}tivier, Michel and Priouret, Pierre},
  volume={22},
  year={2012},
  publisher={Springer Science \& Business Media}
}

@book{borkar2008stochastic,
  title={Stochastic approximation: a dynamical systems viewpoint},
  author={Borkar, Vivek S},
  volume={48},
  year={2008},
  publisher={Springer}
}

\appendix
\AddToHook{env/proof/begin}{\setlength{\parindent}{0pt}}
\newpage
\tableofcontents

\numberwithin{equation}{section}

\section{Proofs for Section \ref{sec_asm}}

In this section, we first prove that the assumptions imposed in \cite{pmlr-v331-liu26b,rubtsov2026gaussian} are sufficient conditions for Assumption \ref{ass:quadratic} to be satisfied. Moreover, we show that these three assumptions are equivalent whenever $ \sum_{x\in\mathcal S}\sum_{b\in\mathcal A}\mu(x,b)P(s\mid x,b)>0$ for every $s\in\mathcal S$.

\begin{assumption}[Assumption from \cite{pmlr-v331-liu26b}]
\label{asp_1}
There exists a constant $C<\infty$ such that
    \[
    \left\|
    (P^{\pi_q}-P^{\pi^\star})(q-Q^\star)
    \right\|_\infty
    \le C\|q-Q^\star\|_\infty^2
    \]
    for every $q\in[0,H]^d$, where $\pi_q$ is a
    deterministic greedy policy with respect to $q$.
\end{assumption}

\begin{assumption}[Assumption from \cite{rubtsov2026gaussian}]
\label{asp_2}
Every state has a unique optimal action, and
    the minimum optimal action gap satisfies
    \[
    \kappa
    :=\min_{s\in\mathcal S}
    \min_{a\ne\pi^\star(s)}
    \bigl\{
    Q^\star(s,\pi^\star(s))-Q^\star(s,a)
    \bigr\}>0.
    \]
\end{assumption}

\begin{lemma}
\label{lem:quadratic_sufficient}
Assumption~\ref{ass:quadratic} holds if either Assumption~\ref{asp_1} or Assumption~\ref{asp_2} holds.
\end{lemma}

\begin{proof}
Recall that
\[
B(q)=D(\mathcal Tq-q),
\qquad
A=D(I-\gamma P^{\pi^\star}),
\]
and
\[
\varphi(\delta)=B(Q^\star+\delta)+A\delta.
\]
Let $q=Q^\star+\delta\in[0,H]^d$. Since
$\mathcal TQ^\star=Q^\star$, we have
\begin{align*}
\varphi(\delta) =D(\mathcal Tq-q)
  +D(I-\gamma P^{\pi^\star})\delta =D\bigl(
    \mathcal Tq-\mathcal TQ^\star
    -\gamma P^{\pi^\star}\delta
  \bigr).
\end{align*}
Using
\[
\mathcal Tq=r+\gamma P^{\pi_q}q,
\qquad
\mathcal TQ^\star=r+\gamma P^{\pi^\star}Q^\star,
\]
we obtain
\begin{equation}
\varphi(\delta)
=\gamma D(P^{\pi_q}-P^{\pi^\star})q.
\label{eq:quadratic_remainder_identity}
\end{equation}

By the definition of $\pi_q$, for every state $s$,
\[
q(s,\pi_q(s))\ge q(s,\pi^\star(s)).
\]
Similarly, the optimality of $\pi^\star$ implies
\[
Q^\star(s,\pi_q(s))
\le Q^\star(s,\pi^\star(s)).
\]
It follows that, componentwise,
\[
(P^{\pi_q}-P^{\pi^\star})q\ge0,
\qquad
(P^{\pi_q}-P^{\pi^\star})Q^\star\le0.
\]
Consequently,
\begin{equation}
0\le
(P^{\pi_q}-P^{\pi^\star})q
\le
(P^{\pi_q}-P^{\pi^\star})(q-Q^\star).
\label{eq:quadratic_policy_comparison}
\end{equation}

We first consider Assumption \ref{asp_1}.
Combining
\eqref{eq:quadratic_remainder_identity}
and \eqref{eq:quadratic_policy_comparison},
and writing
$\mu_{\max}:=\max_{s,a}\mu(s,a)$, we obtain
\begin{align*}
\|\varphi(\delta)\|_\infty
&\le
\gamma\mu_{\max}
\left\|
(P^{\pi_q}-P^{\pi^\star})(q-Q^\star)
\right\|_\infty\\
&\le
\gamma\mu_{\max}C
\|q-Q^\star\|_\infty^2\\
&=
\gamma\mu_{\max}C\|\delta\|_\infty^2.
\end{align*}
Thus Assumption~\ref{ass:quadratic} holds with
$L=\gamma\mu_{\max}C$. We next consider Assumption \ref{asp_2}.
Suppose first that $\|\delta\|_\infty<\frac{\kappa}{2}$.
For every state $s$ and every
$a\ne\pi^\star(s)$,
\begin{align*}
q(s,\pi^\star(s))-q(s,a)
&=Q^\star(s,\pi^\star(s))-Q^\star(s,a) +\delta(s,\pi^\star(s))-\delta(s,a) \ge\kappa-2\|\delta\|_\infty >0.
\end{align*}
Hence $\pi_q=\pi^\star$, and
\eqref{eq:quadratic_remainder_identity} gives
\begin{equation}
\varphi(\delta)=0
\qquad\text{if }
\|\delta\|_\infty<\frac{\kappa}{2}.
\label{eq:quadratic_local_zero}
\end{equation}

Recall that $V(q)(s)=\max_a q(s,a)$. Since
$V(Q^\star)(s)=Q^\star(s,\pi^\star(s))$,
we have
\begin{align*}
&V(Q^\star+\delta)(s)-V(Q^\star)(s)
  -\delta(s,\pi^\star(s))\\
&\qquad=\max_a\bigl\{Q^\star(s,a)+\delta(s,a)\bigr\}
-Q^\star(s,\pi^\star(s))
-\delta(s,\pi^\star(s))
\end{align*}
for an arbitrary perturbation $\delta$. The expression on the right is nonnegative,
since the maximum includes $\pi^\star(s)$.
Moreover, by the optimality of $\pi^\star$,
\begin{align*}
&V(Q^\star+\delta)(s)-V(Q^\star)(s)
  -\delta(s,\pi^\star(s)) \le
\max_a\delta(s,a)-\delta(s,\pi^\star(s)) \le2\|\delta\|_\infty.
\end{align*}
Using the explicit representation of $\varphi(\delta)$, we therefore obtain
\begin{equation}
\|\varphi(\delta)\|_\infty
\le2\gamma\mu_{\max}\|\delta\|_\infty.
\label{eq:quadratic_global_linear}
\end{equation}

Combining
\eqref{eq:quadratic_local_zero}
and \eqref{eq:quadratic_global_linear},
we conclude that
\begin{align*}
\|\varphi(\delta)\|_\infty
&\le
2\gamma\mu_{\max}\|\delta\|_\infty
\mathbf 1\left\{
\|\delta\|_\infty\ge\frac{\kappa}{2}
\right\} \le
\frac{4\gamma\mu_{\max}}{\kappa}
\|\delta\|_\infty^2.
\end{align*}
Thus Assumption~\ref{ass:quadratic} holds with
$L=4\gamma\mu_{\max}/\kappa$. If there is only one action, the Bellman operator is affine and $\varphi$ vanishes identically, and the assumption holds with $L=0$.
\end{proof}

\begin{lemma}
\label{lem:quadratic_equivalence}
Suppose that $\mathcal S$ and $\mathcal A$ are finite,
$|\mathcal A|\ge2$, $0<\gamma<1$, $H>0$, and
$Q^\star\in[0,H]^d$.
Let $D=\operatorname{diag}(\mu)$, where $\mu(s,a)\ge0$,
and assume that
\begin{equation}
\sum_{x\in\mathcal S}\sum_{b\in\mathcal A}
\mu(x,b)P(s\mid x,b)>0
\qquad\text{for every }s\in\mathcal S.
\label{eq:positive_incoming}
\end{equation}
Fix a deterministic optimal policy $\pi^\star$.
For each $q$, let $\pi_q$ be a deterministic greedy
policy with respect to $q$. Then, Assumption \ref{asp_1}, Assumption \ref{asp_2}, and Assumption \ref{ass:quadratic} are equivalent. In particular, one may take $C=\frac{4}{\kappa}, L=\frac{4\gamma\mu_{\max}}{\kappa}, \mu_{\max}:=\max_{s,a}\mu(s,a)$ for Assumption \ref{asp_2}.
\end{lemma}

\begin{proof}
Recall that
\[
B(q)=D(\mathcal Tq-q),
\qquad
A=D(I-\gamma P^{\pi^\star}),
\qquad
\varphi(\delta)=B(Q^\star+\delta)+A\delta.
\]
For $q=Q^\star+\delta$, the Bellman fixed-point equation
gives
\begin{align}
\varphi(\delta) =D\bigl(
\mathcal Tq-\mathcal TQ^\star
-\gamma P^{\pi^\star}\delta
\bigr) \ =\gamma D(P^{\pi_q}-P^{\pi^\star})q.
\label{eq:equivalence_remainder}
\end{align}

We first prove Assumption \ref{asp_2} $\Rightarrow$ Assumption \ref{asp_1}.
Write $\rho=\|q-Q^\star\|_\infty$.
If $\rho<\kappa/2$, then, for every state $s$ and
every $a\ne\pi^\star(s)$,
\begin{align*}
q(s,\pi^\star(s))-q(s,a) \ge
Q^\star(s,\pi^\star(s))-Q^\star(s,a)-2\rho \ge\kappa-2\rho>0.
\end{align*}
Thus $\pi_q=\pi^\star$, and
\[
(P^{\pi_q}-P^{\pi^\star})(q-Q^\star)=0.
\]
If $\rho\ge\kappa/2$, the stochasticity of
$P^{\pi_q}$ and $P^{\pi^\star}$ implies
\begin{align*}
\left\|
(P^{\pi_q}-P^{\pi^\star})(q-Q^\star)
\right\|_\infty \le2\rho \le\frac{4}{\kappa}\rho^2.
\end{align*}
Therefore Assumption \ref{asp_1} holds with $C=4/\kappa$.

We next prove Assumption \ref{asp_1} $\Rightarrow$ Assumption \ref{ass:quadratic}.
The greediness of $\pi_q$ and the optimality of
$\pi^\star$ imply, componentwise,
\[
(P^{\pi_q}-P^{\pi^\star})q\ge0,
\qquad
(P^{\pi_q}-P^{\pi^\star})Q^\star\le0.
\]
Consequently,
\[
0\le
(P^{\pi_q}-P^{\pi^\star})q
\le
(P^{\pi_q}-P^{\pi^\star})(q-Q^\star).
\]
Combining this inequality with
\eqref{eq:equivalence_remainder} yields
\begin{align*}
\|\varphi(\delta)\|_\infty \le
\gamma\mu_{\max}
\left\|
(P^{\pi_q}-P^{\pi^\star})(q-Q^\star)
\right\|_\infty \le
\gamma\mu_{\max}C\|\delta\|_\infty^2.
\end{align*}
Thus Assumption \ref{ass:quadratic} holds with $L=\gamma\mu_{\max}C$.

It remains to prove Assumption \ref{ass:quadratic} $\Rightarrow$ Assumption \ref{asp_2}.
For any feasible perturbation $\delta$, define
\[
h_\delta(s)
:=
\max_{a\in\mathcal A}
\bigl\{Q^\star(s,a)+\delta(s,a)\bigr\}
-Q^\star(s,\pi^\star(s))
-\delta(s,\pi^\star(s)).
\]
By the Bellman equation,
\begin{equation}
\varphi(\delta)(x,b)
=
\gamma\mu(x,b)
\sum_{s\in\mathcal S}
P(s\mid x,b)h_\delta(s).
\label{eq:equivalence_state_remainder}
\end{equation}

Suppose, towards a contradiction, that some state $s_0$
has an optimal action $a_1\ne a_0:=\pi^\star(s_0)$.
Write
\[
v:=Q^\star(s_0,a_0)=Q^\star(s_0,a_1).
\]
Let $e_{s,a}$ denote the coordinate vector corresponding
to $(s,a)$. For sufficiently small $t>0$, define
\[
\delta_t
=
\begin{cases}
t e_{s_0,a_1}, & v<H,\\
-t e_{s_0,a_0}, & v=H.
\end{cases}
\]
In the first case, take $0<t<H-v$; in the second,
take $0<t<H$.
Then $Q^\star+\delta_t\in[0,H]^d$ and
$\|\delta_t\|_\infty=t$. In the first case, increasing the value of $a_1$ raises the maximum at $s_0$ by $t$, while leaving the value of $a_0$ unchanged. In the second case, decreasing the value of $a_0$ leaves the maximum unchanged, since $a_1$ remains optimal at value $H$. Hence, in both cases,
\[
h_{\delta_t}(s)
=
\begin{cases}
t, & s=s_0,\\
0, & s\ne s_0.
\end{cases}
\]
By \eqref{eq:positive_incoming}, there exists
$(x,b)$ such that
\[
\mu(x,b)P(s_0\mid x,b)>0.
\]
Using \eqref{eq:equivalence_state_remainder} and
Assumption \ref{ass:quadratic}, we obtain
\[
\gamma\mu(x,b)P(s_0\mid x,b)t
\le
\|\varphi(\delta_t)\|_\infty
\le Lt^2.
\]
Dividing by $t$ and letting $t\downarrow0$ gives
\[
\gamma\mu(x,b)P(s_0\mid x,b)\le0,
\]
a contradiction. Therefore, every state has a unique optimal action. Since the state and action spaces are finite,
the minimum optimal action gap is strictly positive.
This proves Assumption \ref{asp_2} and completes the proof.
\end{proof}

\section{Proof of Theorem \ref{mb:upper}}\label{app:main}

In this section, we prove Theorem \ref{mb:upper}. The proof of Theorem \ref{mb:upper} is built on Proposition \ref{mb:expansion} and Proposition \ref{mb:finite}. Proposition \ref{mb:expansion} proves upper bounds on the error remainder term $R_n$ and the finite-sample covariance discrepancy $\norm{C_n-\Sigma_\infty}_F$. Proposition \ref{mb:finite} provides an upper bound on the Gaussian approximation of $M_n$. Section \ref{app:support} contains two problem-specific bounds. Lemma \ref{lem:moments} provides moment bounds for the iterate error, showing that \(\|\Delta_t\|_{L^p(\ell_\infty)}\le C\sqrt{p\alpha_t}\), which is essential for controlling the Bellman nonlinearity and Markovian noise. Lemma \ref{lem:weights} establishes uniform boundedness, bounded total variation, and convergence properties of the Polyak–Ruppert averaging weights \(W_{n,t}\). Section \ref{sec_ubsl} provides supporting lemmas, which are mainly probability tools.

\maintheorem*

\begin{proof}[Proof of Theorem~\ref{mb:upper}]

We first compare the finite-sample covariance with the limiting
covariance. Denote $\lambda_\infty=\lambda_{\min}(\Sigma_\infty)>0$.
By Proposition~\ref{mb:expansion}, there exists a constant $C_{\rm cov}$ such that
\[
\norm{C_n-\Sigma_\infty}_F\le C_{\rm cov}n^{\beta-1}.
\]
Since $\beta<1$, we can choose $n_0$ such that the above bound is
at most $\lambda_\infty/2$ for $n\ge n_0$ and
Proposition~\ref{mb:finite} holds in this range. Using the fact that the
operator norm is bounded by the Frobenius norm, for every unit vector
$u$, we have
\[
u^\top C_nu
\ge u^\top\Sigma_\infty u-\norm{C_n-\Sigma_\infty}_{\op}
\ge\frac{\lambda_\infty}{2}.
\]
Thus, $C_n$ is positive definite for all $n\ge n_0$, and its
smallest eigenvalue is bounded below by a constant independent of $n$.
By Lemma~\ref{lem:gaussian}, we obtain
\begin{align}
W_1(Z_n,Z_\infty)
&\le
\frac{\norm{C_n-\Sigma_\infty}_F}
{\sqrt{\lambda_{\min}(C_n)}
+\sqrt{\lambda_{\min}(\Sigma_\infty)}}\notag\\
&\le
\frac{C_{\rm cov}n^{\beta-1}}
{\sqrt{\lambda_\infty/2}+\sqrt{\lambda_\infty}}
\le Cn^{\beta-1}.
\label{eq:full-upper-gaussian}
\end{align}

Next, recall that $X_n=M_n+R_n$ holds on the original probability
space. We use this decomposition to couple $X_n$ and $M_n$ when bounding
$W_1(X_n,M_n)$. By the comparison between the Euclidean and maximum
norms and Cauchy--Schwarz, we have
\begin{align}
W_1(X_n,M_n)
&\le\E\norm{X_n-M_n}_2
=\E\norm{R_n}_2\notag\\
&\le\sqrt d\,\E\norm{R_n}_\infty
\le\sqrt d\,(\E\norm{R_n}_\infty^2)^{1/2}
=\sqrt d\,\norm{R_n}_{L^2(\ell_\infty)}.
\label{eq:full-upper-remainder}
\end{align}
Using \eqref{mb:expansion-bound} at $p=2$, we obtain
\[
W_1(X_n,M_n)
\le C\{n^{-\beta/2}+n^{1/2-\beta}+\elln n^{-1/2}\}.
\]
Combining the triangle inequality, Proposition~\ref{mb:finite}, and
\eqref{eq:full-upper-gaussian}, we obtain
\begin{align}
W_1(X_n,Z_\infty)
&\le W_1(X_n,M_n)+W_1(M_n,Z_n)+W_1(Z_n,Z_\infty)\notag\\
&\le C\{n^{-\beta/2}+n^{1/2-\beta}
+\elln n^{-1/2}+\elln n^{-1/4}+n^{\beta-1}\}.
\label{eq:full-upper-five-terms}
\end{align}
Since $n\ge1$, $\beta>1/2$, and $\elln\ge1$, we have
\[
n^{-\beta/2}\le n^{-1/4},\qquad
\elln n^{-1/2}\le\elln n^{-1/4},\qquad
n^{1/2-\beta}+n^{\beta-1}
\le\elln(n^{1/2-\beta}+n^{\beta-1}).
\]
Substituting these inequalities into
\eqref{eq:full-upper-five-terms} proves \eqref{mb:upper-bound} for $n\ge n_0$.

We now consider the remaining finitely many horizons.
Since $Q_k,Q^\star\in[0,H]^d$, we have
$\norm{X_n}_2\le H\sqrt{dn}$. Also,
$\E\norm{Z_\infty}_2\le\sqrt{\tr(\Sigma_\infty)}$.
Taking an independent coupling gives
\[
W_1(X_n,Z_\infty)
\le H\sqrt{dn}+\sqrt{\tr(\Sigma_\infty)}.
\]
Denote
$a_n=\elln(n^{-1/4}+n^{\beta-1}+n^{1/2-\beta})$.
Since this quantity is positive, the maximum of
$[H\sqrt{dn}+\sqrt{\tr(\Sigma_\infty)}]/a_n$
over the finite set $2\le n<n_0$ is finite. Increasing $C$ to include
this maximum proves the result for all the remaining horizons.
Finally, when $\beta=3/4$, both $n^{\beta-1}$ and $n^{1/2-\beta}$
are equal to $n^{-1/4}$. For fixed $d,b$, we have $\elln=O(\log n)$,
which gives the stated rate and completes the proof.
\end{proof}

\subsection{Proof of Proposition \ref{mb:expansion}}\label{app:expansion}

\begin{restatable}{proposition}{lemmaa}
\label{mb:expansion}
There is an exact decomposition $X_n=M_n+R_n$. For $n\ge1$ and
$p\ge2$,
\begin{equation}
 \lp{R_n}\le C\{p^{3/2}n^{-\beta/2}
                  +pn^{1/2-\beta}+\elln n^{-1/2}\},
 \qquad
 \norm{C_n-\Sigma_\infty}_F\le Cn^{\beta-1}.
 \label{mb:expansion-bound}
\end{equation}
\end{restatable}

\begin{proof}
We first derive the decomposition of the averaged error. We then bound
the initialization term, the nonlinear term, and the centered Markov
term using Lemmas~\ref{lem:moments} and~\ref{lem:weights}. The covariance
bound follows from the same deterministic weight estimates.

For a deterministic $\delta$ in
$\mathcal E=\{\delta:Q^\star+\delta\in[0,H]^d\}$, define
\begin{equation}
 \mathcal U_\delta(y)=U(Q^\star+\delta,y)-U(Q^\star,y),\qquad
 f_\delta(y)=\mathcal U_\delta(y)-\E_{\bar\mu}\mathcal U_\delta(Y).
 \label{eq:expanded-c-eq-fe}
\end{equation}
Recall that
$\norm{V(q)-V(q')}_\infty\le\norm{q-q'}_\infty$.
By the definition of the sampled update, we have
\[
 \sup_y\norm{U(q,y)-U(q',y)}_\infty
 \le(1+\gamma)\norm{q-q'}_\infty.
\]
Subtracting the stationary means on both sides, we obtain
\begin{equation}
 \sup_y\norm{f_\delta(y)-f_{\delta'}(y)}_\infty
 \le2(1+\gamma)\norm{\delta-\delta'}_\infty.
 \label{eq:expanded-c-eq-fe-lip}
\end{equation}
By definition, $\bar\mu f_\delta=0$ and $f_0=0$. Thus,
$\sup_y\norm{f_\delta(y)}_\infty\le C\norm \delta_\infty$.
Since $B(Q^\star)=0$, we also have
\[
 \E_{\bar\mu}\mathcal U_\delta(Y)=B(Q^\star+\delta)=-A\delta+\varphi(\delta).
\]
Substituting $\delta=\Delta_t$ into \eqref{eq:expanded-c-eq-fe} and
using the update rule, we obtain
\begin{equation}
 \Delta_{t+1}
 =(I-\alpha_tA)\Delta_t+
 \alpha_t\{\varepsilon_t+f_{\Delta_t}(Y_t)+\varphi(\Delta_t)\}.
 \label{eq:expanded-c-eq-exact-recursion}
\end{equation}
The terms $\varphi$ and $f_{\Delta_t}$ in this identity account for
the full difference between the original update and the linear drift.
Thus, the recursion above is exact.

Denote $\zeta_t=\varepsilon_t+f_{\Delta_t}(Y_t)+\varphi(\Delta_t)$.
By recursion in \eqref{eq:expanded-c-eq-exact-recursion}, we have
\[
 \Delta_k=\Phi_{0:k-1}\Delta_0+
          \sum_{t=0}^{k-1}\alpha_t\Phi_{t+1:k-1}\zeta_t.
\]
Recall that $X_n=n^{-1/2}\sum_{k=1}^n\Delta_k$.
Summing the above identity and interchanging the two finite sums gives
\begin{align*}
 \frac1{\sqrt n}\sum_{k=1}^n
      \sum_{t=0}^{k-1}\alpha_t\Phi_{t+1:k-1}\zeta_t
 &=\frac1{\sqrt n}\sum_{t=0}^{n-1}
      \left(\alpha_t\sum_{k=t+1}^n\Phi_{t+1:k-1}\right)\zeta_t\\
 &=\frac1{\sqrt n}\sum_{t=0}^{n-1}W_{n,t}\zeta_t.
\end{align*}
Therefore, the decomposition holds with
\begin{align}
 R_n&=R_n^{\mathrm{init}}+R_n^{\mathrm{nonlin}}
                        +R_n^{\mathrm{Markov}},\label{eq:expanded-c-eq-split}\\
 R_n^{\mathrm{init}}
 &=\frac1{\sqrt n}\sum_{k=1}^n\Phi_{0:k-1}\Delta_0,\label{eq:expanded-c-eq-init}\\
 R_n^{\mathrm{nonlin}}
 &=\frac1{\sqrt n}\sum_{t<n}W_{n,t}\varphi(\Delta_t),\label{eq:expanded-c-eq-nonlin}\\
 R_n^{\mathrm{Markov}}
 &=\frac1{\sqrt n}\sum_{t<n}W_{n,t}f_{\Delta_t}(Y_t).
 \label{eq:expanded-c-eq-markov}
\end{align}

We bound the three terms in this decomposition separately.
For the initialization term, integral comparison gives
\[
 \sum_{u=0}^{k-1}\alpha_u
 \ge c_0\int_0^k(x+b)^{-\beta}\,dx
 =\frac{c_0}{1-\beta}
   \bigl((k+b)^{1-\beta}-b^{1-\beta}\bigr).
\]
By \eqref{eq:product-bound}, we have
$\norm{\Phi_{0:k-1}}_\infty\le Ce^{-ck^{1-\beta}}$.
To bound the sum $\sum_{k\ge1}e^{-ck^{1-\beta}}$, let $r=1-\beta>0$.
By a change of variable in the corresponding integral, we have
\[
\int_0^\infty e^{-cx^r}\,dx
=\frac{c^{-1/r}}{r}\int_0^\infty e^{-z}z^{1/r-1}\,dz
=\frac{c^{-1/r}}{r}\Gamma(1/r)<\infty.
\]
It follows by integral comparison that the series is bounded
by a constant independent of the horizon. Since
$\norm{\Delta_0}_\infty\le H$, we obtain
\begin{equation}
 \lp{R_n^{\mathrm{init}}}
 \le\frac H{\sqrt n}\sum_{k=1}^\infty
                  \norm{\Phi_{0:k-1}}_\infty
 \le Cn^{-1/2}.
 \label{eq:expanded-c-eq-init-bound}
\end{equation}
The above estimate holds deterministically, so its constant is
independent of $p$.

For the nonlinear term, by the triangle inequality in $L^p$,
the uniform weight bound, and the quadratic remainder assumption,
we have
\begin{align}
 \lp{R_n^{\mathrm{nonlin}}}
 &\le\frac C{\sqrt n}\sum_{t<n}
                   \norm{\norm{\Delta_t}_\infty^2}_{L^p}
 \notag\\
 &\le Cpn^{1/2-\beta}.
 \label{eq:expanded-c-eq-nonlin-bound}
\end{align}
where the second inequality follows from \eqref{eq:moment-sums}.
Thus, summing the quadratic remainders gives a contribution of order
$n^{1/2-\beta}$.

We next consider the centered Markov term. For every fixed $\delta$,
$f_\delta$ is centered under $\bar\mu$. However, after substituting the
iterate, its conditional mean is
\[
\E[f_{\Delta_t}(Y_t)\mid\cF_{t-1}]
=(Kf_{\Delta_t})(Y_{t-1}),
\]
which is not necessarily zero. We therefore use the Poisson
equation to express this term as a martingale difference and two
remaining differences. For each $\delta\in\mathcal E$, define
\begin{equation}
 h_\delta=\sum_{j=0}^\infty K^jf_\delta,\qquad v_\delta=Kh_\delta.
 \label{eq:expanded-c-eq-poisson}
\end{equation}
Since $\bar\mu f_\delta=0$, $f_0=0$, and
\eqref{eq:expanded-c-eq-fe-lip} holds, Lemma~\ref{lem:poisson} implies
that the series converges absolutely and satisfies
\begin{align}
 \sup_y\norm{h_\delta(y)}_\infty&\le C\norm \delta_\infty,\label{eq:expanded-c-eq-h-size}\\
 \sup_y\norm{h_\delta(y)-h_{\delta'}(y)}_\infty
 &\le C\norm{\delta-\delta'}_\infty.
 \label{eq:expanded-c-eq-h-lip}
\end{align}
Since a Markov kernel does not increase the sup norm, both
bounds also hold for $v_\delta$. Applying $K$ to the convergent series
term by term, we obtain
\begin{equation}
 h_\delta-Kh_\delta=f_\delta.
 \label{eq:expanded-c-eq-poisson-identity}
\end{equation}
The Lipschitz bounds imply measurability in the parameter, so
these identities can also be evaluated at the random iterate $\Delta_t$.
For $t\ge1$, define
\begin{equation}
 m_t=h_{\Delta_t}(Y_t)-v_{\Delta_t}(Y_{t-1}).
 \label{eq:expanded-c-eq-mt}
\end{equation}
Since $\Delta_t$ is $\cF_{t-1}$-measurable, the Markov property gives
\[
 \E[h_{\Delta_t}(Y_t)\mid\cF_{t-1}]
 =\sum_yh_{\Delta_t}(y)K(Y_{t-1},y)
 =v_{\Delta_t}(Y_{t-1}).
\]
Thus, $\E[m_t\mid\cF_{t-1}]=0$. By
\eqref{eq:expanded-c-eq-h-size}, we also have
$\norm{m_t}_\infty\le C\norm{\Delta_t}_\infty$.
Substituting the definition into the Poisson identity gives
\begin{equation}
 f_{\Delta_t}(Y_t)
 =m_t+v_{\Delta_t}(Y_{t-1})-v_{\Delta_t}(Y_t).
 \label{eq:expanded-c-eq-poisson-decomp}
\end{equation}

Since $W_{n,t}$ is deterministic, each coordinate of
$W_{n,t}m_t$ is a martingale difference. Applying
Lemma \ref{lem:martingale} coordinatewise and using
\eqref{eq:weight-uniform}, we have
\begin{align}
 \lp{\frac1{\sqrt n}\sum_{t=1}^{n-1}W_{n,t}m_t}
 &\le\frac{Cp}{\sqrt n}
          \left(\sum_{t=1}^{n-1}\lp{\Delta_t}^{\,2}\right)^{1/2}
 \notag\\
 &\le\frac{Cp}{\sqrt n}(Cpn^{1-\beta})^{1/2}
 \le Cp^{3/2}n^{-\beta/2}.
 \label{eq:expanded-c-eq-martingale-bound}
\end{align}
The coordinatewise estimate changes the constant only through
its dependence on $d$. We use the factor $Cp$ in the martingale bound,
although the scalar calculation also gives the factor $C\sqrt p$.
The former is sufficient for the claimed remainder estimate.

For $n\ge2$, denote $g_t=v_{\Delta_t}$. By summation by parts,
we have
\begin{align}
 &\sum_{t=1}^{n-1}W_{n,t}\{g_t(Y_{t-1})-g_t(Y_t)\}
 \notag\\
 &\quad=W_{n,1}g_1(Y_0)-W_{n,n-1}g_{n-1}(Y_{n-1})
 \notag\\
 &\qquad+\sum_{t=1}^{n-2}
          \{W_{n,t+1}g_{t+1}(Y_t)-W_{n,t}g_t(Y_t)\}.
 \label{eq:expanded-c-eq-summation}
\end{align}
Since $\norm{\Delta_t}_\infty\le H$, both endpoint terms are
bounded by $C$. For each interior term, we have the decomposition
\begin{align}
 W_{n,t+1}g_{t+1}(Y_t)-W_{n,t}g_t(Y_t)
 ={}&(W_{n,t+1}-W_{n,t})g_t(Y_t)
 \notag\\
 &+W_{n,t+1}\{g_{t+1}(Y_t)-g_t(Y_t)\}.
 \label{eq:expanded-c-eq-two-variations}
\end{align}
For the first term, by \eqref{eq:weight-total}, we have
\begin{align*}
 \sum_{t=1}^{n-2}
  \norm{(W_{n,t+1}-W_{n,t})g_t(Y_t)}_\infty \le C\sum_{t=1}^{n-2}
                  \norm{W_{n,t+1}-W_{n,t}}_\infty \le C\elln.
\end{align*}
For the second term, using the Lipschitz bound for $v_\delta$ and
the one-step movement bound in \eqref{eq:pathwise}, we obtain
\[
 \sup_y\norm{g_{t+1}(y)-g_t(y)}_\infty
 \le C\norm{\Delta_{t+1}-\Delta_t}_\infty
 \le CH\alpha_t.
\]
Combining the endpoint estimates and the two bounds above, we
obtain, for every sample path,
\begin{equation}
 \norm{\sum_{t=1}^{n-1}W_{n,t}
    \{v_{\Delta_t}(Y_{t-1})-v_{\Delta_t}(Y_t)\}}_\infty
 \le C(1+\elln+n^{1-\beta}).
 \label{eq:expanded-c-eq-coboundary}
\end{equation}
Dividing by $\sqrt n$ and taking the $L^p$ norm gives
\begin{equation}
 C\{\elln n^{-1/2}+n^{1/2-\beta}\}.
 \label{eq:expanded-c-eq-coboundary-bound}
\end{equation}
These bounds hold for every sample path and do not use
conditional centering of the two remaining differences.

The Poisson decomposition was applied from $t=1$ onward, without
introducing $Y_{-1}$. We bound the time-zero term directly as follows.
\begin{equation}
 \norm{\frac1{\sqrt n}W_{n,0}f_{\Delta_0}(Y_0)}_\infty
 \le\frac C{\sqrt n}\norm{\Delta_0}_\infty
 \le Cn^{-1/2}.
 \label{eq:expanded-c-eq-time-zero}
\end{equation}
Combining \eqref{eq:expanded-c-eq-poisson-decomp},
\eqref{eq:expanded-c-eq-martingale-bound},
\eqref{eq:expanded-c-eq-coboundary-bound}, and
\eqref{eq:expanded-c-eq-time-zero}, we have
\[
 \lp{R_n^{\mathrm{Markov}}}
 \le C\{p^{3/2}n^{-\beta/2}
          +n^{1/2-\beta}+\elln n^{-1/2}\}.
\]
Combining this estimate with \eqref{eq:expanded-c-eq-split},
\eqref{eq:expanded-c-eq-init-bound}, and
\eqref{eq:expanded-c-eq-nonlin-bound}, we obtain
\[
 \lp{R_n}\le
 C\{p^{3/2}n^{-\beta/2}
       +(p+1)n^{1/2-\beta}
       +(\elln+1)n^{-1/2}\}.
\]
Using $p\ge2$ and $\elln\ge1$, the above inequality gives the first inequality in \eqref{mb:expansion-bound}. When $n=1$, the martingale and coboundary sums
are empty. The bounded initialization, nonlinear, and time-zero terms
give the same inequality after increasing $C$.

We now prove the second inequality in \eqref{mb:expansion-bound}.
Denote $T=A^{-1}$. For any matrix $B$, we have
\begin{equation}
 BSB^\top-TST^\top
 =(B-T)SB^\top+TS(B-T)^\top.
 \label{eq:expanded-cov-eq-covariance-identity}
\end{equation}
Expanding the right-hand side verifies the identity, since
the two cross terms cancel. Using
$\norm{XYZ}_F\le\norm{X}_F\norm{Y}_{\op}\norm{Z}_{\op}$ and the
same inequality after transposing, we obtain
\[
 \norm{BSB^\top-TST^\top}_F
 \le\norm{S}_{\op}
       (\norm{B}_{\op}+\norm{T}_{\op})\norm{B-T}_F.
\]
Taking $B=W_{n,t}$, the two operator norms are uniformly
bounded by \eqref{eq:weight-uniform}, \eqref{eq:product-bound}, and
norm equivalence in fixed dimension. Also,
$\norm{B-T}_F\le\sqrt d\,\norm{B-T}_\infty$. Therefore, we have
\begin{align*}
 \norm{C_n-\Sigma_\infty}_F
 &\le\frac1n\sum_{t=0}^{n-1}
        \norm{W_{n,t}SW_{n,t}^\top-TST^\top}_F\\
 &\le\frac{C}{n}\sum_{t=0}^{n-1}
        \norm{W_{n,t}-A^{-1}}_\infty
 \le Cn^{\beta-1},
\end{align*}
where the last inequality follows from
\eqref{eq:weight-average}. This completes the proof.
\end{proof}

\subsection{Proof of Proposition \ref{mb:finite}}\label{app:normal}

\begin{restatable}{proposition}{lemmab}
\label{mb:finite}
For all sufficiently large $n$, if $Z_n\sim\N(0,C_n)$, then
\begin{equation}
 W_1(M_n,Z_n)\le C\elln n^{-1/4}.
 \label{mb:finite-bound}
\end{equation}
\end{restatable}

\begin{proof}
By Lemma~\ref{lem:weights} and the boundedness of the fixed-point
noise, there exists a constant $K_0<\infty$ such that
\begin{equation}\label{eq:expanded-normal-eq-increment}
\norm{W_{n,t}\varepsilon_t}_2\le K_0,\qquad
\norm{W_{n,t}\varepsilon'_t}_2\le K_0
\end{equation}
almost surely, uniformly in $n,t$.
We first partition the trajectory into retained intervals separated
by gaps. Let
\begin{equation}\label{eq:expanded-normal-eq-q-m}
q=\lceil\sqrt n\rceil,\qquad
m=\lceil C_1\log(en)\rceil,
\end{equation}
where $C_1$ is a sufficiently large constant to be chosen below.
For all sufficiently large $n$, we have $m<q$.
We partition $\{0,\ldots,n-1\}$ by first discarding $m$ indices,
then retaining the next $q$ indices, discarding the next $m$ indices,
and retaining the next $q$ indices. We continue alternating in this way.
The last retained interval can contain fewer than $q$ indices.
All final indices that are not retained are included in the discarded
set. Denote the retained intervals by $I_1,\ldots,I_N$ and let
\[
I=\bigcup_{j=1}^NI_j,\qquad
J=\{0,\ldots,n-1\}\setminus I.
\]
By this construction, the number of retained intervals is
bounded by $N\le C(n/q+1)$. Each gap contains at most $m$ indices,
so the total number of discarded indices satisfies
\begin{equation}\label{eq:expanded-normal-eq-gaps-count}
|J|\le C(n/q+1)m\le C\sqrt n\,\ell_n.
\end{equation}
Define
\begin{equation}\label{eq:expanded-normal-eq-blocks}
B_j=\frac1{\sqrt n}\sum_{t\in I_j}W_{n,t}\varepsilon_t,
\qquad
G_n^{\gap}=\frac1{\sqrt n}\sum_{t\in J}W_{n,t}\varepsilon_t.
\end{equation}
By the definitions above, we have
\begin{equation}\label{eq:expanded-normal-eq-decomp}
M_n=\sum_{j=1}^N B_j+G_n^{\gap}.
\end{equation}

Applying Lemma~\ref{lem:coupling} to the trajectory segments
$(Y_t:t\in I_j)$, we obtain independent stationary copies of the
retained segments. Each copy has the same dependence within its
interval as a stationary trajectory of that length.
Using the same deterministic weights, define
\begin{equation}\label{eq:expanded-normal-eq-ind-blocks}
B'_j=\frac1{\sqrt n}\sum_{t\in I_j}W_{n,t}\varepsilon'_t.
\end{equation}
By choosing the fixed constant $C_1$ sufficiently large, we have,
for all sufficiently large $n$,
\begin{equation}\label{eq:expanded-normal-eq-good-event}
\Prob(\mathcal E_n^c)\le CN e^{-c_{\rm mix}m}\le n^{-5},
\qquad
\mathcal E_n=\{B_j=B'_j\text{ for every }j\}.
\end{equation}
The copied segments are defined on an extension of the
original probability space. In this construction, the original
discarded observations and retained observations keep their joint law.

We first bound the sum over the discarded indices. By the martingale difference property of \eqref{eq:noise}, increments at
distinct times are orthogonal, including when the indices in $J$ are
not consecutive. Thus, using \eqref{eq:expanded-normal-eq-increment},
we have
\begin{align}
\E\|G_n^{\gap}\|_2^2
&=\frac1n\sum_{t\in J}\E\|W_{n,t}\varepsilon_t\|_2^2
\le\frac{K_0^2|J|}{n},\label{eq:expanded-normal-eq-gap-L2}\\
\E\|G_n^{\gap}\|_2
&\le K_0\sqrt{|J|/n}
\le Cn^{-1/4}\sqrt{\ell_n}.\label{eq:expanded-normal-eq-gap-L1}
\end{align}
The term at $t=0$ also has mean zero, so it is included in
the orthogonality calculation of \eqref{eq:noise}.
Alternatively, it can be bounded directly by $K_0/\sqrt n$.
The above calculation does not require independence between
$G_n^{\gap}$ and the sum over the retained intervals.

Next, denote $L_j=|I_j|\le q$. Within each stationary copied interval,
the noise increments are bounded martingale differences. Applying
Azuma--Hoeffding coordinatewise and integrating the tail gives,
for every fixed $p\ge2$,
\begin{equation}\label{eq:expanded-normal-eq-block-moment}
\left(\E\left\|\sum_{t\in I_j}W_{n,t}\varepsilon'_t\right\|_2^p\right)^{1/p}
\le C_{d,p}\sqrt{L_j}.
\end{equation}
Taking $p=3$, we obtain
\begin{equation}\label{eq:expanded-normal-eq-block-third}
\E\|B'_j\|_2^3\le Cn^{-3/2}L_j^{3/2}.
\end{equation}
Summing over the retained intervals and using
$L_j^{3/2}\le\sqrt q\,L_j$, we have
\begin{align}
\sum_{j=1}^N\E\|B'_j\|_2^3
&\le Cn^{-3/2}\sum_jL_j^{3/2}
\le Cn^{-3/2}\sqrt q\sum_jL_j\\
&\le C\sqrt{q/n}\le Cn^{-1/4}.\label{eq:expanded-normal-eq-third-sum}
\end{align}
Thus, the sum of the third moments has order $n^{-1/4}$, which
will determine the polynomial order in the Gaussian approximation.

Denote $B'_n=\sum_jB'_j$. Each copied sum has mean zero.
Using independence between the copied intervals and martingale
orthogonality within each interval, we obtain
\begin{equation}\label{eq:expanded-normal-eq-CI}
\mathrm{Cov}(B'_n)=C_n^I,
\qquad
C_n^I=\frac1n\sum_{t\in I}W_{n,t}S W_{n,t}^\top.
\end{equation}
where the single-time second moment is $S$ by stationarity
and \eqref{eq:noise}. Since the terms omitted from the full covariance
are precisely those indexed by $J$, we have
\begin{equation}\label{eq:expanded-normal-eq-CI-Cn}
\|C_n^I-C_n\|_F
=\left\|\frac1n\sum_{t\in J}W_{n,t}S W_{n,t}^\top\right\|_F
\le C|J|/n\le C\ell_n n^{-1/2}.
\end{equation}
Combining the above bound with \eqref{mb:expansion-bound}, we obtain
\begin{equation}\label{eq:expanded-normal-eq-CI-Sigma}
\|C_n^I-\Sigma_\infty\|_F\le C(n^{\beta-1}+\ell_n n^{-1/2})=o(1).
\end{equation}
Since $\Sigma_\infty\succ0$, there exist fixed constants
$0<\lambda_*<\Lambda_*<\infty$ such that, for all sufficiently
large $n$,
\begin{equation}\label{eq:expanded-normal-eq-eig}
\lambda_*I_d\preceq C_n^I,C_n\preceq\Lambda_*I_d.
\end{equation}
This is where Assumption~\ref{ass:cov} is used in the proof.
We now apply the Gaussian approximation bound for independent sums.
Using the covariance of the retained sum, define
\begin{equation}\label{eq:expanded-normal-eq-whiten}
\xi_j=(C_n^I)^{-1/2}B'_j,
\qquad
\sum_j\E\xi_j\xi_j^\top=I_d.
\end{equation}
By \eqref{eq:expanded-normal-eq-eig} and \eqref{eq:expanded-normal-eq-third-sum},
\begin{equation}\label{eq:expanded-normal-eq-rho3}
\rho_3=\sum_j\E\|\xi_j\|_2^3
\le\lambda_*^{-3/2}\sum_j\E\|B'_j\|_2^3
\le Cn^{-1/4}.
\end{equation}
For sufficiently large $n$, we have $\rho_3\le1/2$.
By Lemma~\ref{lem:independent}, the extension to a general covariance
given after its proof, and \eqref{eq:expanded-normal-eq-rho3}, we obtain
\begin{equation}\label{eq:expanded-normal-eq-w1-ind}
W_1\bigl(B'_n,\N(0,C_n^I)\bigr)
\le C\rho_3(1+|\log\rho_3|)
\le C\ell_n n^{-1/4}.
\end{equation}
where the last inequality follows because
$x\mapsto x(1-\log x)$ is increasing on $(0,1)$ and
$\rho_3\le Cn^{-1/4}$.
For all sufficiently large $n$, using the eigenvalue bounds in
\eqref{eq:expanded-normal-eq-eig}, we apply Lemma~\ref{lem:gaussian}
to obtain
\begin{equation}\label{eq:expanded-normal-eq-target-comp}
W_1(\N(0,C_n^I),\N(0,C_n))
\le C\norm{C_n^I-C_n}_F
\le C\ell_n n^{-1/2}.
\end{equation}

It remains to compare the copied sum with the original sum.
Denote $B_n=\sum_jB_j$, so $M_n=B_n+G_n^{\gap}$.
There are at most $n$ summands in each of $B_n$ and $B'_n$.
By \eqref{eq:expanded-normal-eq-increment}, we have
\[
\|B_n\|_2\le K_0\sqrt n,\qquad \|B'_n\|_2\le K_0\sqrt n.
\]
Since $B_n=B'_n$ on the event $\mathcal E_n$, we have
\begin{equation}\label{eq:expanded-normal-eq-coupling-W1}
\E\|B_n-B'_n\|_2
=\E\bigl[\|B_n-B'_n\|_2\mathbf1_{\mathcal E_n^c}\bigr]
\le 2K_0\sqrt n\,\Prob(\mathcal E_n^c)
\le Cn^{-9/2}.
\end{equation}
Let $Z_n^I\sim\N(0,C_n^I)$ and $Z_n\sim\N(0,C_n)$.
We couple the original and copied retained sums as above, and couple
$M_n$ and $B_n$ through their decomposition on the original space.
Combining the preceding estimates with the triangle inequality gives
\begin{align}
W_1(M_n,Z_n)
&\le \E\|G_n^{\gap}\|_2
+\E\|B_n-B'_n\|_2
+W_1(B'_n,Z_n^I)
+W_1(Z_n^I,Z_n)\\
&\le C\left\{n^{-1/4}\sqrt{\ell_n}+n^{-9/2}
+n^{-1/4}\ell_n+n^{-1/2}\ell_n\right\}\\
&\le Cn^{-1/4}\ell_n.\label{eq:expanded-normal-eq-w1-final}
\end{align}

This proves \eqref{mb:finite-bound} and completes the proof.
\end{proof}

\subsection{Problem Specific Bounds}\label{app:support}

\begin{lemma}\label{lem:moments}
Under Assumption~\ref{ass:mix}, for every
$t\ge0$ and $p\ge2$,
\begin{equation}
 \lp{\Delta_t}\le C\sqrt{p\alpha_t}.
 \label{eq:last-iterate}
\end{equation}
Also, pathwise,
\begin{equation}
 \norm{\Delta_t}_\infty\le H,\qquad
 \norm{\Delta_{t+1}-\Delta_t}_\infty\le H\alpha_t.
 \label{eq:pathwise}
\end{equation}
Consequently,
\begin{equation}
 \sum_{t<n}\lp{\Delta_t}^{\,2}\le Cpn^{1-\beta},\qquad
 \sum_{t<n}\norm{\norm{\Delta_t}_\infty^2}_{L^p}
 \le Cpn^{1-\beta}.
 \label{eq:moment-sums}
\end{equation}
\end{lemma}

\begin{proof}
We first bound the cumulative centered update over an arbitrary
interval. We then apply this bound on intervals with fixed cumulative
stepsize and compare the iterates with the mean recursion.

We begin by showing that the iterates remain bounded.
Suppose $Q_t\in[0,H]^d$. For the updated coordinate $i=(s_t,a_t)$,
the update rule gives
\[
Q_{t+1}(i)=(1-\alpha_t)Q_t(i)
+\alpha_t\bigl[r(i)+\gamma V(Q_t)(s_{t+1})\bigr].
\]
Since $0\le r(i)+\gamma V(Q_t)(s_{t+1})\le1+\gamma H=H$ and
$0\le\alpha_t\le1$, the updated coordinate is a convex combination
of two values in $[0,H]$. The other coordinates remain unchanged.
By induction from $Q_0$, we have $Q_t\in[0,H]^d$ for all $t$.
Since $Q^\star\in[0,H]^d$, this gives $\norm{\Delta_t}_\infty\le H$.
The update changes only one coordinate and its magnitude is at most
$H\alpha_t$. Hence,
\[
\norm{Q_{t+1}-Q_t}_\infty
=\norm{\Delta_{t+1}-\Delta_t}_\infty\le H\alpha_t.
\]
This proves \eqref{eq:pathwise}.

We now express the update as its stationary mean and a centered
perturbation. For $y=(s,a,s')$, recall that
\[
U(q,y)=\mathbf u_{s,a}\bigl[r(s,a)+\gamma V(q)(s')-q(s,a)\bigr],
\quad B(q)=\E_{\bar\mu}U(q,Y).
\]
By the definition of $D=\operatorname{diag}(\mu)$, we have
$B(q)=D(\mathcal Tq-q)$ and $B(Q^\star)=0$.
For any $q,q'\in[0,H]^d$, the Lipschitz property of the maximum gives
\begin{align*}
\norm{U(q,y)-U(q',y)}_\infty
&\le(1+\gamma)\norm{q-q'}_\infty\le2\norm{q-q'}_\infty,\\
\norm{B(q)-B(q')}_\infty
&\le(1+\gamma)\norm{q-q'}_\infty\le2\norm{q-q'}_\infty.
\end{align*}
Both functions are uniformly bounded on the cube. Define
\[
F_q(y)=U(q,y)-B(q).
\]
By the bounds above, $\bar\mu F_q=0$,
$\sup_{q,y}\norm{F_q(y)}_\infty\le C$, and
$\sup_y\norm{F_q(y)-F_{q'}(y)}_\infty\le C\norm{q-q'}_\infty$.
Substituting the definition into the original recursion, we have
\begin{equation}
Q_{t+1}=Q_t+\alpha_t B(Q_t)+\alpha_t F_{Q_t}(Y_t).
\label{eq:expanded-moment-eq-meanfieldnoise}
\end{equation}
Although $F_{Q_t}(Y_t)$ is centered under $\bar\mu$ for each
fixed parameter, it is not necessarily a martingale difference.
The iterate $Q_t$ and the observation $Y_t$ both depend on the past
trajectory. We therefore apply the Poisson equation.
For every $q\in[0,H]^d$, define
\begin{equation}
\psi_q(y)=\sum_{j=0}^{\infty}(K^jF_q)(y),
\qquad (K^jf)(y)=\int f(z)K^j(y,dz).
\label{eq:expanded-moment-eq-poisson}
\end{equation}
Since $F_q$ is centered, uniformly bounded, and Lipschitz in
$q$, Lemma~\ref{lem:poisson} gives
\begin{equation}
\sup_y\norm{\psi_q(y)}_\infty\le C,
\qquad
\sup_y\norm{\psi_q(y)-\psi_{q'}(y)}_\infty
\le C\norm{q-q'}_\infty.
\label{eq:expanded-moment-eq-poissonbounds}
\end{equation}
Applying $K$ to the series term by term gives
\begin{equation}
\psi_q-K\psi_q=F_q.
\label{eq:expanded-moment-eq-poissonidentity}
\end{equation}
Denote $g_q=K\psi_q$. Since a Markov kernel does not increase
the sup norm, the same boundedness and Lipschitz estimates hold for
$g_q$.

Let $\cF_t=\sigma(Q_0,Y_0,\ldots,Y_t)$.
For $t\ge1$, $Q_t$ is $\cF_{t-1}$-measurable.
By \eqref{eq:expanded-moment-eq-poissonidentity}, we have
\begin{equation}
F_{Q_t}(Y_t)=M_t+g_{Q_t}(Y_{t-1})-g_{Q_t}(Y_t),
\qquad
M_t:=\psi_{Q_t}(Y_t)-g_{Q_t}(Y_{t-1}).
\label{eq:expanded-moment-eq-decomp}
\end{equation}
Since $Q_t$ is measurable with respect to the past, the Markov
property gives
\[
\E[M_t\mid\cF_{t-1}]
=K\psi_{Q_t}(Y_{t-1})-g_{Q_t}(Y_{t-1})=0.
\]
By \eqref{eq:expanded-moment-eq-poissonbounds}, we also have
$\norm{M_t}_\infty\le C$ almost surely.
Thus, $(M_t)_{t\ge1}$ is a bounded vector-valued martingale difference
sequence.

For integers $1\le u<v$, define
\[
S_k^M=\sum_{t=u}^{k-1}\alpha_tM_t,\qquad u\le k\le v.
\]
Applying the maximal Azuma--Hoeffding inequality to each
coordinate, we obtain
\[
\Prob\!\left(\max_{u\le k\le v}|(S_k^M)_i|>x\right)
\le2\exp\!\left(-\frac{c x^2}{\sum_{t=u}^{v-1}\alpha_t^2}\right).
\]
Taking a union bound over the fixed $d$ coordinates and
integrating $px^{p-1}\Prob(\cdot>x)$ over $x>0$, we have, for $p\ge2$,
\begin{equation}
\norm{\max_{u\le k\le v}\norm{S_k^M}_\infty}_{L^p}
\le C\sqrt{p\sum_{t=u}^{v-1}\alpha_t^2}.
\label{eq:expanded-moment-eq-martmax}
\end{equation}
The dependence on the fixed dimension is included in $C$.
We now bound the remaining differences in the Poisson decomposition.
Denote $g_t=g_{Q_t}$. For $u<k\le v$, summation by parts gives
\begin{align}
&\sum_{t=u}^{k-1}\alpha_t[g_t(Y_{t-1})-g_t(Y_t)]\nonumber\\
&\quad=\alpha_u g_u(Y_{u-1})
-\alpha_{k-1}g_{k-1}(Y_{k-1})
+\sum_{t=u}^{k-2}
\bigl[\alpha_{t+1}g_{t+1}(Y_t)-\alpha_tg_t(Y_t)\bigr].
\label{eq:expanded-moment-eq-summationbyparts}
\end{align}
When $k=u$, the sum is empty and equals zero, so this index
does not change the maximum over $u\le k\le v$.
Since $\alpha_t$ is decreasing and $g_q$ is uniformly bounded, both
endpoint terms have sup norm at most $C\alpha_u$.
For the interior terms, we use the decomposition
\begin{align*}
\alpha_{t+1}g_{t+1}(Y_t)-\alpha_tg_t(Y_t)
={}(\alpha_{t+1}-\alpha_t)g_t(Y_t)+\alpha_{t+1}[g_{t+1}(Y_t)-g_t(Y_t)].
\end{align*}
For the first term, by monotonicity of the stepsizes, we have
\[
\sum_{t=u}^{k-2}|\alpha_{t+1}-\alpha_t|\norm{g_t}_\infty
\le C\sum_{t=u}^{k-2}(\alpha_t-\alpha_{t+1})
\le C\alpha_u.
\]
For the second term, using
\eqref{eq:expanded-moment-eq-poissonbounds} and \eqref{eq:pathwise},
we have
\[
\norm{g_{t+1}-g_t}_\infty
\le C\norm{Q_{t+1}-Q_t}_\infty
\le C\alpha_t.
\]
Substituting this inequality into the sum gives
\[
\sum_{t=u}^{k-2}\alpha_{t+1}
\norm{g_{t+1}-g_t}_\infty
\le C\sum_{t=u}^{v-1}\alpha_t^2.
\]
All the preceding estimates hold for each sample path and
simultaneously over $k$. Combining them, we obtain
\begin{equation}
\max_{u\le k\le v}
\norm{\sum_{t=u}^{k-1}\alpha_t
[g_t(Y_{t-1})-g_t(Y_t)]}_\infty
\le C\left(\alpha_u+\sum_{t=u}^{v-1}\alpha_t^2\right).
\label{eq:expanded-moment-eq-remaindermax}
\end{equation}
Combining \eqref{eq:expanded-moment-eq-decomp},
\eqref{eq:expanded-moment-eq-martmax}, and
\eqref{eq:expanded-moment-eq-remaindermax}, we conclude that
\begin{equation}
{\quad
\norm{\max_{u\le k\le v}
\norm{\sum_{t=u}^{k-1}\alpha_tF_{Q_t}(Y_t)}_\infty}_{L^p}
\le C\left(
\sqrt{p\sum_{t=u}^{v-1}\alpha_t^2}
+\alpha_u+\sum_{t=u}^{v-1}\alpha_t^2\right).
\quad}
\label{eq:expanded-moment-eq-mainnoise}
\end{equation}
For $u=0$, the first term $\alpha_0F_{Q_0}(Y_0)$ has norm at
most $C\alpha_0$. We bound this term directly and apply the preceding
calculation from $t=1$ onward.
Thus, \eqref{eq:expanded-moment-eq-mainnoise} also holds for $u=0$
after increasing $C$ if necessary.

We next divide the time indices into intervals with a fixed cumulative
stepsize. Let
\[
T_0=\frac{\log4}{a_\star},\qquad t_0=0,
\quad
 t_{j+1}=\min\left\{v>t_j:
\sum_{t=t_j}^{v-1}\alpha_t\ge T_0\right\}.
\]
Since $\sum_t\alpha_t=\infty$ for $\beta<1$, each endpoint is
well defined. For a block $[u,v]=[t_j,t_{j+1}]$, by the minimality
of $v$ and the bound $\alpha_t\le1$, we have
\begin{equation}
T_0\le\sum_{t=u}^{v-1}\alpha_t<T_0+1.
\label{eq:expanded-moment-eq-blockmass}
\end{equation}
We first compare the stepsizes at the two endpoints.
Using $\alpha_t=c_0(t+b)^{-\beta}$ and the shift condition, we have
\begin{align*}
\log\frac{\alpha_u}{\alpha_v}
&=\beta\sum_{t=u}^{v-1}\log\!\left(1+\frac1{t+b}\right)\\
&\le\beta\sum_{t=u}^{v-1}\frac1{t+b}\\
&=\frac\beta{c_0}\sum_{t=u}^{v-1}
(t+b)^{\beta-1}\alpha_t\\
&\le\frac\beta{c_0}(u+b)^{\beta-1}
\sum_{t=u}^{v-1}\alpha_t\\
&\le\frac{a_\star}{4}(T_0+1)
=\frac{\log4+a_\star}{4}<\log2.
\end{align*}
where the last inequality follows from
$a_\star=\rho(1-\gamma)\le1$. Therefore,
\begin{equation}
\frac{\alpha_u}{\alpha_v}<2,
\quad\text{and hence}\quad
\alpha_u\le2\alpha_t\qquad (u\le t\le v).
\label{eq:expanded-moment-eq-ratio}
\end{equation}
Since the stepsizes are nonincreasing, we also have
\begin{equation}
\sum_{t=u}^{v-1}\alpha_t^2
\le\alpha_u\sum_{t=u}^{v-1}\alpha_t
\le(T_0+1)\alpha_u.
\label{eq:expanded-moment-eq-squaresum}
\end{equation}
Substituting \eqref{eq:expanded-moment-eq-squaresum} into
\eqref{eq:expanded-moment-eq-mainnoise} gives a bound for the noise on
each block. Since $\alpha_u\le1$ and $p\ge2$, we have
$\alpha_u\le\sqrt{p\alpha_u}$. Hence,
\begin{equation}
{\quad
\norm{\max_{u\le k\le v}
\norm{\sum_{t=u}^{k-1}\alpha_tF_{Q_t}(Y_t)}_\infty}_{L^p}
\le C\sqrt{p\alpha_u}.
\quad}
\label{eq:expanded-moment-eq-blocknoise}
\end{equation}

For the same block $[u,v]$, we define the mean recursion by
\[
z_u=Q_u,
\qquad z_{t+1}=z_t+\alpha_tB(z_t)
\quad(u\le t<v).
\]
We first show that the mean-update map
$G_\alpha(q)=q+\alpha B(q)$ preserves $[0,H]^d$.
Its $i$th coordinate is
$(1-\alpha\mu_i)q_i+\alpha\mu_i(\mathcal Tq)_i$,
which is a convex combination of two values in $[0,H]$.
For $q\in[0,H]^d$ and $0\le\alpha\le1$, the Bellman contraction gives
\begin{align*}
|G_\alpha(q)_i-Q_i^\star|
&\le(1-\alpha\mu_i)|q_i-Q_i^\star|
+\alpha\mu_i|(\mathcal Tq)_i-(\mathcal TQ^\star)_i|\\
&\le[1-\alpha\mu_i(1-\gamma)]
\norm{q-Q^\star}_\infty\\
&\le(1-a_\star\alpha)\norm{q-Q^\star}_\infty.
\end{align*}
Taking the maximum over $i$ and iterating over the block, we
obtain
\begin{align}
\norm{z_v-Q^\star}_\infty
&\le\prod_{t=u}^{v-1}(1-a_\star\alpha_t)
\norm{Q_u-Q^\star}_\infty\nonumber\\
&\le\exp\!\left(-a_\star\sum_{t=u}^{v-1}\alpha_t\right)
\norm{\Delta_u}_\infty
\le\frac14\norm{\Delta_u}_\infty.
\label{eq:expanded-moment-eq-detcontract}
\end{align}

Denote $E_k=Q_k-z_k$.
Subtracting the two recursions and using $E_u=0$, we have
\begin{equation}
E_k=\sum_{t=u}^{k-1}\alpha_tF_{Q_t}(Y_t)
+\sum_{t=u}^{k-1}\alpha_t[B(Q_t)-B(z_t)].
\label{eq:expanded-moment-eq-erroridentity}
\end{equation}
Set
\[
N_{u,v}=\max_{u\le k\le v}
\norm{\sum_{t=u}^{k-1}\alpha_tF_{Q_t}(Y_t)}_\infty.
\]
Since $B$ is $2$-Lipschitz, the preceding identity implies
\[
\norm{E_k}_\infty
\le N_{u,v}+2\sum_{t=u}^{k-1}\alpha_t\norm{E_t}_\infty.
\]
Applying the discrete Gronwall inequality to each sample
path, we obtain
\begin{align*}
\max_{u\le k\le v}\norm{E_k}_\infty \le N_{u,v}\prod_{t=u}^{v-1}(1+2\alpha_t) \le N_{u,v}\exp\!\left(2\sum_{t=u}^{v-1}\alpha_t\right) \le e^{2(T_0+1)}N_{u,v}.
\end{align*}
Taking the $L^p$ norm and using
\eqref{eq:expanded-moment-eq-blocknoise}, we conclude that
\begin{equation}
{\quad
\norm{\max_{u\le k\le v}\norm{Q_k-z_k}_\infty}_{L^p}
\le C\sqrt{p\alpha_u}.
\quad}
\label{eq:expanded-moment-eq-comparison}
\end{equation}
In the calculation above, we first sum the noise increments
and then take the norm. This retains cancellation in the cumulative
noise. Taking the norm of each increment before summing would not
give the square-root bound.

We now bound the error at the block endpoints. Combining the triangle
inequality, \eqref{eq:expanded-moment-eq-detcontract}, and
\eqref{eq:expanded-moment-eq-comparison}, we have
\begin{equation}
\norm{\Delta_v}_{L^p(\ell_\infty)}
\le\frac14\norm{\Delta_u}_{L^p(\ell_\infty)}
+C\sqrt{p\alpha_u}.
\label{eq:expanded-moment-eq-blockrecursion}
\end{equation}
To compare the error with the current stepsize, define
\[
d_j=\frac{\norm{\Delta_{t_j}}_{L^p(\ell_\infty)}}
{\sqrt{p\alpha_{t_j}}}.
\]
By \eqref{eq:expanded-moment-eq-ratio} and
\eqref{eq:expanded-moment-eq-blockrecursion}, we obtain
\begin{equation}
d_{j+1}
\le\frac14\sqrt{\frac{\alpha_{t_j}}{\alpha_{t_{j+1}}}}d_j
+C\sqrt{\frac{\alpha_{t_j}}{\alpha_{t_{j+1}}}}
\le\frac{\sqrt2}{4}d_j+C\sqrt2
\le\frac12d_j+C.
\label{eq:expanded-moment-eq-normalizedrecursion}
\end{equation}
Iterating the last inequality gives
\[
d_j\le2^{-j}d_0+C\sum_{i=0}^{j-1}2^{-i}
\le d_0+2C.
\]
Since
$d_0\le H/\sqrt{p\alpha_0}\le H/\sqrt{2\alpha_0}$
and $\alpha_0$ is fixed, the initial normalized error is bounded by
a constant independent of $p$. Therefore,
\begin{equation}
\norm{\Delta_{t_j}}_{L^p(\ell_\infty)}
\le C\sqrt{p\alpha_{t_j}}
\qquad\text{for all }j\ge0.
\label{eq:expanded-moment-eq-endpointbound}
\end{equation}

Finally, we extend the endpoint bound to all intermediate
times. Fix $t\ge0$ and choose $j$ such that
$u=t_j\le t\le t_{j+1}=v$.
By contraction of the mean recursion up to time $t$ and the comparison
bound, we have
\begin{align*}
\norm{\Delta_t}_{L^p(\ell_\infty)}
&\le\norm{Q_t-z_t}_{L^p(\ell_\infty)}
+\norm{z_t-Q^\star}_{L^p(\ell_\infty)}\\
&\le C\sqrt{p\alpha_u}
+\norm{\Delta_u}_{L^p(\ell_\infty)}\\
&\le C\sqrt{p\alpha_u}.
\end{align*}
Using $\alpha_u\le2\alpha_t$ from
\eqref{eq:expanded-moment-eq-ratio}, we obtain
\[
\norm{\Delta_t}_{L^p(\ell_\infty)}
\le C\sqrt{p\alpha_t},
\]
which proves \eqref{eq:last-iterate}.

It remains to bound the sums of quadratic moments.
By \eqref{eq:last-iterate} and the sum of the power stepsizes, we have
\[
\sum_{t=0}^{n-1}\norm{\Delta_t}_{L^p(\ell_\infty)}^2
\le Cp\sum_{t=0}^{n-1}\alpha_t
\le Cpn^{1-\beta}.
\]
For the second bound, by Minkowski's inequality and the
identity
$\norm{\norm{\Delta_t}_\infty^2}_{L^p}
=\norm{\Delta_t}_{L^{2p}(\ell_\infty)}^2$, we obtain
\begin{align*}
\norm{\sum_{t=0}^{n-1}\norm{\Delta_t}_\infty^2}_{L^p}
&\le\sum_{t=0}^{n-1}
\norm{\norm{\Delta_t}_\infty^2}_{L^p}\\
&=\sum_{t=0}^{n-1}\norm{\Delta_t}_{L^{2p}(\ell_\infty)}^2\\
&\le C(2p)\sum_{t=0}^{n-1}\alpha_t
\le Cpn^{1-\beta}.
\end{align*}
This proves \eqref{eq:moment-sums} and completes the proof.
\end{proof}

\begin{lemma}\label{lem:weights}
For all $n\ge1$,
\begin{align}
 \sup_{t<n}\norm{W_{n,t}}_\infty&\le C,\label{eq:weight-uniform}\\
 \sum_{t=0}^{n-2}\norm{W_{n,t+1}-W_{n,t}}_\infty
 &\le C\elln,\label{eq:weight-total}\\
 \frac1n\sum_{t<n}\norm{W_{n,t}-A^{-1}}_\infty
 &\le Cn^{\beta-1}.\label{eq:weight-average}
\end{align}
The sum in \eqref{eq:weight-total} is zero for $n=1$.
For every $0\le t<n$,
\begin{equation}
 \norm{W_{n,t}-A^{-1}}_\infty
 \le C\left((t+b)^{\beta-1}
       +\exp\left\{-\frac{a_\star}{2}
                     \sum_{u=t+1}^n\alpha_u\right\}\right).
 \label{eq:weight-pointwise}
\end{equation}
Also,
\begin{equation}
 \norm{\Phi_{i:j}}_\infty
 \le\exp\left(-a_\star\sum_{u=i}^j\alpha_u\right),
 \qquad \norm{A^{-1}}_\infty\le a_\star^{-1}.
 \label{eq:product-bound}
\end{equation}
\end{lemma}

\begin{proof}
We first bound the matrix products and compare the stepsizes at
different indices. We then use these estimates to compare the
infinite-horizon weights with $A^{-1}$ and bound the finite-sample
tail. Finally, we prove the bound for adjacent weights.

Recall that $A=D(I-\gamma P^{\pi^\star})$.
For $0\le\alpha\le1$, we have
\[
 I-\alpha A=I-\alpha D+\alpha\gamma DP^{\pi^\star}.
\]
Since $\mu_i\le1$, all diagonal entries of $I-\alpha D$ are
nonnegative. The entries of $DP^{\pi^\star}$ are also nonnegative, so the full
matrix has nonnegative entries. For row $i$, the row sum is
\[
 1-\alpha\mu_i+\alpha\gamma\mu_i
 =1-\alpha\mu_i(1-\gamma)\le1-a_\star\alpha.
\]
The infinity norm of a nonnegative matrix equals its largest
row sum. Thus, by submultiplicativity and $1-x\le e^{-x}$, we have
\begin{equation}
 \norm{\Phi_{i:j}}_\infty
 \le\prod_{u=i}^j(1-a_\star\alpha_u)
 \le\exp\left(-a_\star\sum_{u=i}^j\alpha_u\right).
 \label{eq:expanded-weight-eq-product}
\end{equation}
When $i>j$, the same inequality holds by the empty-product
convention.

The matrix $A$ is invertible since $\norm{P^{\pi^\star}}_\infty=1$ and $\gamma<1$.
Its inverse can be written as
\[
 A^{-1}=(I-\gamma P^{\pi^\star})^{-1}D^{-1}
 =\left(\sum_{r=0}^\infty\gamma^r(P^{\pi^\star})^r\right)D^{-1}.
\]
Taking the infinity norm, we obtain
\begin{equation}
 \norm{A^{-1}}_\infty
 \le\frac{1}{1-\gamma}\frac1\rho
 =\frac1{a_\star}.
 \label{eq:expanded-weight-eq-inverse}
\end{equation}
Since $\sum_u\alpha_u=\infty$ for $\beta<1$, it follows from
\eqref{eq:expanded-weight-eq-product} that $\Phi_{i:j}\to0$ as
$j\to\infty$ for each fixed $i$.

For $k>t$, define the cumulative stepsize and its comparison parameter by
\begin{equation}
 s_{t,k}=\sum_{u=t+1}^{k-1}\alpha_u,\qquad
 \eta_t=\frac{\beta}{c_0}(t+b)^{\beta-1}.
 \label{eq:expanded-weight-eq-clock}
\end{equation}
Since $\beta-1$ is negative, $\eta_t$ decreases with $t$. Note that $b$ is chosen to satisfy $\frac{\beta}{c_0}b^{\beta-1}\le\frac{a_\star}{4}$, and thus we have
$\eta_t\le\eta_0\le a_\star/4$.
Using the formula for the stepsizes, we obtain
\begin{align}
 \log\frac{\alpha_t}{\alpha_k}
 &=\beta\log\frac{k+b}{t+b}
 =\beta\sum_{u=t}^{k-1}\log\left(1+\frac1{u+b}\right)
 \notag\\
 &\le\beta\sum_{u=t}^{k-1}\frac1{u+b}
 =\sum_{u=t}^{k-1}
   \frac{\beta}{c_0}(u+b)^{\beta-1}\alpha_u
 \notag\\
 &\le\eta_t\sum_{u=t}^{k-1}\alpha_u
 =\eta_t(\alpha_t+s_{t,k}).
 \label{eq:expanded-weight-eq-ratio}
\end{align}
Thus, the relative change in the stepsize can be bounded
using the cumulative stepsize that appears in the product bound.
Exponentiating the preceding inequality gives
\begin{equation}
 \frac{\alpha_t}{\alpha_k}
 \le e^{\eta_t\alpha_t}e^{\eta_t s_{t,k}}
 \le e^{a_\star/4}e^{(a_\star/4)s_{t,k}}.
 \label{eq:expanded-weight-eq-ratio-exp}
\end{equation}
The exponent in this bound has a smaller coefficient than
the contraction exponent in \eqref{eq:expanded-weight-eq-product}.
We use this difference to bound the sums below.

For fixed $t$ and $\lambda>0$, the cumulative sums $s_{t,k}$ satisfy
\[
 s_{t,t+1}=0,\qquad
 s_{t,k+1}-s_{t,k}=\alpha_k.
\]
Since the cumulative sum diverges, the intervals
$[s_{t,k},s_{t,k+1}]$, $k=t+1,t+2,\ldots$, partition $[0,\infty)$
up to their common endpoints. Each interval has length $\alpha_k\le1$.
For $x\in[s_{t,k},s_{t,k+1}]$, we have
\[
 e^{-\lambda s_{t,k}}\le e^\lambda e^{-\lambda x},
 \qquad 1+s_{t,k}\le1+x.
\]
Integrating over the interval, we obtain
\[
 \alpha_k(1+s_{t,k})e^{-\lambda s_{t,k}}
 \le e^\lambda\int_{s_{t,k}}^{s_{t,k+1}}
                   (1+x)e^{-\lambda x}\,dx.
\]
Summing these inequalities over $k$ gives
\begin{equation}
 \sum_{k>t}\alpha_k(1+s_{t,k})e^{-\lambda s_{t,k}}
 \le e^\lambda\int_0^\infty(1+x)e^{-\lambda x}\,dx
 =e^\lambda\left(\frac1\lambda+\frac1{\lambda^2}\right).
 \label{eq:expanded-weight-eq-exp-sum}
\end{equation}
Since the factor $1+s_{t,k}$ is at least one, the sum without it is also
bounded uniformly in $t$.

We now define the infinite-horizon weight by
\begin{equation}
 W_{\infty,t}
 =\alpha_t\sum_{k=t+1}^\infty\Phi_{t+1:k-1}.
 \label{eq:expanded-weight-eq-infinite-weight}
\end{equation}
To show that this series converges, we insert the factor
$\alpha_k/\alpha_k$ and apply
\eqref{eq:expanded-weight-eq-product} and
\eqref{eq:expanded-weight-eq-ratio-exp}. We have
\begin{align*}
 \alpha_t\sum_{k>t}\norm{\Phi_{t+1:k-1}}_\infty \le\sum_{k>t}\alpha_k\frac{\alpha_t}{\alpha_k}
                       e^{-a_\star s_{t,k}} \le e^{a_\star/4}
       \sum_{k>t}\alpha_k e^{-(3a_\star/4)s_{t,k}}
 \le C.
\end{align*}
where the last inequality follows from
\eqref{eq:expanded-weight-eq-exp-sum} with $\lambda=3a_\star/4$.
This proves absolute convergence and gives
\[
 \sup_{t\ge0}\norm{W_{\infty,t}}_\infty\le C.
\]
The same calculation bounds every finite partial sum and
therefore proves \eqref{eq:weight-uniform}.

Next, we compare the infinite-horizon weight with the inverse drift.
Since every factor in the product commutes with $A$, for $k>t$,
we have
\[
 \alpha_k A\Phi_{t+1:k-1}
 =\Phi_{t+1:k-1}-\Phi_{t+1:k}.
\]
Summing over $k$ from the first index to $m>t$ gives
\begin{equation}
 A\sum_{k=t+1}^m\alpha_k\Phi_{t+1:k-1}
 =I-\Phi_{t+1:m}.
 \label{eq:expanded-weight-eq-finite-telescope}
\end{equation}
Taking $m\to\infty$ and using
\eqref{eq:expanded-weight-eq-product}, we obtain
\begin{equation}
 \sum_{k>t}\alpha_k\Phi_{t+1:k-1}=A^{-1}.
 \label{eq:expanded-weight-eq-telescope}
\end{equation}
The identity above uses $\alpha_k$ in each summand, while
the definition of the infinite-horizon weight uses $\alpha_t$ in
every summand. Their difference therefore measures the effect of
changing stepsizes.
Subtracting \eqref{eq:expanded-weight-eq-telescope} from
\eqref{eq:expanded-weight-eq-infinite-weight}, we have
\begin{equation}
 W_{\infty,t}-A^{-1}
 =\sum_{k>t}\alpha_k
     \left(\frac{\alpha_t}{\alpha_k}-1\right)\Phi_{t+1:k-1}.
 \label{eq:expanded-weight-eq-infinite-difference}
\end{equation}
Since the stepsizes decrease, the ratio is at least one.
For $x\ge0$, we have $e^x-1=\int_0^x e^y\,dy\le xe^x$.
Taking $x=\log(\alpha_t/\alpha_k)$ and applying
\eqref{eq:expanded-weight-eq-ratio}, we obtain
\[
 0\le\frac{\alpha_t}{\alpha_k}-1
 \le\eta_t(\alpha_t+s_{t,k})
            e^{\eta_t(\alpha_t+s_{t,k})}.
\]
Substituting this inequality into the difference above gives
\begin{align}
 \norm{W_{\infty,t}-A^{-1}}_\infty
 &\le\eta_t e^{\eta_t\alpha_t}
       \sum_{k>t}\alpha_k(\alpha_t+s_{t,k})
                    e^{-(a_\star-\eta_t)s_{t,k}}
 \notag\\
 &\le C\eta_t
       \sum_{k>t}\alpha_k(1+s_{t,k})
                    e^{-(3a_\star/4)s_{t,k}}
 \notag\\
 &\le C\eta_t
 \le C(t+b)^{\beta-1}.
 \label{eq:expanded-weight-eq-interior}
\end{align}
where we used $\alpha_t\le1$, $\eta_t\le a_\star/4$, and
\eqref{eq:expanded-weight-eq-exp-sum}.
This proves the first part of the pointwise estimate
\eqref{eq:weight-pointwise}.

We next bound the tail beyond the finite horizon. Denote
$T_{n,t}=\sum_{u=t+1}^n\alpha_u$.
Factoring the products at time $n$, we have
\begin{align}
 W_{\infty,t}-W_{n,t}
 &=\alpha_t\sum_{k=n+1}^\infty\Phi_{t+1:k-1}
 \notag\\
 &=\alpha_t\Phi_{t+1:n}
          \sum_{k=n+1}^\infty\Phi_{n+1:k-1}
 =\frac{\alpha_t}{\alpha_n}\Phi_{t+1:n}W_{\infty,n}.
 \label{eq:expanded-weight-eq-tail-identity}
\end{align}
The factor $\alpha_t/\alpha_n$ is not uniformly bounded, so
we estimate it together with the contracting product.
Taking $k=n$ in \eqref{eq:expanded-weight-eq-ratio}, we obtain
\[
 \frac{\alpha_t}{\alpha_n}
 \le\exp\left\{\eta_t\left(
          \alpha_t+\sum_{u=t+1}^{n-1}\alpha_u\right)\right\}
 \le e^{a_\star/4}e^{(a_\star/4)T_{n,t}}.
\]
Combining this bound with \eqref{eq:expanded-weight-eq-product}
and the uniform bound for $W_{\infty,n}$, we have
\begin{equation}
 \norm{W_{\infty,t}-W_{n,t}}_\infty
 \le C e^{-(3a_\star/4)T_{n,t}}
 \le C e^{-(a_\star/2)T_{n,t}}.
 \label{eq:expanded-weight-eq-tail}
\end{equation}
By adding and subtracting the infinite-horizon weight, we have
\[
 W_{n,t}-A^{-1}
 =(W_{\infty,t}-A^{-1})-(W_{\infty,t}-W_{n,t}).
\]
Combining \eqref{eq:expanded-weight-eq-interior} and
\eqref{eq:expanded-weight-eq-tail} with the triangle inequality proves
\eqref{eq:weight-pointwise}.

We now average this pointwise bound over the time indices.
For the first term, using $b\ge1$ and $\beta-1<0$, we have
\begin{equation}
 \sum_{t=0}^{n-1}(t+b)^{\beta-1}
 \le\sum_{j=1}^n j^{\beta-1}
 \le1+\int_1^n x^{\beta-1}\,dx
 \le\frac{n^\beta}{\beta}.
 \label{eq:expanded-weight-eq-interior-sum}
\end{equation}
For the second term, let $c=a_\star/2$.
Since $\alpha_u\ge\alpha_n$ for $u\le n$, we have
$T_{n,t}\ge(n-t)\alpha_n$. It follows that
\begin{align}
 \sum_{t=0}^{n-1}e^{-cT_{n,t}}
 &\le\sum_{j=1}^n e^{-cj\alpha_n}
 \le\sum_{j=1}^\infty e^{-cj\alpha_n}
 =\frac1{e^{c\alpha_n}-1}
 \notag\\
 &\le\frac1{c\alpha_n}
 =\frac{(n+b)^\beta}{cc_0}
 \le Cn^\beta.
 \label{eq:expanded-weight-eq-terminal-sum}
\end{align}
The last constant can depend on the fixed shift $b$.
In the last step, we used $n+b\le(1+b)n$ for $n\ge1$.
Summing \eqref{eq:expanded-weight-eq-interior-sum} and
\eqref{eq:expanded-weight-eq-terminal-sum} and dividing by $n$ proves
\eqref{eq:weight-average}.

It remains to prove \eqref{eq:weight-total}. The sum is empty when
$n=1$, so we assume $n\ge2$ below.
For $0\le t\le n-2$, separating the $k=t+1$ term in
\eqref{eq:weights} and factoring $I-\alpha_{t+1}A$ out of the
remaining products gives
\begin{equation}
 W_{n,t}
 =\alpha_t I+
 \frac{\alpha_t}{\alpha_{t+1}}
       (I-\alpha_{t+1}A)W_{n,t+1}.
 \label{eq:expanded-weight-eq-backward}
\end{equation}
Subtracting $W_{n,t+1}$ from both sides, we obtain
\begin{equation}
 W_{n,t}-W_{n,t+1}
 =\alpha_t(I-AW_{n,t+1})
  +\left(\frac{\alpha_t}{\alpha_{t+1}}-1\right)W_{n,t+1}.
 \label{eq:expanded-weight-eq-adjacent}
\end{equation}
For the first term, use the identity
$I-AW_{n,t+1}=A(A^{-1}-W_{n,t+1})$.
Applying \eqref{eq:weight-pointwise} at $t+1$, we have
\begin{align*}
 \alpha_t\norm{I-AW_{n,t+1}}_\infty
 &\le C\alpha_t(t+b+1)^{\beta-1}
  +C\alpha_t
       \exp\left(-c\sum_{u=t+2}^n\alpha_u\right)\\
 &\le\frac{C}{t+b}
  +C\alpha_t
       \exp\left(-c\sum_{u=t+2}^n\alpha_u\right).
\end{align*}
where the second inequality follows from
$(t+b+1)^{\beta-1}\le(t+b)^{\beta-1}$ and
$\alpha_t=c_0(t+b)^{-\beta}$.
For the second term in \eqref{eq:expanded-weight-eq-adjacent}, by
concavity of $x\mapsto x^\beta$, we have
\[
 0\le\frac{\alpha_t}{\alpha_{t+1}}-1
 =\left(1+\frac1{t+b}\right)^\beta-1
 \le\frac{\beta}{t+b}.
\]
Combining the two estimates with \eqref{eq:weight-uniform},
we obtain
\begin{equation}
 \norm{W_{n,t}-W_{n,t+1}}_\infty
 \le\frac{C}{t+b}
  +C\alpha_t\exp\left(-c\sum_{u=t+2}^n\alpha_u\right).
 \label{eq:expanded-weight-eq-adjacent-bound}
\end{equation}

For the sum of the first term in
\eqref{eq:expanded-weight-eq-adjacent-bound}, the harmonic-sum bound
gives
\[
 \sum_{t=0}^{n-2}\frac1{t+b}
 \le C\log(e(n+b))\le C\elln.
\]
We bound the sum of the second term by telescoping. Define
\[
 R_t=\sum_{u=t+2}^n\alpha_u\quad(0\le t\le n-2),
 \qquad R_{n-1}=0.
\]
By definition, $R_t-R_{t+1}=\alpha_{t+2}$.
Since $b\ge1$, we also have
\[
 \frac{\alpha_t}{\alpha_{t+2}}
 =\left(1+\frac2{t+b}\right)^\beta\le3^\beta.
\]
Using $e^x-1\ge x$, we obtain
\[
 e^{-cR_{t+1}}-e^{-cR_t}
 =e^{-cR_t}(e^{c\alpha_{t+2}}-1)
 \ge c\alpha_{t+2}e^{-cR_t}.
\]
Combining the two inequalities above, we have
\begin{align}
 \sum_{t=0}^{n-2}\alpha_t e^{-cR_t}
 &\le\frac{3^\beta}{c}
      \sum_{t=0}^{n-2}
         (e^{-cR_{t+1}}-e^{-cR_t})
 \notag\\
 &=\frac{3^\beta}{c}(1-e^{-cR_0})
 \le\frac{3^\beta}{c}.
 \label{eq:expanded-weight-eq-backward-sum}
\end{align}
Summing \eqref{eq:expanded-weight-eq-adjacent-bound} and
using the preceding bounds proves \eqref{eq:weight-total}.
This completes the proof.
\end{proof}

\subsection{Supporting Lemmas} \label{sec_ubsl}

\begin{lemma}[Theorem~2.3 in \cite{glynn1996liapounov}]\label{sr:poisson}
We use the finite-state, uniformly geometrically ergodic specialization of the
bounded fundamental-kernel conclusion.
Let $K$ have invariant law $\bar\mu$ and satisfy \eqref{eq:mix}, and put
$\Pi F=\bar\mu F$. On bounded centered functions, the potential operator is
\[
 \mathscr RF=(I-K+\Pi)^{-1}F=\sum_{j\ge0}K^jF,
 \qquad (I-K)\mathscr RF=F.
\]
It is linear and bounded. In the present TV convention its bound can be taken as
\[
 \|\mathscr RF\|_{\sup}\le C_P\|F\|_{\sup},\qquad
 C_P=\frac{2C_{\rm mix}}{1-e^{-c_{\rm mix}}}.
\]
The displayed constant follows directly by summing
$\|K^jF\|_{\sup}\le2C_{\rm mix}e^{-c_{\rm mix}j}\|F\|_{\sup}$;
thus the series converges in sup norm. Vector-valued functions are treated coordinatewise.
\end{lemma}

\begin{lemma}[Theorems~3.5 and~4.1 in \cite{pinelis1994}]\label{sr:martingale}
We use their scalar specializations, including equation~(4.2) of the cited paper.
Let $d_j$ be real martingale differences, $S_k=\sum_{j\le k}d_j$,
$d^*=\max_j|d_j|$, and
$s^2=\sum_j\E[d_j^2\mid\mathcal F_{j-1}]$.
\begin{enumerate}[label=(\roman*),nosep]
\item For $p\ge2$ and finite $p$th moments,
\[
 \|\max_k|S_k|\|_{L^p}
 \le C\{p\|d^*\|_{L^p}+\sqrt p\|s\|_{L^p}\},
\]
with an absolute constant $C$ independent of $p$ and the number of increments.
\item If $|d_j|\le a_j$ almost surely for deterministic $a_j$ and $V=\sum_ja_j^2>0$,
then for $x\ge0$,
\[
 \Pr\{\max_k|S_k|\ge x\}\le2e^{-x^2/(2V)}.
\]
\end{enumerate}
\end{lemma}

\begin{lemma}[Theorem~9.2 in \cite{bhatia1997}]\label{sr:sylvester}
We specialize the cited result to symmetric positive-definite matrices, using
the identification $A=U$, $B=-V$ in that paper. The Frobenius bound is an
immediate consequence of the integral representation; see also Section~10
of the cited paper.
For real symmetric positive-definite $U,V$ and any real matrix $H$ of compatible size,
the equation $UE+EV=H$ has the unique solution
\[
 E=\int_0^\infty e^{-tU}H e^{-tV}\,dt,\qquad
 \|E\|_F\le\frac{\|H\|_F}{\lambda_{\min}(U)+\lambda_{\min}(V)}.
\]
The integrand's Frobenius norm is at most
$e^{-t(\lambda_{\min}(U)+\lambda_{\min}(V))}\|H\|_F$.
No commutativity of $U,V,H$ is assumed.
\end{lemma}

\begin{lemma}[Proposition~4.7 in \cite{levin2017}]\label{sr:coupling}
Part~(i) is the maximal coupling assertion; see also Remark~4.8 of the cited book.
Part~(ii) is a direct consequence of the definition of total variation.
For probability laws $\nu,\nu'$ on a finite space:
\begin{enumerate}[label=(\roman*),nosep]
\item There is a coupling $(Z,Z')$ with these marginals such that
$\Pr(Z\ne Z')=\|\nu-\nu'\|_{\TV}$.
\item For any Markov kernel $T$, $\|\nu T-\nu'T\|_{\TV}\le\|\nu-\nu'\|_{\TV}$.
\end{enumerate}
The second assertion follows by integrating $T(\cdot,E)\in[0,1]$ against $\nu-\nu'$.
Both assertions apply to finite trajectory-segment spaces as well as to individual states.
\end{lemma}

\begin{lemma}[Equation~(3.5) in \cite{raic2019}]\label{sr:raic}
We use the bound in Section~3 of arXiv version~2.
Let $\xi_j\in\R^d$ be independent centered vectors satisfying
$\sum_j\E\xi_j\xi_j^\top=I_d$, and let $Z\sim\N(0,I_d)$.
Writing $k_d=11.1+0.83\log d$, the stated bound implies
\[
 W_1\left(\sum_j\xi_j,Z\right)
 \le\sum_j\E\left[\|\xi_j\|_2^2\min\{4.5,k_d\|\xi_j\|_2\}\right]
 \le k_d\sum_j\E\|\xi_j\|_2^3.
\]
The last inequality is used when the third absolute moments are finite.
No identical-distribution assumption or smallness condition on their sum is required.
\end{lemma}

\begin{lemma}[Theorem~5.10 in \cite{villani2009}]\label{sr:wasserstein}
We specialize Kantorovich duality in Chapter~5 to Euclidean distance as cost.
The map and mean bounds below are immediate consequences.
For probability laws with finite first moments,
\[
 W_1(U,V)=\inf_{\pi\in\Gamma(\mathcal L(U),\mathcal L(V))}
                  \int\|u-v\|_2\,d\pi
 =\sup_{\operatorname{Lip}_2(h)\le1}|\E h(U)-\E h(V)|.
\]
Consequently, an $L$-Lipschitz map $T$ satisfies
$W_1(TU,TV)\le L W_1(U,V)$, and
$W_1(U,V)\ge\|\E U-\E V\|_2$.
These include coordinate projection and deterministic linear maps, with constants
$1$ and $\|T\|_{\op}$ respectively.
\end{lemma}

\begin{lemma}[Poisson equation]\label{lem:poisson}
Suppose \eqref{eq:mix} holds. Let $F_\theta$ be a family of bounded
vector-valued functions with $\bar\mu F_\theta=0$. Write
$\norm F_{\sup}=\sup_y\norm{F(y)}_\infty$. Then
\[
 h_\theta=\sum_{j=0}^\infty K^jF_\theta,\qquad v_\theta=Kh_\theta
\]
are well defined and satisfy $h_\theta-Kh_\theta=F_\theta$.
Furthermore,
\begin{align}
 \norm{h_\theta}_{\sup}+\norm{v_\theta}_{\sup}
 &\le C\norm{F_\theta}_{\sup},\label{eq:poisson-size}\\
 \norm{h_\theta-h_{\theta'}}_{\sup}
 +\norm{v_\theta-v_{\theta'}}_{\sup}
 &\le C\norm{F_\theta-F_{\theta'}}_{\sup}.
 \label{eq:poisson-lipschitz}
\end{align}
If $\theta_t$ is $\cF_{t-1}$-measurable, the family is jointly
measurable, and $\E\norm{F_{\theta_t}}_{\sup}<\infty$, then for $t\ge1$,
\[
 h_{\theta_t}(Y_t)-v_{\theta_t}(Y_{t-1})
\]
is a martingale difference.
\end{lemma}

\begin{proof}
Apply the bounded linear potential operator in Lemma~\ref{sr:poisson} to each
centered $F_\theta$. Its norm bound and linearity give both the size estimate and the
same estimate applied to $F_\theta-F_{\theta'}$; $K$ is a sup-norm contraction.
For a predictable parameter, integrability follows from \eqref{eq:poisson-size}, and
the Markov property gives
\[
 \E[h_{\theta_t}(Y_t)\mid\mathcal F_{t-1}]
 =(Kh_{\theta_t})(Y_{t-1})=v_{\theta_t}(Y_{t-1}).
\]
Thus the asserted difference is a martingale difference. The decomposition used in
Section~\ref{app:expansion} is the exact identity
\[
 F_{\theta_t}(Y_t)
 =\{h_{\theta_t}(Y_t)-v_{\theta_t}(Y_{t-1})\}
   +v_{\theta_t}(Y_{t-1})-v_{\theta_t}(Y_t).
\]
The dependence on $\theta_t$ is retained in the last two terms when summing over time.
\end{proof}

\begin{lemma}[Martingale moment bounds]\label{lem:martingale}
Let $(D_j)_{j=1}^N$ be vector-valued martingale differences in
$\R^d$ with finite $p$th moments, where $p\ge2$. Then
\begin{equation}
 \lp{\sum_{j=1}^ND_j}
 \le C_dp\left(\sum_{j=1}^N\lp{D_j}^{\,2}\right)^{1/2}.
 \label{eq:martingale-moments}
\end{equation}
If $\norm{D_j}_\infty\le a_j$ almost surely for deterministic $a_j$,
then
\begin{equation}
 \norm{\max_{0\le k\le N}
               \norm{\sum_{j=1}^kD_j}_\infty}_{L^p}
 \le C_d\sqrt{p\sum_{j=1}^Na_j^2}.
 \label{eq:martingale-maximum}
\end{equation}
\end{lemma}

\begin{proof}
Apply Lemma~\ref{sr:martingale}(i) coordinatewise.
For scalar martingale differences $d_j$, conditional Jensen and Minkowski give
\[
 \bigl\|\max_j|d_j|\bigr\|_{L^p}
 \le\left(\sum_j\|d_j\|_{L^p}^2\right)^{1/2},\qquad
 \left\|\left(\sum_j\E[d_j^2\mid\mathcal F_{j-1}]\right)^{1/2}\right\|_{L^p}
 \le\left(\sum_j\|d_j\|_{L^p}^2\right)^{1/2}.
\]
Since $\|Z\|_{L^p(\ell_\infty)}\le d^{1/p}\max_i\|Z_i\|_{L^p}$,
the first assertion follows, with the required factor $C_dp$.
For the second assertion, part~(ii) of Lemma~\ref{sr:martingale} and a coordinate union bound yield,
with $V=\sum_ja_j^2>0$,
\[
 \Pr\left\{\max_{k\le N}\left\|\sum_{j\le k}D_j\right\|_\infty>x\right\}
 \le2d e^{-x^2/(2V)}.
\]
Integrating this tail gives an $L^p$ bound by
$(2d)^{1/p}(2V)^{1/2}\Gamma(1+p/2)^{1/p}\le C_d\sqrt{pV}$.
The case $V=0$ is immediate.
\end{proof}

\begin{lemma}[Gaussian covariance comparison]\label{lem:gaussian}
For positive-definite matrices $V_1,V_2$,
\begin{equation}
 W_1(\N(0,V_1),\N(0,V_2))
 \le
 \frac{\norm{V_1-V_2}_F}
 {\sqrt{\lambda_{\min}(V_1)}+\sqrt{\lambda_{\min}(V_2)}}.
 \label{eq:gaussian-comparison}
\end{equation}
\end{lemma}

\begin{proof}
Let $U=V_1^{1/2}$, $V=V_2^{1/2}$, and $E=U-V$.
Then $UE+EV=V_1-V_2$, without a commutativity assumption.
Lemma~\ref{sr:sylvester} gives
\[
 \|E\|_F\le\frac{\|V_1-V_2\|_F}
 {\sqrt{\lambda_{\min}(V_1)}+\sqrt{\lambda_{\min}(V_2)}}.
\]
Couple the Gaussians as $UZ,VZ$ with $Z\sim\N(0,I_d)$.
Then $W_1(UZ,VZ)\le\E\|EZ\|_2\le\|E\|_F$, proving the claim.
\end{proof}

\begin{lemma}[Coupling separated trajectory intervals]\label{lem:coupling}
Suppose \eqref{eq:mix} holds. Let $I_1,\ldots,I_N$ be ordered
intervals of time indices. Suppose at least $m$ indices precede
$I_1$ and at least $m$ indices lie between successive intervals.
There are independent stationary trajectory segments
$(Y'_t:t\in I_j)$, each of the corresponding length, that can be
coupled with the original chain so that
\begin{equation}
 \Prob\{\text{some retained segment differs from its copy}\}
 \le CN e^{-c_{\rm mix}m}.
 \label{eq:block-coupling}
\end{equation}
\end{lemma}

\begin{proof}
Write $I_j=\{a_j,\ldots,b_j\}$, let $\mathscr Y_j$ be its trajectory segment,
and let $\nu_j$ be its stationary segment law.
For $j\ge2$, conditional on the preceding retained segments, the law at $a_j$ is
$K^{a_j-b_{j-1}}(Y_{b_{j-1}},\cdot)$.
The gap condition and \eqref{eq:mix} bound its total variation distance from $\bar\mu$
by $Ce^{-c_{\rm mix}m}$. The initial discarded interval gives the same bound for $j=1$.
Applying the segment transition kernel preserves this bound, by
Lemma~\ref{sr:coupling}(ii).
Telescoping the joint law against the product of the $\nu_j$ therefore gives
\[
 \left\|\mathcal L(\mathscr Y_1,\ldots,\mathscr Y_N)-\bigotimes_{j=1}^N\nu_j\right\|_{\TV}
 \le CN e^{-c_{\rm mix}m}.
\]
The maximal coupling in Lemma~\ref{sr:coupling}(i) proves the desired bound.
Its product marginal consists of independent stationary segments.
Sampling the copies conditionally on the retained original segments realizes this coupling
on an extension of the original probability space; all retained and discarded original observations
keep their original joint law.
\end{proof}

\begin{lemma}[Independent-sum Gaussian approximation]\label{lem:independent}
Let $\xi_1,\ldots,\xi_N$ be independent centered vectors in $\R^d$,
with $\sum_j\E\xi_j\xi_j^\top=I_d$. Let
$\rho_3=\sum_j\E\norm{\xi_j}_2^3$ and $Z\sim\N(0,I_d)$.
If $0<\rho_3\le1/2$, then
\begin{equation}
 W_1\left(\sum_j\xi_j,Z\right)
 \le C_d\rho_3(1+|\log\rho_3|).
 \label{eq:independent-normal}
\end{equation}
\end{lemma}

\begin{proof}
Lemma~\ref{sr:raic}, with $k_d=11.1+0.83\log d$, gives the stronger estimate
\[
 W_1\left(\sum_j\xi_j,Z\right)\le k_d\sum_j\E\|\xi_j\|_2^3=k_d\rho_3.
\]
This implies \eqref{eq:independent-normal}. Independence, centering, and identity total covariance
are exactly the hypotheses of that lemma; identical distributions are not required.
\end{proof}

\section{Proof of Theorem \ref{lb:theorem}}\label{lb:main-proof}

In this section, we establish the lower bound. Proposition \ref{lb:prop-bias} and Proposition \ref{lb:prop-cov} are discussed in the main body. In the proofs below, vector norms without a subscript are Euclidean norms, and matrix norms without a subscript are operator norms.

\theoremstyle{plain}
\newtheorem*{lowerrestatement}{Theorem \ref{lb:theorem}}
\theoremstyle{definition}
\newtheorem*{biasrestatement}{Proposition \ref{lb:prop-bias}}
\newtheorem*{covrestatement}{Proposition \ref{lb:prop-cov}}

\lbthm*

\begin{proof}[Proof of Theorem~\ref{lb:theorem}]
We first verify the assumptions for the fixed problem.
Solving $\mu^\top P=\mu^\top$ gives
$\mu=(3/5,2/5)^\top$. Let $\Pi=\mathbf{1}\mu^\top$.
By direct multiplication,
\[
 P=\Pi+\frac12(I-\Pi),\qquad
 \Pi^2=\Pi,\qquad \Pi(I-\Pi)=(I-\Pi)\Pi=0.
\]
Therefore,
\begin{equation}\label{lb:mixing-explicit}
 P^j=\Pi+2^{-j}(I-\Pi),\qquad
 \sup_i\norm{P^j(i,\cdot)-\mu}_{\TV}\le2^{-j}.
\end{equation}
With the unique action suppressed, we identify the transition
triple $Y_t$ with the pair $(s_t,s_{t+1})$.
Conditional on $Y_0=(i,j)$, its law after $h\ge1$ steps is obtained
by starting the state chain at $j$, applying $h-1$ transitions, and
then drawing one further transition.
Applying a Markov kernel cannot increase total variation distance.
Thus its distance from the stationary pair law is at most
$2^{-(h-1)}$. Increasing the constant also covers $h=0$.
This proves geometric mixing for the transition chain.
Exploration holds since $\min_i\mu_i=2/5$.

The Bellman system and its inverse are
\[
 I-\gamma P=\frac1{25}\begin{pmatrix}9&-4\\-6&11\end{pmatrix},
 \qquad
 (I-\gamma P)^{-1}
 =\frac13\begin{pmatrix}11&4\\6&9\end{pmatrix}.
\]
Multiplying the inverse by $r$ gives $Q^\star=(4/3,3)^\top$.
Also,
\[
 A=D(I-\gamma P)
   =\frac1{125}\begin{pmatrix}27&-12\\-12&22\end{pmatrix},
 \qquad
 \lambda_\pm(A)=\frac{49\pm\sqrt{601}}{250}>0.
\]
Since there is one action, the mean update at $Q^\star+\delta$ is
exactly $-A\delta$. Hence the quadratic remainder assumption holds with
zero remainder.

To calculate the fixed-point noise covariance, write
$\xi_{ij}=r_i+\gamma Q_j^\star-Q_i^\star$. The four values are
\begin{equation}\label{lb:xi}
 (\xi_{ij})_{i,j=1}^2
 =\begin{pmatrix}-4/15&16/15\\-14/15&2/5\end{pmatrix}.
\end{equation}
Their conditional means are zero in both rows.
Only the coordinate corresponding to $s_t$ is updated, so the
off-diagonal entries of $S$ vanish. Its diagonal entries are
\begin{align*}
 S_{11}
 &=\frac35\left\{\frac45\frac{16}{225}
                  +\frac15\frac{256}{225}\right\}
 =\frac{64}{375},\\
 S_{22}
 &=\frac25\left\{\frac3{10}\frac{196}{225}
                  +\frac7{10}\frac4{25}\right\}
 =\frac{56}{375}.
\end{align*}

Thus $S\succ0$, and invertibility of $A$ gives
$\Sigma_\infty\succ0$.
For reference, the inverse and the limiting covariance are
\begin{equation}\label{lb:inverse-covariance}
 A^{-1}=\begin{pmatrix}55/9&10/3\\10/3&15/2\end{pmatrix},
 \qquad
 \Sigma_\infty=\frac1{243}
       \begin{pmatrix}1952&1752\\1752&2502\end{pmatrix}.
\end{equation}
The initialization $Q_0=0$ belongs to $[0,5]^2$.
Since $\alpha_t\le1$, each updated coordinate is a convex
combination of its old value and a value in this interval.
Thus the iterates remain bounded and all the first moments used
below are finite.

Suppose first that $1/2<\beta\le3/4$.
Let $w=v/\norm v_2$, which is a fixed unit vector.
For every coupling of $X_n$ and $Z_\infty$, we have
\[
 \E\norm{X_n-Z_\infty}_2
 \ge\left|\E[w^\top(X_n-Z_\infty)]\right|.
\]
Taking the infimum over couplings and using
$\E Z_\infty=0$, Proposition~\ref{lb:prop-bias} gives
\begin{align}
 W_1(X_n,Z_\infty)
 &\ge |w^\top\E X_n|\notag\\
 &=\frac{c_0\norm v_2}{1-\beta}n^{1/2-\beta}
                 +o(n^{1/2-\beta})\notag\\
 &\ge \frac{c_0\norm v_2}{2(1-\beta)}n^{1/2-\beta}
 \label{lb:mean-lower-proof}
\end{align}
for all sufficiently large $n$.
The coefficient is strictly positive because $v\ne0$.
This proves the first regime, including $\beta=3/4$.

Now suppose that $3/4<\beta<1$ and put
$\delta_n=n^{\beta-1}$.
By Proposition~\ref{lb:prop-cov}, $C_n\to\Sigma_\infty$.
Since the latter is positive definite, the eigenvalues of $C_n$
are bounded below by a fixed positive constant for all sufficiently
large $n$. We can therefore use the finite-covariance approximation
from Propositions~\ref{mb:expansion} and~\ref{mb:finite}.
The identity $X_n=M_n+R_n$ gives a coupling on the original space,
so
\begin{align}
 W_1(X_n,Z_n)
 &\le \E\norm{R_n}_2+W_1(M_n,Z_n)\notag\\
 &\le C\{n^{-\beta/2}+n^{1/2-\beta}
             +\elln n^{-1/2}+\elln n^{-1/4}\}.
 \label{lb:finite-transfer}
\end{align}
We compare each term with $\delta_n$. The four ratios are
\begin{equation}\label{lb:error-ratios}
 n^{1-3\beta/2},\qquad
 n^{3/2-2\beta},\qquad
 \elln n^{1/2-\beta},\qquad
 \elln n^{3/4-\beta}.
\end{equation}
All four converge to zero in the stated range.
For the first ratio it is enough that $\beta>2/3$.
The second and fourth require $\beta>3/4$, while the third requires
only $\beta>1/2$. Since $\elln=O(\log n)$ for fixed $b$, it does
not change any of these strict exponent comparisons.
Consequently,
\begin{equation}\label{lb:negligible-transfer}
 W_1(X_n,Z_n)=o(\delta_n).
\end{equation}

Fix a coordinate $j$ and write
$\sigma_j^2=(\Sigma_\infty)_{jj}>0$ and $g_j=G_{jj}>0$.
The covariance expansion gives
\[
 (C_n)_{jj}=\sigma_j^2+g_j\delta_n+o(\delta_n).
\]
Let $\sigma_{n,j}=\sqrt{(C_n)_{jj}}$.
Rationalizing the difference of square roots, we obtain
\begin{equation}\label{lb:std-expansion}
 \sigma_{n,j}-\sigma_j
 =\frac{(C_n)_{jj}-\sigma_j^2}{\sigma_{n,j}+\sigma_j}
 =\frac{g_j}{2\sigma_j}\delta_n+o(\delta_n).
\end{equation}
By Lemma~\ref{lb:lem-gaussian}, projection onto coordinate $j$
and the exact scalar Gaussian formula give
\begin{align}
 W_1(Z_n,Z_\infty) \ge \sqrt{\frac2\pi}\,|\sigma_{n,j}-\sigma_j| =\sqrt{\frac2\pi}\frac{g_j}{2\sigma_j}\delta_n
                        +o(\delta_n).
 \label{lb:gaussian-lower-proof}
\end{align}
The leading coefficient is strictly positive.
Applying the reverse triangle inequality and
\eqref{lb:negligible-transfer}, we conclude that
\begin{align*}
 W_1(X_n,Z_\infty) \ge W_1(Z_n,Z_\infty)-W_1(X_n,Z_n) \ge c\delta_n
\end{align*}
for all sufficiently large $n$.
At $\beta=3/4$, the last ratio in \eqref{lb:error-ratios} does
not tend to zero under the available bound. That endpoint has
already been covered by \eqref{lb:mean-lower-proof}.
Increasing the threshold $n_0$ to include all the preceding
requirements completes the proof.
\end{proof}

\subsection{Proof of Proposition \ref{lb:prop-bias}}\label{lb:mean-proof}

\llemmaa*

\begin{proof}
We first derive a closed recursion for the joint first moments.
For $i\in\{1,2\}$, define the two-dimensional vector
$z_t^{(i)}=\E[\Delta_t\mathbf{1}\{s_t=i\}]$, and let
\begin{equation}\label{lb:z-definition}
 z_t=\begin{pmatrix}z_t^{(1)}\\z_t^{(2)}\end{pmatrix}
 =\begin{pmatrix}
 \E[\Delta_{t,1}\mathbf{1}\{s_t=1\}]\\
 \E[\Delta_{t,2}\mathbf{1}\{s_t=1\}]\\
 \E[\Delta_{t,1}\mathbf{1}\{s_t=2\}]\\
 \E[\Delta_{t,2}\mathbf{1}\{s_t=2\}]
 \end{pmatrix},\qquad
 L_0=\begin{pmatrix}1&0&1&0\\0&1&0&1\end{pmatrix}.
\end{equation}
By definition, $L_0z_t=\E\Delta_t$.
The iterate and the current state need not be independent.
We keep their joint moments when taking expectations.

Conditional on $s_t=i,s_{t+1}=j$, the error update is
\begin{equation}\label{lb:conditional-update}
 \Delta_{t+1}
 =\{I+\alpha_t \mathbf u_i(\gamma \mathbf u_j-\mathbf u_i)^\top\}\Delta_t
       +\alpha_t \mathbf u_i\xi_{ij},
\end{equation}
where $\xi_{ij}$ is given in \eqref{lb:xi}.
Given the current state and the past, the next state has law
$P(s_t,\cdot)$. Therefore,
\[
 \E[\Delta_t\mathbf{1}\{s_t=i,s_{t+1}=j\}]
 =P_{ij}z_t^{(i)}.
\]
The state law is stationary at every time, so
$\Prob(s_t=i,s_{t+1}=j)=\mu_iP_{ij}$.
Multiplying \eqref{lb:conditional-update} by the indicator of the
next state and taking expectations gives
\begin{equation}\label{lb:block-moment-recursion}
 z_{t+1}^{(j)}
 =\sum_{i=1}^2P_{ij}
       \{I+\alpha_t \mathbf u_i(\gamma \mathbf u_j-\mathbf u_i)^\top\}z_t^{(i)}
   +\alpha_t\sum_{i=1}^2\mu_iP_{ij}\mathbf u_i\xi_{ij}.
\end{equation}
The forcing is constant in time because of stationary
initialization.

Writing $z_{t,k}$ for the $k$th coordinate of $z_t$, the four rows
of \eqref{lb:block-moment-recursion} are
\begin{align}
 z_{t+1,1}
 &=\frac45z_{t,1}+\frac3{10}z_{t,3}
       -\alpha_t\frac4{25}z_{t,1}-\alpha_t\frac{16}{125},\notag\\
 z_{t+1,2}
 &=\frac45z_{t,2}+\frac3{10}z_{t,4}
       +\alpha_t\left(\frac6{25}z_{t,3}-\frac3{10}z_{t,4}\right)
       -\alpha_t\frac{14}{125},\notag\\
 z_{t+1,3}
 &=\frac15z_{t,1}+\frac7{10}z_{t,3}
       +\alpha_t\left(-\frac15z_{t,1}+\frac4{25}z_{t,2}\right)
       +\alpha_t\frac{16}{125},\notag\\
 z_{t+1,4}
 &=\frac15z_{t,2}+\frac7{10}z_{t,4}
       -\alpha_t\frac7{50}z_{t,4}+\alpha_t\frac{14}{125}.
 \label{lb:four-moment-rows}
\end{align}
For example, the forcing in the first row is
$\mu_1P_{11}\xi_{11}=-16/125$.
For the second row it is
$\mu_2P_{21}\xi_{21}=-14/125$.
The other two forcing entries have the opposite signs.

Thus, the recursion can be written as
\begin{equation}\label{lb:z-recursion}
 z_{t+1}=(T_0+\alpha_tT_1)z_t+\alpha_t g,
 \qquad g=\frac1{125}\begin{pmatrix}-16\\-14\\16\\14\end{pmatrix},
\end{equation}
where
\begin{equation}\label{lb:T-matrices}
 T_0=\begin{pmatrix}
 4/5&0&3/10&0\\0&4/5&0&3/10\\
 1/5&0&7/10&0\\0&1/5&0&7/10
 \end{pmatrix},\qquad
 T_1=\begin{pmatrix}
 -4/25&0&0&0\\0&0&6/25&-3/10\\
 -1/5&4/25&0&0\\0&0&0&-7/50
 \end{pmatrix}.
\end{equation}

We next change coordinates so that the overall mean is the first
block. Define
\begin{equation}\label{lb:coordinate-matrices}
 J_0=\begin{pmatrix}3/5&0\\0&3/5\\2/5&0\\0&2/5\end{pmatrix},\qquad
 F_0=\begin{pmatrix}1&0\\0&1\\-1&0\\0&-1\end{pmatrix},\qquad
 M_0=[J_0\ F_0].
\end{equation}
The inverse is
\[
 M_0^{-1}=\begin{pmatrix}
 1&0&1&0\\0&1&0&1\\
 2/5&0&-3/5&0\\0&2/5&0&-3/5
 \end{pmatrix}.
\]
Write $z=M_0(u,f)^\top$.
Then
\begin{equation}\label{lb:uf-meaning}
 u=z^{(1)}+z^{(2)}=L_0z,\qquad
 f=\frac25z^{(1)}-\frac35z^{(2)},\qquad
 z^{(1)}=\frac35u+f,\quad z^{(2)}=\frac25u-f.
\end{equation}
In particular, $u_t=\E\Delta_t$.
The second block is zero when each joint moment equals the
stationary state probability times the overall mean.

Multiplying the matrices in \eqref{lb:T-matrices} by this change
of coordinates gives
\begin{equation}\label{lb:transformed-recursion}
 M_0^{-1}(T_0+aT_1)M_0
 =\begin{pmatrix}I-aA&aB_0\\aC_0&\frac12I+aE_0\end{pmatrix},
 \qquad
 M_0^{-1}g=\begin{pmatrix}0\\g_f\end{pmatrix},
\end{equation}
where
\begin{align}
 B_0&=\frac1{25}\begin{pmatrix}-9&4\\-6&11\end{pmatrix},&
 C_0&=\frac1{625}\begin{pmatrix}21&-36\\24&-9\end{pmatrix},\notag\\
 E_0&=\frac1{250}\begin{pmatrix}14&-24\\-24&9\end{pmatrix},&
 g_f&=\frac1{125}\begin{pmatrix}-16\\-14\end{pmatrix}.
 \label{lb:transformed-matrices}
\end{align}
The zero in the forcing is a two-dimensional vector.
Thus the mean has no direct additive forcing, but it is coupled
to the second block through $aB_0f$.

For a constant stepsize $a>0$, consider an equilibrium $z(a)$ of
\eqref{lb:z-recursion}. In the transformed coordinates, its equations
are
\begin{align*}
 u&=(I-aA)u+aB_0f,\\
 f&=aC_0u+(\tfrac12I+aE_0)f+ag_f.
\end{align*}
Dividing the first equation by $a$ gives $Au=B_0f$.
Substituting $u=A^{-1}B_0f$ into the second gives
\begin{equation}\label{lb:equilibrium-system}
 \mathcal H(a)f=ag_f,\qquad
 \mathcal H(a)=\frac12I-aN_0,\qquad
 N_0=E_0+C_0A^{-1}B_0
     =\frac1{375}\begin{pmatrix}48&-88\\-63&28\end{pmatrix}.
\end{equation}
Since $\mathcal H(0)=I/2$, this matrix is invertible in a
neighborhood of zero. The equilibrium is therefore uniquely given by
\begin{equation}\label{lb:equilibrium-formula}
 f(a)=a\mathcal H(a)^{-1}g_f,\qquad
 u(a)=A^{-1}B_0f(a),\qquad
 z(a)=M_0\begin{pmatrix}u(a)\\f(a)\end{pmatrix}.
\end{equation}
The entries are analytic near zero and all three vectors vanish
at zero. These properties allow us to control the change in the
equilibrium as the stepsize decreases.

We give the exact mean to verify the first-order coefficient.
For the two-by-two system,
\[
 \det\mathcal H(a)=\frac14-\frac{38}{375}a-\frac{56}{1875}a^2,
 \qquad
 7500\det\mathcal H(a)=1875-760a-224a^2.
\]
Using the inverse formula for a two-by-two matrix gives
\[
 f(a)=\frac{a}{3125\det\mathcal H(a)}
           \begin{pmatrix}112a-200\\112a-175\end{pmatrix}.
\]
Multiplying by $A^{-1}B_0$ and collecting the numerator terms,
we obtain
\begin{equation}\label{lb:rational-mean}
 u_1(a)=\frac{8a(56a-155)}{3(224a^2+760a-1875)},\qquad
 u_2(a)=\frac{2a(85-112a)}{224a^2+760a-1875}.
\end{equation}
Alternatively, expanding the inverse in
\eqref{lb:equilibrium-formula} gives
\begin{align}
 f(a)&=2ag_f+O(a^2),\notag\\
 u(a)&=2aA^{-1}B_0g_f+O(a^2)
       =a\begin{pmatrix}248/1125\\-34/375\end{pmatrix}+O(a^2).
 \label{lb:equilibrium-expansion}
\end{align}
For the last equality, one can first compute
\[
 B_0g_f=\frac1{3125}\begin{pmatrix}88\\-58\end{pmatrix}
\]
and then use the inverse in \eqref{lb:inverse-covariance}.
This verifies that the vector $v$ in the proposition is nonzero.

It remains to compare the nonconstant-stepsize recursion with these
equilibria. Lemma~\ref{lb:lem-stability}, applied to
\eqref{lb:transformed-recursion}, gives a single norm in which all
sufficiently small stepsize matrices contract. Its product bound is
\begin{equation}\label{lb:moment-product}
 \norm{(T_0+\alpha_tT_1)\cdots(T_0+\alpha_kT_1)}
 \le C\exp\left(-c\sum_{u=k}^t\alpha_u\right),\qquad k\le t.
\end{equation}
The finitely many earlier factors are included in the constant.
The proof of that lemma gives the quadratic estimates for both
blocks and verifies the bound for the full varying product.

Choose a fixed $t_0$ so large that $z(\alpha_t)$ is defined in the
analytic neighborhood above for every $t\ge t_0$.
Let $\chi_t=z_t-z(\alpha_t)$.
Subtracting the equilibrium equation from \eqref{lb:z-recursion},
we have
\begin{equation}\label{lb:tracking-recursion}
 \chi_{t+1}=(T_0+\alpha_tT_1)\chi_t
                +z(\alpha_t)-z(\alpha_{t+1}).
\end{equation}
The derivative of $z(a)$ is bounded on a sufficiently small closed
interval. Hence
\begin{align}
 \norm{z(\alpha_t)-z(\alpha_{t+1})} \le C|\alpha_t-\alpha_{t+1}| \le C\frac{\alpha_t}{t+b}.
 \label{lb:equilibrium-movement}
\end{align}
The last inequality follows from the mean value theorem applied
to $c_0x^{-\beta}$.

Unrolling \eqref{lb:tracking-recursion} and using
\eqref{lb:moment-product} gives
\begin{align}
 \norm{\chi_t} \le C\exp\left(-c\sum_{u=t_0}^{t-1}\alpha_u\right)\norm{\chi_{t_0}}
 +C\sum_{k=t_0}^{t-1}
      \exp\left(-c\sum_{u=k+1}^{t-1}\alpha_u\right)
                   \frac{\alpha_k}{k+b}.
 \label{lb:tracking-convolution}
\end{align}
Lemma~\ref{lb:lem-tracking} proves that the sum on the right is
$O((t+b)^{-1})$. It does so by separating the early indices, for
which the remaining product is exponentially small, from the later
indices, for which a geometric sum is uniformly bounded.
The initial term is also exponentially small.
Consequently,
\begin{equation}\label{lb:tracking-result}
 \norm{z_t-z(\alpha_t)}=O((t+b)^{-1}).
\end{equation}
Applying $L_0$ and combining this estimate with
\eqref{lb:equilibrium-expansion}, we obtain
\[
 \E\Delta_t=\alpha_t v+O(\alpha_t^2+(t+b)^{-1}),
\]
which proves \eqref{lb:mean-iterate}.

Finally, sum this expansion over $t=1,\ldots,n$.
By Lemma~\ref{lb:lem-sums},
\begin{align*}
 \sum_{t=1}^n\alpha_t
 &=\frac{c_0}{1-\beta}n^{1-\beta}+O(1),\\
 \sum_{t=1}^n\alpha_t^2&=O(1),\qquad
 \sum_{t=1}^n(t+b)^{-1}=O(\log(en)).
\end{align*}
After division by $\sqrt n$, we have
\[
 \E X_n=\frac{c_0}{1-\beta}v\,n^{1/2-\beta}
                         +O\left(\frac{\log(en)}{\sqrt n}\right).
\]
The ratio of the last term to $n^{1/2-\beta}$ is bounded by a
constant times $\log(en)n^{\beta-1}$, which tends to zero.
This proves \eqref{lb:mean-average}.

The only dependence on the deterministic initialization occurs
in the fixed initial error in \eqref{lb:tracking-convolution}.
For stationary states and deterministic $\Delta_0$, one has
$z_0=J_0\Delta_0$. Every fixed bounded initialization therefore
gives the same leading coefficient.
This completes the proof.
\end{proof}

\subsection{Proof of Proposition \ref{lb:prop-cov}}\label{lb:cov-proof}

\llemmab*

\begin{proof}
We first expand the infinite-horizon weight and then calculate the
effect of truncating it at time $n$.
All limits below are taken with $c_0,b,\beta$ fixed.
Let
\[
 W_{\infty,t}=\alpha_t\sum_{k=t+1}^\infty\Phi_{t+1:k-1},
 \qquad R=A^{-1},\qquad \tau=t+b,\qquad a=\alpha_t.
\]
Lemma~\ref{lb:lem-weights} proves convergence of this series and the
uniform bounds used below. It also proves the telescoping identity
\[
 \sum_{k=t+1}^\infty\alpha_k\Phi_{t+1:k-1}=A^{-1}.
\]
Subtracting this identity from the definition of the infinite-horizon
weight gives
\begin{equation}\label{lb:infinite-difference}
 W_{\infty,t}-A^{-1}
 =\sum_{j=1}^\infty(\alpha_t-\alpha_{t+j})
                         \Phi_{t+1:t+j-1}.
\end{equation}
We will prove
\begin{equation}\label{lb:infinite-asymptotic}
 W_{\infty,t}
 =A^{-1}+\frac\beta{c_0}(t+b)^{\beta-1}A^{-2}
                           +o((t+b)^{\beta-1}).
\end{equation}

Fix $L>0$ and first consider the summands with
$1\le j\le\lfloor L/a\rfloor$.
Since $\tau a=c_0\tau^{1-\beta}\to\infty$, we have
\[
 \max_{j\le L/a}\frac j\tau\le\frac L{\tau a}\longrightarrow0.
\]
The Taylor expansion of $(1+x)^{-\beta}$ at zero therefore gives,
uniformly over these indices,
\begin{align}
 \alpha_t-\alpha_{t+j} =a\{1-(1+j/\tau)^{-\beta}\} =\frac{\beta a j}{\tau}
           \left(1+O\left(\frac j\tau\right)\right).
 \label{lb:local-step-difference}
\end{align}
The error is uniform for each fixed $L$ because the second derivative
of $(1+x)^{-\beta}$ is bounded on a fixed neighborhood of zero.

We also need a uniform approximation to the matrix product.
For $1\le h\le j-1$, the mean value theorem gives
\[
 |\alpha_{t+h}-a|\le C\frac{ah}{\tau}.
\]
Summing this inequality and using $j\le L/a$ gives
\begin{equation}\label{lb:local-clock-error}
 \left|\sum_{h=1}^{j-1}\alpha_{t+h}-a(j-1)\right|
 \le C\frac{aj^2}{\tau}
 \le C\frac{L^2}{\tau a}\longrightarrow0.
\end{equation}
Since the stepsizes decrease,
\begin{equation}\label{lb:local-square-error}
 \sum_{h=1}^{j-1}\alpha_{t+h}^2\le ja^2\le La\longrightarrow0.
\end{equation}

The matrix $A$ is symmetric and positive definite.
For sufficiently large $t$, the convergent matrix logarithm series
gives
\[
 \log(I-\alpha_u A)=-\alpha_u A+\mathcal E_u,\qquad
 \norm{\mathcal E_u}\le C\alpha_u^2,\qquad u\ge t.
\]
For example, once $\alpha_t\norm A\le1/2$, the terms of degree
at least two have norm bounded by
\[
 \sum_{r\ge2}(\alpha_u\norm A)^r/r\le C\alpha_u^2.
\]
All these matrices are functions of $A$ and commute.
Thus their logarithms can be added in the product, giving
\[
 \Phi_{t+1:t+j-1}
 =\exp\{-Aa(j-1)+\mathcal E_{t,j}\},\qquad
 \max_{j\le L/a}\norm{\mathcal E_{t,j}}
 \le C\left(\frac{L^2}{\tau a}+La\right)\longrightarrow0.
\]
The main exponent is bounded on this range of indices.
Since the two exponents commute, the exponential series gives
\[
 \norm{e^{B+E}-e^B}
 =\norm{e^B(e^E-I)}
 \le e^{\norm B}(e^{\norm E}-1)
 \le e^{\norm B+\norm E}\norm E.
\]
Taking $B=-Aa(j-1)$ and $E=\mathcal E_{t,j}$, we obtain
\begin{equation}\label{lb:local-product-limit}
 \max_{1\le j\le L/a}
 \norm{\Phi_{t+1:t+j-1}-e^{-Aa(j-1)}}\longrightarrow0.
\end{equation}
This calculation includes the empty product at $j=1$.

Multiply the truncated sum in \eqref{lb:infinite-difference} by
$\tau a$. By \eqref{lb:local-step-difference},
\[
 (\tau a)(\alpha_t-\alpha_{t+j})
 =\beta a^2j\,(1+o(1))
\]
uniformly for $j\le L/a$.
The sum of $a^2j$ over this range is bounded by a constant depending
only on $L$.
We can therefore use \eqref{lb:local-product-limit} term by term
in the truncated sum. We obtain
\begin{align}
 &\tau a\sum_{j=1}^{\lfloor L/a\rfloor}
       (\alpha_t-\alpha_{t+j})\Phi_{t+1:t+j-1}\notag\\
 &\qquad=\beta\sum_{j=1}^{\lfloor L/a\rfloor}
            a(aj)e^{-Aa(j-1)}+o(1)
 \longrightarrow \beta\int_0^L x e^{-Ax}\,dx.
 \label{lb:interior-riemann}
\end{align}
The last step is a Riemann sum with mesh $a$.
Replacing $a(j-1)$ by $aj$ in the exponential changes the sum by
$O(a)$ for fixed $L$, since the matrix exponential has a bounded
derivative on that interval.

We now control the omitted part of the infinite sum.
For $k>t$, let
\[
 s_{t,k}=\sum_{u=t+1}^{k-1}\alpha_u,\qquad
 \eta_t=\frac\beta{c_0}(t+b)^{\beta-1}
        =\frac\beta{\tau a}.
\]
By Lemma~\ref{lb:lem-weights},
\[
 0\le\alpha_t-\alpha_k
 \le\alpha_k\eta_t(\alpha_t+s_{t,k})
                         e^{\eta_t(\alpha_t+s_{t,k})}.
\]
Let $\lambda=\lambda_{\min}(A)$.
The shift condition gives $\eta_t\le\lambda/4$, and the product
is bounded by $e^{-\lambda s_{t,k}}$.
Since $\tau a\eta_t=\beta$ and $\alpha_t\le1$, the normalized
tail has norm at most
\begin{equation}\label{lb:normalized-infinite-tail}
 C\sum_{k>t+\lfloor L/a\rfloor}
       \alpha_k(1+s_{t,k})e^{-(3\lambda/4)s_{t,k}}.
\end{equation}
At the first index in this tail, the cumulative stepsize tends to
$L$. This follows from \eqref{lb:local-clock-error} with the
corresponding number of factors.
The interval comparison in Lemma~\ref{lb:lem-weights} consequently
bounds the limit superior of \eqref{lb:normalized-infinite-tail} by
\[
 C\int_{L/2}^\infty(1+x)e^{-cx}\,dx.
\]
The integral tends to zero as $L\to\infty$.
Taking the limit first in $t$ for fixed $L$ and then in $L$, we obtain
\begin{equation}\label{lb:infinite-coefficient}
 \tau a(W_{\infty,t}-A^{-1})
 \longrightarrow \beta\int_0^\infty x e^{-Ax}\,dx
 =\beta A^{-2}.
\end{equation}
The matrix integral is evaluated in Lemma~\ref{lb:lem-lyapunov}.
Since $(\tau a)^{-1}=c_0^{-1}(t+b)^{\beta-1}$, this proves
\eqref{lb:infinite-asymptotic}.
The same estimates give the domination
\begin{equation}\label{lb:infinite-domination}
 \norm{W_{\infty,t}-A^{-1}}\le C(t+b)^{\beta-1},
\end{equation}
including the finitely many early indices after increasing $C$.

We next average the covariance of these infinite-horizon weights.
Let $q_t=(t+b)^{\beta-1}$ and write
\[
 W_{\infty,t}=R+\frac\beta{c_0}q_t A^{-2}+E_t,\qquad
 \norm{E_t}=o(q_t),\qquad
 \norm{W_{\infty,t}-R}\le Cq_t.
\]
Expanding the two weights on the two sides of $S$ gives
\begin{align}
 W_{\infty,t}SW_{\infty,t}^\top-RSR^\top
 &=\frac\beta{c_0}q_t
       \{A^{-2}SA^{-\top}+A^{-1}S(A^{-2})^\top\}
       +H_t,\notag\\
 \norm{H_t}&\le C\norm{E_t}+Cq_t^2.
 \label{lb:interior-covariance-expansion}
\end{align}
The first two remainder terms are $E_tSR^\top$ and $RSE_t^\top$.
The remaining term is the product of the two weight differences,
which is bounded by $Cq_t^2$.

By Lemma~\ref{lb:lem-sums},
\[
 \frac1n\sum_{t=0}^{n-1}q_t
 =\frac1\beta n^{\beta-1}+O(n^{-1}),\qquad
 \frac1n\sum_{t=0}^{n-1}q_t^2
 =O(n^{2\beta-2})=o(n^{\beta-1}).
\]
The same lemma, applied to $E_t=o(q_t)$, gives
$n^{-1}\sum_{t<n}\norm{E_t}=o(n^{\beta-1})$.
Therefore,
\begin{align}
 \frac1n\sum_{t=0}^{n-1}
      \{W_{\infty,t}SW_{\infty,t}^\top-\Sigma_\infty\} =\frac{n^{\beta-1}}{c_0}
      \{A^{-2}SA^{-\top}+A^{-1}S(A^{-2})^\top\}
           +o(n^{\beta-1}).
 \label{lb:averaged-interior}
\end{align}
The factor $\beta$ from the local weight expansion cancels with
the factor $1/\beta$ in the power sum.

It remains to replace the infinite-horizon weights by the weights
truncated at time $n$. Let $a_n=\alpha_n$ and write $t=n-k$.
For fixed $L>0$, consider $1\le k\le L/a_n$.
Since $na_n\to\infty$, all these indices satisfy $k/n\to0$
uniformly. For $n-k\le u\le n$, we have
\begin{equation}\label{lb:terminal-step-ratio}
 1\le\frac{\alpha_u}{a_n}
 \le\left(1-\frac{k}{n+b}\right)^{-\beta}
 =1+o(1)
\end{equation}
uniformly over this range.
The product comparison used in
\eqref{lb:local-clock-error}--\eqref{lb:local-product-limit} can
now be applied with the common stepsize $a_n$ throughout the terminal
interval. The cumulative stepsize error is at most
\[
 L\max_{n-k\le u\le n}|\alpha_u/a_n-1|=o(1),
\]
and the sum of squared stepsizes is $O(La_n)=o(1)$.
Thus all the products in the finite sum are uniformly approximated
by the corresponding constant-stepsize exponentials.

In the notation $x_k=ka_n$, this gives
\begin{align}
 W_{n,n-k}
 &=\frac{\alpha_{n-k}}{a_n}
       \left\{a_n\sum_{j=1}^k e^{-Aa_n(j-1)}+o(1)\right\}
       \notag\\
 &=\int_0^{x_k}e^{-As}\,ds+o(1)
  =R(I-e^{-Ax_k})+o(1),
 \label{lb:terminal-finite-limit}
\end{align}
uniformly for $1\le k\le L/a_n$.
The error in summing the product approximation is bounded by
$ka_n$ times the largest product error, so it tends to zero
uniformly on this range.
The Riemann sum for the integral is also uniform in its upper
endpoint because the integrand has bounded derivative on $[0,L]$.
By \eqref{lb:infinite-domination},
\begin{equation}\label{lb:terminal-infinite-limit}
 \max_{1\le k\le L/a_n}\norm{W_{\infty,n-k}-R}\longrightarrow0,
\end{equation}
since the smallest time index in this range tends to infinity.
These estimates also cover grid points tending to zero.

Define the difference of the two covariance summands by
\[
 F_{n,k}=W_{n,n-k}SW_{n,n-k}^\top
               -W_{\infty,n-k}SW_{\infty,n-k}^\top,
 \qquad 1\le k\le n.
\]
The two uniform limits above imply convergence on every bounded
interval of the grid $x_k=ka_n$ to the continuous matrix function
\begin{equation}\label{lb:terminal-integrand}
 F(x)=R(I-e^{-Ax})S(I-e^{-A^\top x})R^\top-\Sigma_\infty.
\end{equation}
To justify summation over all $k$, we use the exact tail identity
from Lemma~\ref{lb:lem-weights}:
\[
 W_{\infty,t}-W_{n,t}
 =\frac{\alpha_t}{\alpha_n}\Phi_{t+1:n}W_{\infty,n}.
\]
Its exponential bound and uniform boundedness of the weights give
\begin{align}
 \norm{F_{n,k}}
 &\le C\norm{W_{n,n-k}-W_{\infty,n-k}}\notag\\
 &\le C\exp\left(-c\sum_{u=n-k+1}^n\alpha_u\right)
 \le Ce^{-cka_n}.
 \label{lb:terminal-domination}
\end{align}
In the last inequality, every stepsize in the sum is at least $a_n$.
The limiting integrand has the same type of bound. Expanding its
two factors and using $\norm{e^{-Ax}}\le e^{-\lambda x}$ gives
\[
 \norm{F(x)}
 \le\norm R^2\norm S\{2e^{-\lambda x}+e^{-2\lambda x}\}
 \le Ce^{-\lambda x}.
\]
Thus the continuous tail is also integrable.

We can now apply Lemma~\ref{lb:lem-riemann} with mesh $a_n$.
Its hypotheses are precisely the local uniform convergence in
\eqref{lb:terminal-finite-limit}--\eqref{lb:terminal-infinite-limit}
and the domination \eqref{lb:terminal-domination}.
It gives
\[
 a_n\sum_{k=1}^nF_{n,k}\longrightarrow\int_0^\infty F(x)\,dx.
\]
Since
\[
 \frac1{na_n}
 =\frac{(n+b)^\beta}{c_0n}
 =\frac{n^{\beta-1}}{c_0}(1+o(1)),
\]
the terminal contribution is
\begin{align}
 &\frac1n\sum_{t=0}^{n-1}
   \{W_{n,t}SW_{n,t}^\top-W_{\infty,t}SW_{\infty,t}^\top\}\notag\\
 &\qquad=\frac{n^{\beta-1}}{c_0}\int_0^\infty F(x)\,dx
                                      +o(n^{\beta-1}).
 \label{lb:averaged-terminal}
\end{align}

We evaluate the integral without assuming that $S$ commutes with $A$.
Expanding \eqref{lb:terminal-integrand} gives
\[
 F(x)=-Re^{-Ax}SR^\top-RSe^{-A^\top x}R^\top
                   +Re^{-Ax}Se^{-A^\top x}R^\top.
\]
Using $\int_0^\infty e^{-Ax}\,dx=A^{-1}$ and the definition of
$\mathcal X$, we obtain
\begin{equation}\label{lb:terminal-coefficient}
 \int_0^\infty F(x)\,dx
 =-A^{-2}SA^{-\top}-A^{-1}S(A^{-2})^\top
                         +A^{-1}\mathcal X A^{-\top}.
\end{equation}
Adding \eqref{lb:averaged-interior} and
\eqref{lb:averaged-terminal}, the first two terms in
\eqref{lb:terminal-coefficient} cancel the interior correction.
We are left with
\[
 C_n-\Sigma_\infty
 =\frac{n^{\beta-1}}{c_0}A^{-1}\mathcal X A^{-\top}
                              +o(n^{\beta-1}).
\]
Finally, $S\succ0$ implies $\mathcal X\succ0$ by
Lemma~\ref{lb:lem-lyapunov}. Since $A$ is invertible and $c_0>0$,
the same is true of $G$.
This proves Proposition~\ref{lb:prop-cov}.
\end{proof}

\subsection{Problem Specific Bounds}\label{lb:support}

\begin{lemma}\label{lb:lem-weights}
For the matrix $A$ in \eqref{lb:matrices}, let
$\lambda=\lambda_{\min}(A)$. The infinite-horizon weights are well defined and uniformly bounded. More precisely,
\begin{align}
 \norm{\Phi_{i:j}}&\le
       \exp\left(-\lambda\sum_{u=i}^j\alpha_u\right),
       &\norm{A^{-1}}&\le\lambda^{-1},\notag\\
 \sup_t\norm{W_{\infty,t}}+\sup_{n,t<n}\norm{W_{n,t}}&\le C,
       &\norm{W_{\infty,t}-A^{-1}}&\le C(t+b)^{\beta-1}.
 \label{lb:deterministic-weight-bounds}
\end{align}
The exact identities are
\begin{align}
 \sum_{k=t+1}^\infty\alpha_k\Phi_{t+1:k-1}&=A^{-1},
 \label{lb:weight-telescope}\\
 W_{\infty,t}-W_{n,t}
 &=\frac{\alpha_t}{\alpha_n}\Phi_{t+1:n}W_{\infty,n}.
 \label{lb:weight-exact-tail}
\end{align}
The tail satisfies
\begin{equation}\label{lb:weight-tail-bound}
 \norm{W_{\infty,t}-W_{n,t}}
 \le C\exp\left(-\frac{3\lambda}{4}
                      \sum_{u=t+1}^n\alpha_u\right),\qquad t<n.
\end{equation}
For $s_{t,k}=\sum_{u=t+1}^{k-1}\alpha_u$ and any $d>0$, one also has
\begin{equation}\label{lb:clock-tail-sum}
 \sum_{k\ge m}\alpha_k(1+s_{t,k})e^{-d s_{t,k}}
 \le e^d\int_{s_{t,m}}^\infty(1+x)e^{-dx}\,dx,
 \qquad m\ge t+1.
\end{equation}
\end{lemma}

\begin{proof}
The explicit eigenvalues satisfy
$0<\lambda\le\lambda_{\max}(A)<1$ and
$\lambda\ge a_\star=2/25$.
For $0\le\alpha\le1$, the matrix $I-\alpha A$ is symmetric
with nonnegative eigenvalues. Its largest eigenvalue is
$1-\lambda\alpha$, so
\[
 \norm{I-\alpha A}=1-\lambda\alpha\le e^{-\lambda\alpha}.
\]
Multiplying this inequality proves the product bound.
The inverse bound follows from the smallest eigenvalue.
Both assertions also respect the empty-product convention.

To compare stepsizes, define
\[
 \eta_t=\frac\beta{c_0}(t+b)^{\beta-1}.
\]
The shift condition and $\lambda\ge a_\star$ give
$\eta_t\le\eta_0\le\lambda/4$.
For $k>t$, using $\log(1+x)\le x$, we have
\begin{align}
 \log\frac{\alpha_t}{\alpha_k}
 &=\beta\sum_{u=t}^{k-1}\log\left(1+\frac1{u+b}\right)\notag\\
 &\le\frac\beta{c_0}\sum_{u=t}^{k-1}
                         (u+b)^{\beta-1}\alpha_u\notag\\
 &\le\eta_t\sum_{u=t}^{k-1}\alpha_u
   =\eta_t(\alpha_t+s_{t,k}).
 \label{lb:ratio-bound}
\end{align}
The second inequality uses the fact that $\beta-1<0$.
Exponentiating gives
\begin{equation}\label{lb:ratio-exponential}
 \frac{\alpha_t}{\alpha_k}
 \le e^{\lambda/4}e^{(\lambda/4)s_{t,k}}.
\end{equation}

We next prove \eqref{lb:clock-tail-sum}.
The intervals $[s_{t,k},s_{t,k+1}]$ have lengths $\alpha_k\le1$.
Since $\sum_u\alpha_u=\infty$, they partition $[0,\infty)$
up to endpoints. For a point $x$ in the interval indexed by $k$,
\[
 e^{-d s_{t,k}}\le e^d e^{-dx},\qquad
 1+s_{t,k}\le1+x.
\]
Multiplying and integrating over that interval gives
\[
 \alpha_k(1+s_{t,k})e^{-d s_{t,k}}
 \le e^d\int_{s_{t,k}}^{s_{t,k+1}}(1+x)e^{-dx}\,dx.
\]
Summing from $k=m$ onward proves the stated bound.
In particular, the full sum is uniformly bounded, since
\[
 \int_0^\infty(1+x)e^{-dx}\,dx=\frac1d+\frac1{d^2}.
\]

Insert $\alpha_k/\alpha_k$ into the infinite weight series.
By the product estimate and \eqref{lb:ratio-exponential},
\begin{align*}
 \alpha_t\sum_{k>t}\norm{\Phi_{t+1:k-1}} \le\sum_{k>t}\alpha_k\frac{\alpha_t}{\alpha_k}
                                     e^{-\lambda s_{t,k}} \le e^{\lambda/4}\sum_{k>t}\alpha_k
                                     e^{-(3\lambda/4)s_{t,k}}
 \le C.
\end{align*}
This proves absolute convergence and the uniform bound for
$W_{\infty,t}$. The same estimate applies to every finite partial
sum, proving the uniform bound for $W_{n,t}$.

Since every product factor is a polynomial in $A$, it commutes
with $A$. For $k>t$,
\[
 \alpha_kA\Phi_{t+1:k-1}
 =\Phi_{t+1:k-1}-\Phi_{t+1:k}.
\]
Summing to a finite upper index $m$ gives
\[
 A\sum_{k=t+1}^m\alpha_k\Phi_{t+1:k-1}=I-\Phi_{t+1:m}.
\]
The product on the right tends to zero as $m\to\infty$.
This proves \eqref{lb:weight-telescope}.

Subtracting the telescoping identity from the infinite weight
series gives
\[
 W_{\infty,t}-A^{-1}
 =\sum_{k>t}\alpha_k
       \left(\frac{\alpha_t}{\alpha_k}-1\right)\Phi_{t+1:k-1}.
\]
For $x\ge0$, $e^x-1\le xe^x$.
Using this inequality and \eqref{lb:ratio-bound}, we obtain
\[
 0\le\frac{\alpha_t}{\alpha_k}-1
 \le\eta_t(\alpha_t+s_{t,k})
                         e^{\eta_t(\alpha_t+s_{t,k})}.
\]
Thus, by $\alpha_t\le1$ and $\eta_t\le\lambda/4$,
\begin{align*}
 \norm{W_{\infty,t}-A^{-1}} \le C\eta_t\sum_{k>t}\alpha_k(1+s_{t,k}) e^{-(3\lambda/4)s_{t,k}} \le C\eta_t\le C(t+b)^{\beta-1}.
\end{align*}
This proves the remaining bound in
\eqref{lb:deterministic-weight-bounds}.

For the omitted tail, factor the products at time $n$:
\begin{align*}
 W_{\infty,t}-W_{n,t}
 &=\alpha_t\sum_{k=n+1}^\infty\Phi_{t+1:k-1} =\alpha_t\Phi_{t+1:n}\sum_{k=n+1}^\infty\Phi_{n+1:k-1}
 =\frac{\alpha_t}{\alpha_n}\Phi_{t+1:n}W_{\infty,n}.
\end{align*}
This proves \eqref{lb:weight-exact-tail}.
Taking $k=n$ in \eqref{lb:ratio-bound}, we have
\[
 \frac{\alpha_t}{\alpha_n}
 \le e^{\lambda/4}
       \exp\left(\frac\lambda4\sum_{u=t+1}^n\alpha_u\right).
\]
Combining this with the product estimate and the uniform bound
for $W_{\infty,n}$ proves \eqref{lb:weight-tail-bound}.
\end{proof}

Under the stationary initialization used in the lower bound,
$C_n$ is also the actual covariance of $M_n$.
For this one-action model with deterministic $Q_0$, the filtration
defined in Section~\ref{mb:preliminaries} reduces to
$\mathcal F_t=\sigma(s_0,\ldots,s_{t+1})$.
The Bellman equation gives
$\E[\varepsilon_t\mid\mathcal F_{t-1}]=0$ for $t\ge1$ and
$\E\varepsilon_0=0$.
For $s<t$, the earlier increment is measurable with respect to
$\mathcal F_{t-1}$, so
\[
 \E[\varepsilon_s\varepsilon_t^\top]
 =\E\{\varepsilon_s\E[\varepsilon_t^\top\mid\mathcal F_{t-1}]\}=0.
\]
The stationary marginal second moment equals $S$ at every time.
Since the weights are deterministic, expanding the covariance
therefore gives
\[
 \Cov(M_n)
 =\frac1n\sum_{s,t<n}W_{n,s}
                \E[\varepsilon_s\varepsilon_t^\top]W_{n,t}^\top
 =\frac1n\sum_{t<n}W_{n,t}SW_{n,t}^\top=C_n.
\]

\begin{lemma}\label{lb:lem-stability}
Let $A\succ0$ be symmetric and let $B_0,C_0,E_0$ be fixed
matrices of compatible sizes. Define
\[
 \mathcal B(a)=\begin{pmatrix}
 I-aA&aB_0\\aC_0&\frac12I+aE_0
 \end{pmatrix}.
\]
There exist $a_0,c>0$ such that, for $0<a\le a_0$,
\begin{equation}\label{lb:common-quadratic-bound}
 \norm{\mathcal B(a)(u,f)^\top}_2^2
 \le(1-ca)(\norm u_2^2+\norm f_2^2).
\end{equation}
For the matrices in \eqref{lb:T-matrices} and every decreasing
power stepsize, this implies
\[
 \norm{(T_0+\alpha_tT_1)\cdots(T_0+\alpha_kT_1)}
 \le C\exp\left(-c'\sum_{u=k}^t\alpha_u\right),\qquad k\le t.
\]
\end{lemma}

\begin{proof}
Let $\lambda=\lambda_{\min}(A)>0$, $L=\norm A$, and denote
$K_B=\norm{B_0}$, $K_C=\norm{C_0}$, and $K_E=\norm{E_0}$.
Write the two blocks after applying the matrix as
\[
 u'=(I-aA)u+aB_0f,\qquad
 f'=aC_0u+(\tfrac12I+aE_0)f.
\]
Choose $a$ small enough that $aL\le1$ and $aL^2\le\lambda/2$.
Positive definiteness and symmetry give
\[
 \norm{(I-aA)u}_2^2
 \le(1-2\lambda a+L^2a^2)\norm u_2^2,
 \qquad \norm{I-aA}\le1.
\]
Expanding the square of the first block and applying Young's
inequality to its cross term, we have
\begin{align*}
 \norm{u'}_2^2
 &\le(1-2\lambda a+L^2a^2)\norm u_2^2
       +2aK_B\norm u_2\norm f_2+a^2K_B^2\norm f_2^2\\
 &\le(1-2\lambda a+L^2a^2+\lambda a/2)\norm u_2^2
       +(2K_B^2a/\lambda+K_B^2a^2)\norm f_2^2\\
 &\le(1-\lambda a)\norm u_2^2+C_Ba\norm f_2^2,
\end{align*}
where we also require $a\le1$ and can take
$C_B=K_B^2(1+2/\lambda)$.

For the second block, choose $aK_E\le1/8$.
Then $\norm{I/2+aE_0}\le5/8$.
Using $\norm{x+y}_2^2\le(3/2)\norm x_2^2+3\norm y_2^2$ with
$x=(I/2+aE_0)f$ and $y=aC_0u$, we obtain
\begin{align*}
 \norm{f'}_2^2
 \le\frac32\left(\frac58\right)^2\norm f_2^2
                        +3K_C^2a^2\norm u_2^2 \le\frac34\norm f_2^2+C_Ca^2\norm u_2^2,
 \qquad C_C=3K_C^2.
\end{align*}
Adding the two estimates gives
\[
 \norm{u'}_2^2+\norm{f'}_2^2
 \le(1-\lambda a+C_Ca^2)\norm u_2^2
                         +(3/4+C_Ba)\norm f_2^2.
\]
Decrease $a_0$ so that, in addition to the earlier conditions,
\[
 C_Ca_0\le\lambda/2,\qquad
 (C_B+\lambda/2)a_0\le1/4.
\]
Both coefficients are then at most $1-\lambda a/2$.
This proves \eqref{lb:common-quadratic-bound} with $c=\lambda/2$.

Taking square roots and using $\sqrt{1-x}\le e^{-x/2}$ gives
\[
 \norm{\mathcal B(a)}\le e^{-\lambda a/4},\qquad 0<a\le a_0.
\]
This estimate holds in the same Euclidean norm for every such $a$.
Thus, once the stepsizes are at most $a_0$, multiplying the norm
bounds gives an exponential in their sum.
By \eqref{lb:transformed-recursion}, the original matrices are
conjugate to these block matrices by the fixed invertible matrix
$M_0$. Conjugating the full product contributes the fixed factor
$\norm{M_0}\norm{M_0^{-1}}$.

There are only finitely many indices before the stepsizes enter
this range. Products containing these earlier factors can be split
into a later contracting part and a finite earlier part.
The norms of all possible earlier parts, multiplied by the
exponential of their finite accumulated stepsizes, have a finite
maximum. Including this maximum in $C$ proves the bound for all
$k\le t$. The common norm estimate just established is what
allows the factors to have different stepsizes.
\end{proof}

\subsection{Supporting Lemmas}\label{lb:support2}

\begin{lemma}[Proposition A.1 in \cite{jiang2024generalized}]\label{sr:integral}
If $f$ is continuous and non-increasing on $[a,b+1]$, where $a\le b$ are integers, then
\[
 \int_a^{b+1}f(x)\,dx\le\sum_{k=a}^b f(k)
 \le f(a)+\int_a^b f(x)\,dx.
\]
In particular, this applies to every decreasing power function used in Lemma~\ref{lb:lem-sums}.
\end{lemma}

\begin{lemma}[Lemma~2.4(a) in \cite{jiang2024generalized}]\label{sr:chung}
We use the asymptotic consequence of the cited lemma in arXiv version~2,
with the shift denoted by $\eta$.
Let $0<\nu<1$, $q,c,d>0$, and $\eta>0$.
Suppose a nonnegative scalar sequence satisfies, from some fixed index onward,
\[
 a_{k+1}\le\left(1-\frac{c}{(k+\eta)^\nu}\right)a_k
                   +\frac{d}{(k+\eta)^{\nu+q}}.
\]
Then $a_k=O((k+\eta)^{-q})$.
More precisely, after shifting the initial index so that
$\eta\ge c^{1/\nu}$ and $\eta>(q/c)^{1/(1-\nu)}$,
the cited lemma bounds $a_k$ by a constant times
$(k+\eta)^{-q}+\exp[-c(k+\eta)^{1-\nu}/(1-\nu)]$.
The finite prefix affects only the constants.
\end{lemma}

\begin{lemma}[Theorem~2.24 in \cite{folland1999}]\label{sr:dct}
If measurable scalar functions $f_n$ converge almost everywhere to $f$ and
$|f_n|\le g$ for some $g\in L^1$, then $f\in L^1$ and
$\int f_n\to\int f$. The same conclusion holds in any fixed matrix norm for
fixed-size matrix-valued functions with $\|F_n\|\le g$ and $F_n\to F$ almost everywhere.
\end{lemma}

\begin{lemma}\label{lb:lem-sums}
Let $\alpha_t=c_0(t+b)^{-\beta}$ with $b\ge1$ and $1/2<\beta<1$.
For fixed $c_0,b,\beta$,
\begin{align}
 \sum_{t=1}^n\alpha_t
 &=\frac{c_0}{1-\beta}n^{1-\beta}+O(1),
 &\sum_{t=1}^n\alpha_t^2&=O(1),\notag\\
 \sum_{t=1}^n(t+b)^{-1}&=O(\log(en)),
 &\sum_{t=0}^\infty\frac{\alpha_t}{t+b}&<\infty.
 \label{lb:power-sums}
\end{align}
If $q_t=(t+b)^{\beta-1}$, then
\begin{equation}\label{lb:q-sums}
 \sum_{t=0}^{n-1}q_t=\frac1\beta n^\beta+O(1),\qquad
 \sum_{t=0}^{n-1}q_t^2=O(n^{2\beta-1}).
\end{equation}
For any sequence of matrices with $\norm{E_t}=o(q_t)$,
\begin{equation}\label{lb:averaged-little-o}
 \frac1n\sum_{t=0}^{n-1}\norm{E_t}=o(n^{\beta-1}).
\end{equation}
\end{lemma}

\begin{proof}
Apply the integral test in Lemma~\ref{sr:integral} to
$(x+b)^{-\beta}$, $(x+b)^{-2\beta}$, $(x+b)^{-1}$,
$(x+b)^{-1-\beta}$, $(x+b)^{\beta-1}$, and $(x+b)^{2\beta-2}$.
Their elementary integrals give \eqref{lb:power-sums}--\eqref{lb:q-sums};
fixed shifts change the displayed divergent power sums by $O(1)$.
For \eqref{lb:averaged-little-o}, choose a fixed $T$ with
$\|E_t\|\le\epsilon q_t$ for $t\ge T$ and write
\[
 n^{-\beta}\sum_{t<n}\|E_t\|
 \le n^{-\beta}\sum_{t<T}\|E_t\|+\epsilon n^{-\beta}\sum_{T\le t<n}q_t.
\]
Its limit superior is at most $\epsilon/\beta$; let $\epsilon\downarrow0$.
\end{proof}

\begin{lemma}\label{lb:lem-tracking}
Let $\alpha_t=c_0(t+b)^{-\beta}$ with $0<\beta<1$.
Suppose the deterministic matrices $B_t$ satisfy
\[
 \norm{B_t\cdots B_k}
 \le C\exp\left(-c\sum_{u=k}^t\alpha_u\right),\qquad k\le t,
\]
and, from a fixed time $t_0$ onward,
\[
 \chi_{t+1}=B_t\chi_t+d_t,\qquad
 \norm{d_t}\le C\frac{\alpha_t}{t+b}.
\]
For every fixed initial error, $\norm{\chi_t}=O((t+b)^{-1})$.
\end{lemma}

\begin{proof}
Unroll the matrix recursion using the assumed product bound.
For a sufficiently large fixed $C$, the resulting scalar upper bound is the solution of
\[
 u_{t_0}=C\|\chi_{t_0}\|,\qquad
 u_{t+1}=e^{-c\alpha_t}u_t+C\frac{\alpha_t}{t+b},\qquad
 \|\chi_t\|\le u_t.
\]
The last inequality follows from the unrolled products, and does not require a one-step
contraction bound on $B_t$.
Eventually $e^{-c\alpha_t}\le1-(c/2)\alpha_t$, so
\[
 u_{t+1}\le\left(1-\frac{cc_0/2}{(t+b)^\beta}\right)u_t
              +\frac{Cc_0}{(t+b)^{\beta+1}}.
\]
Lemma~\ref{sr:chung}, with $\nu=\beta$, $q=1$, and shift $\eta=b$, gives
$u_t=O((t+b)^{-1})$. A finite initial prefix only changes the constant.
\end{proof}

\begin{lemma}\label{lb:lem-riemann}
Let $a_n>0$ satisfy $a_n\to0$ and $na_n\to\infty$.
Let $F_{n,k}$, $1\le k\le n$, be matrices of a fixed size.
Suppose there is a continuous matrix function $F$ on $[0,\infty)$
such that, for every fixed $L>0$,
\begin{equation}\label{lb:riemann-local-hypothesis}
 \max_{1\le k\le\lfloor L/a_n\rfloor}
       \norm{F_{n,k}-F(ka_n)}\longrightarrow0.
\end{equation}
Assume also that $\norm{F_{n,k}}\le Ce^{-cka_n}$ and
$\norm{F(x)}\le Ce^{-cx}$ for fixed $C,c>0$.
Then
\begin{equation}\label{lb:riemann-conclusion}
 a_n\sum_{k=1}^nF_{n,k}\longrightarrow\int_0^\infty F(x)\,dx.
\end{equation}
\end{lemma}

\begin{proof}
Define the matrix-valued step function
\[
 \widetilde F_n(x)=
 \begin{cases}F_{n,\lceil x/a_n\rceil},&0<x\le na_n,\\0,&x>na_n.\end{cases}
\]
For every fixed $x>0$, $na_n\to\infty$, the local uniform approximation,
and continuity of $F$ imply $\widetilde F_n(x)\to F(x)$.
Moreover, $\|\widetilde F_n(x)\|\le Ce^{-cx}$, since
$a_n\lceil x/a_n\rceil\ge x$.
Lemma~\ref{sr:dct}, applied entrywise, now gives
\[
 a_n\sum_{k=1}^nF_{n,k}=\int_0^\infty\widetilde F_n(x)\,dx
 \longrightarrow\int_0^\infty F(x)\,dx.
\]
\end{proof}

\begin{lemma}\label{lb:lem-lyapunov}
Let $A\succ0$ be symmetric and $S\succ0$.
Then
\begin{equation}\label{lb:matrix-laplace-integrals}
 \int_0^\infty e^{-As}\,ds=A^{-1},\qquad
 \int_0^\infty s e^{-As}\,ds=A^{-2}.
\end{equation}
The matrix
$\mathcal X=\int_0^\infty e^{-As}Se^{-A^\top s}\,ds$
is the unique solution to
$A\mathcal X+\mathcal X A^\top=S$, and $\mathcal X\succ0$.
For \eqref{lb:matrices}, its value is
\begin{equation}\label{lb:explicit-lyapunov}
 \mathcal X=\frac1{6615}
        \begin{pmatrix}3392&1752\\1752&3762\end{pmatrix}.
\end{equation}
\end{lemma}

\begin{proof}
Diagonalizing the symmetric positive-definite $A$ and applying
\[
 \int_0^\infty e^{-\lambda s}ds=\lambda^{-1}
 \quad\text{and}\quad
 \int_0^\infty s e^{-\lambda s}ds=\lambda^{-2}
\]
proves \eqref{lb:matrix-laplace-integrals}.
Lemma~\ref{sr:sylvester}, with $U=V=A$, gives the covariance integral and uniqueness.
For $u\ne0$, its quadratic form is
$\int_0^\infty\|S^{1/2}e^{-A^\top s}u\|_2^2ds>0$, proving positivity.
For the particular matrices in \eqref{lb:matrices}, direct multiplication of the matrix
in \eqref{lb:explicit-lyapunov} gives
$A\mathcal X+\mathcal X A^\top=\diag(64/375,56/375)=S$;
uniqueness identifies it as the claimed covariance.
\end{proof}

\begin{lemma}\label{lb:lem-gaussian}
For random vectors $U,V$ with finite first moments and a
$1$-Lipschitz function $h$,
\begin{equation}\label{lb:lipschitz-test}
 |\E h(U)-\E h(V)|\le W_1(U,V).
\end{equation}
In particular, $W_1(U,V)\ge\norm{\E U-\E V}_2$.
For centered scalar Gaussians with standard deviations
$\sigma,\tau\ge0$,
\begin{equation}\label{lb:scalar-gaussian-distance}
 W_1(\N(0,\sigma^2),\N(0,\tau^2))
 =\sqrt{\frac2\pi}\,|\sigma-\tau|.
\end{equation}
Projection onto any coordinate does not increase $W_1$.
Consequently, if $V_n\to V\succ0$ and
$(V_n)_{jj}=V_{jj}+g_j\delta_n+o(\delta_n)$ with
$g_j>0$ and $\delta_n\downarrow0$, then
\begin{equation}\label{lb:gaussian-marginal-expansion}
 W_1(\N(0,V_n),\N(0,V))
 \ge\sqrt{\frac2\pi}\frac{g_j}{2\sqrt{V_{jj}}}\delta_n
                                      +o(\delta_n).
\end{equation}
\end{lemma}

\begin{proof}
Lemma~\ref{sr:wasserstein} gives the Lipschitz test inequality, the mean lower bound,
and contraction under coordinate projection.
If $G\sim\N(0,1)$, the coupling $(\sigma G,\tau G)$ gives an upper bound
$\sqrt{2/\pi}|\sigma-\tau|$; the $1$-Lipschitz test $x\mapsto|x|$ gives the same lower bound.
This proves \eqref{lb:scalar-gaussian-distance}.
For the variance expansion, write $\sigma_n=\sqrt{(V_n)_{jj}}$ and
$\sigma=\sqrt{V_{jj}}>0$. Then
\[
 \sigma_n-\sigma
 =\frac{(V_n)_{jj}-V_{jj}}{\sigma_n+\sigma}
 =\frac{g_j}{2\sqrt{V_{jj}}}\delta_n+o(\delta_n).
\]
Projection and the scalar formula give \eqref{lb:gaussian-marginal-expansion}.
\end{proof}

\section{Proofs for Section \ref{sec_syn}}\label{syn:proofs}

In this section, we prove the synchronous upper and lower bounds. The upper bound follows from Proposition~\ref{syn:proxy}, which separates the Bellman nonlinearity, the Gaussian approximation of the sampled fixed-policy recursion, and the deterministic covariance comparison. The lower bound uses the same MDP as in the asynchronous setting. Section~\ref{syn:support} provides supporting lemmas used below. Throughout this section, $Q_t$ and $X_n$ refer to the synchronous algorithm \eqref{syn:algorithm}.

\subsection{Proof of Theorem~\ref{syn:upper}}\label{syn:upper-proof}

\synupperthm*

\begin{proof}
Let $\lambda_{\mathrm{syn}}=\lambda_{\min}(\Sigma_\infty^{\mathrm{syn}})>0$.
By \eqref{syn:covariance-bound}, there exists $n_0$ such that
$\lambda_{\min}(C_n^{\mathrm{syn}})\ge\lambda_{\mathrm{syn}}/2$ for $n\ge n_0$.
Lemma~\ref{lem:gaussian} then gives
\[
 W_1(Z_n^{\mathrm{syn}},Z_\infty^{\mathrm{syn}})
 \le C\|C_n^{\mathrm{syn}}-\Sigma_\infty^{\mathrm{syn}}\|_F
 \le Cn^{\beta-1}.
\]
Using the coupling in \eqref{syn:linear-recursion}, the triangle inequality, and Proposition~\ref{syn:proxy}, we obtain
\begin{align*}
 W_1(X_n,Z_\infty^{\mathrm{syn}})
 &\le\E\|X_n-X_n^{\mathrm{lin}}\|_2
     +W_1(X_n^{\mathrm{lin}},Z_n^{\mathrm{syn}})
     +W_1(Z_n^{\mathrm{syn}},Z_\infty^{\mathrm{syn}})\\
 &\le C\{n^{-1/2}\log(en)+n^{-\beta}+n^{\beta-1}\}.
\end{align*}
The finitely many $2\le n<n_0$ are covered by increasing $C$, since
$\|X_n\|_2\le H\sqrt{dn}$ and $Z_\infty^{\mathrm{syn}}$ has a finite first moment.
Substitution of $\beta=1/2$ proves the endpoint bound.
For fixed $\beta>1/2$, both $n^{-1/2}\log(en)$ and $n^{-\beta}$ are $o(n^{\beta-1})$.
\end{proof}

\subsection{Proof of Proposition~\ref{syn:proxy}}\label{syn:proxy-proof}

\synproxyprop*

\begin{proof}
Write $\cF_t^{\mathrm{syn}}=\sigma(Q_0,\omega_0,\ldots,\omega_t)$ and
$\cF_{-1}^{\mathrm{syn}}=\sigma(Q_0)$.
The current block $\omega_t$ is independent of $\cF_{t-1}^{\mathrm{syn}}$.
Both $Q_t$ and $L_t$ are $\cF_{t-1}^{\mathrm{syn}}$-measurable and remain in $[0,H]^d$.
For later use, set
\[
 B_t=(1-\alpha_t)I+\alpha_t\gamma\widehat P_t^{\pi^\star},\qquad
 \rho_{i:j}=\prod_{u=i}^j(1-a_{\mathrm{syn}}\alpha_u),
\]
with $\rho_{i:j}=1$ when $i>j$.
Every sampled matrix is nonnegative with row sums $1-a_{\mathrm{syn}}\alpha_t$.
Thus, for every configuration of the sample blocks, including arbitrary replacements of any of them,
\begin{equation}\label{syn:sampled-contraction}
 \|B_j\cdots B_i\|_\infty\le\rho_{i:j}.
\end{equation}

\paragraph{Reduction to the fixed-policy recursion.}
Let
\[
 (\widehat{\mathcal T}_tq)(s,a)
 =r(s,a)+\gamma V(q)(S'_t(s,a)),\qquad
 g_t(q)=\widehat{\mathcal T}_tq-(r+\gamma\widehat P_t^{\pi^\star}q).
\]
With one action, $g_t=0$.
Otherwise, optimality of $\pi^\star$ gives, for every $q\in[0,H]^d$,
\begin{equation}\label{syn:gap-localization}
 \|g_t(q)\|_\infty
 \le2\gamma\|q-Q^\star\|_\infty
       \mathbf1\{\|q-Q^\star\|_\infty\ge\kappa/2\}.
\end{equation}
Below this threshold the greedy action is $\pi^\star$ at every state, and in general
$0\le V(q)(s)-q(s,\pi^\star(s))\le2\|q-Q^\star\|_\infty$.
Lemma~\ref{syn:moment-lemma} therefore implies
\[
 \E\|g_t(Q_t)\|_\infty
 \le2\gamma(2/\kappa)^5\E\|Q_t-Q^\star\|_\infty^6
 \le C\alpha_t^3.
\]
Since $3\beta>1$, these expectations are summable, including at $\beta=1/2$.
Subtracting \eqref{syn:linear-recursion} from \eqref{syn:algorithm} gives
\[
 Q_{t+1}-L_{t+1}=B_t(Q_t-L_t)+\alpha_tg_t(Q_t).
\]
Unrolling this recursion, applying \eqref{syn:sampled-contraction}, and then using the response bound in Lemma~\ref{syn:weight-lemma}, we obtain
\begin{align*}
 \E\|X_n-X_n^{\mathrm{lin}}\|_2
 &\le\frac{\sqrt d}{\sqrt n}\sum_{t<n}
       \alpha_t\sum_{k=t+1}^n\rho_{t+1:k-1}\,
                         \E\|g_t(Q_t)\|_\infty\\
 &\le\frac C{\sqrt n}\sum_{t<n}\alpha_t^3
 \le Cn^{-1/2}.
\end{align*}
This proves \eqref{syn:localization-bound}. The covariance bound
\eqref{syn:covariance-bound} is part of Lemma~\ref{syn:weight-lemma}.

\paragraph{Mean and covariance of the fixed-policy average.}
Write
\[
 e_t=L_t-Q^\star,\qquad
 \zeta_t=\gamma(P^{\pi^\star}-\widehat P_t^{\pi^\star}),\qquad
 \eta_t=\varepsilon_t^{\mathrm{syn}}-\zeta_te_t.
\]
Then
\begin{equation}\label{syn:martingale-recursion}
 e_{t+1}=(I-\alpha_tA^{\mathrm{syn}})e_t+\alpha_t\eta_t,
 \qquad \E[\eta_t\mid\cF_{t-1}^{\mathrm{syn}}]=0.
\end{equation}
Since $Q_0$ is deterministic, the mean is
$m_t:=\E e_t=\Phi^{\mathrm{syn}}_{0:t-1}e_0$.
Product contraction and $\beta<1$ give
\begin{equation}\label{syn:mean-bound}
 \|m_t\|_2\le C\exp\{-c(t+b)^{1-\beta}\},\qquad
 \|\E X_n^{\mathrm{lin}}\|_2\le Cn^{-1/2}.
\end{equation}
The constant in the first bound absorbs the fixed initial shift.
Set
\[
 F_n=X_n^{\mathrm{lin}}-\E X_n^{\mathrm{lin}},\qquad
 \Gamma_n=\Cov(F_n),\qquad J_t=\E[\eta_t\eta_t^\top].
\]
Although $\zeta_t$ and $\varepsilon_t^{\mathrm{syn}}$ use the same block, that block is independent of $e_t$.
Expanding the covariance accordingly gives the exact identity
\begin{align*}
 J_t-S^{\mathrm{syn}}
 ={}-\E[\varepsilon_t^{\mathrm{syn}}m_t^\top\zeta_t^\top]
     -\E[\zeta_tm_t(\varepsilon_t^{\mathrm{syn}})^\top] +\E[\zeta_te_te_t^\top\zeta_t^\top].
\end{align*}
Boundedness of the sampled matrices, Lemma~\ref{syn:moment-lemma}, and \eqref{syn:mean-bound} imply
\begin{equation}\label{syn:noise-covariance}
 \|J_t-S^{\mathrm{syn}}\|_F
 \le C\{\|m_t\|_2+\E\|e_t\|_2^2\}\le C\alpha_t.
\end{equation}
In particular, the cross terms involve the mean $m_t$, rather than $\E\|e_t\|_2$.
Unrolling \eqref{syn:martingale-recursion} and using martingale orthogonality gives
\begin{equation}\label{syn:true-covariance}
 F_n=\frac1{\sqrt n}\sum_{t<n}W^{\mathrm{syn}}_{n,t}\eta_t,\qquad
 \Gamma_n=\frac1n\sum_{t<n}W^{\mathrm{syn}}_{n,t}J_t
                             (W^{\mathrm{syn}}_{n,t})^\top.
\end{equation}
Uniform boundedness of the weights now yields
\begin{equation}\label{syn:covariance-correction}
 \|\Gamma_n-C_n^{\mathrm{syn}}\|_F
 \le\frac Cn\sum_{t<n}\alpha_t\le Cn^{-\beta}.
\end{equation}

\paragraph{Sensitivity to the independent sample blocks.}
Regard $F_n=f_n(\omega_0,\ldots,\omega_{n-1})$.
Use $\mathcal D_i$ for the difference caused by replacing input block $i$ by another admissible block; the bounds below are uniform over the original blocks and all replacements.
For distinct $i,j$, $\mathcal D_i\mathcal D_j$ is the corresponding four-term difference.
We claim that
\begin{equation}\label{syn:sensitivity}
 \|\mathcal D_i f_n\|_2\le Cn^{-1/2},\qquad
 \|\mathcal D_i\mathcal D_j f_n\|_2
 \le b_{ij}:=Cn^{-1/2}\alpha_{\min(i,j)}
                 \rho_{\min(i,j)+1:\max(i,j)-1}.
\end{equation}
A replacement at time $i$ changes $L_{i+1}$ by at most $C\alpha_i$, since $L_i$ stays in the cube.
At a later time $k$, its effect is at most $C\alpha_i\rho_{i+1:k-1}$.
Summing over the average and applying Lemma~\ref{syn:weight-lemma} proves the first bound.
For $i<j$, the four-term difference first appears at time $j+1$.
At that time it is
$\alpha_j\gamma(\widehat P_j^{\pi^\star}-\widehat P_j^{\pi^\star\prime})$
times the change in $L_j$ caused by block $i$.
Its norm is at most $C\alpha_i\alpha_j\rho_{i+1:j-1}$.
Subsequent common updates are linear and satisfy \eqref{syn:sampled-contraction}.
The response bound starting at $j$ proves the second assertion.

The row sums of these deterministic sensitivity bounds satisfy
\begin{equation}\label{syn:sensitivity-sums}
 \max_i\sum_{j\ne i}b_{ij}\le Cn^{-1/2}.
\end{equation}
For $j>i$, this follows from the response bound in Lemma~\ref{syn:weight-lemma}.
For $j<i$, use its telescoping identity
$\sum_{j<i}\alpha_j\rho_{j+1:i-1}\le a_{\mathrm{syn}}^{-1}$.

\paragraph{Gaussian approximation by resampling.}
Let $\omega'$ be an independent copy of $\omega$.
For $\mathcal J\subseteq\mathcal I_n:=\{0,\ldots,n-1\}$, let $\omega^{\mathcal J}$ replace the blocks in $\mathcal J$ by their copies.
For $j\notin\mathcal J$, write
$\mathcal D_j f_n(\omega^{\mathcal J})
=f_n(\omega^{\mathcal J})-f_n(\omega^{\mathcal J\cup\{j\}})$,
and put
\begin{equation}\label{syn:resampling-matrix}
 \mathcal K_n=\frac12\sum_{\mathcal J\subsetneq\mathcal I_n}
 \vartheta_{\mathcal J}\sum_{j\notin\mathcal J}
 \mathcal D_jf_n(\omega^{\mathcal J})
                       (\mathcal D_jf_n(\omega))^\top,
 \qquad
 \vartheta_{\mathcal J}=\frac1{\binom n{|\mathcal J|}(n-|\mathcal J|)}.
\end{equation}
Lemma~\ref{syn:resampling-lemma}, applied coordinatewise, gives
$\E\mathcal K_n=\Gamma_n$ and
$\sum_{\mathcal J:j\notin\mathcal J}\vartheta_{\mathcal J}=1$ for every $j$.
Replacing one of the $2n$ independent blocks in $(\omega,\omega')$ changes $\mathcal K_n$ in Frobenius norm by at most
\[
 C\left\{n^{-1}+n^{-1/2}\sum_{j\ne i}b_{ij}\right\}\le Cn^{-1}.
\]
Here the terms $j=i$ use the first bound in \eqref{syn:sensitivity}; all other terms use the four-term bound there.
The variance inequality in Lemma~\ref{syn:resampling-lemma} therefore implies
\begin{equation}\label{syn:resampling-variance}
 \E\|\mathcal K_n-\Gamma_n\|_F^2\le C/n,\qquad
 \E\|\mathcal K_n-C_n^{\mathrm{syn}}\|_F
 \le C(n^{-1/2}+n^{-\beta}),
\end{equation}
where the second estimate also uses \eqref{syn:covariance-correction}.

For large $n$, the eigenvalues of $C_n^{\mathrm{syn}}$ are bounded above and below by fixed positive constants, by \eqref{syn:covariance-bound}.
Fix a Euclidean $1$-Lipschitz function $h$ and $0<\delta\le1/2$.
Lemma~\ref{syn:stein-lemma}, with $V=C_n^{\mathrm{syn}}$, supplies a smoothed test $h_\delta$ and a solution $\psi_\delta$ to
\[
 \tr(C_n^{\mathrm{syn}}D^2\psi_\delta(x))-x^\top\nabla\psi_\delta(x)
 =h_\delta(x)-\E h_\delta(Z_n^{\mathrm{syn}}),
\]
with
\[
 \|h_\delta-h\|_{\sup}\le C\delta,\qquad
 \|D^2\psi_\delta\|_{\sup}\le C,\qquad
 \|D^3\psi_\delta\|_{\sup}\le C(1+|\log\delta|).
\]
Apply the covariance identity in Lemma~\ref{syn:resampling-lemma} with
$g=\partial_a\psi_\delta\circ f_n$ and $f=f_{n,a}$, and sum over $a$.
Taylor expansion at $F_n=f_n(\omega)$ gives
\begin{equation}\label{syn:stein-resampling}
 \E[F_n^\top\nabla\psi_\delta(F_n)]
 =\E\langle\mathcal K_n,D^2\psi_\delta(F_n)\rangle_F+\mathcal E_n,
 \qquad
 |\mathcal E_n|\le C(1+|\log\delta|)n^{-1/2}.
\end{equation}
Note that each Taylor remainder is bounded by
$C\|D^3\psi_\delta\|_{\sup}
\|\mathcal D_j f_n(\omega)\|_2^2
\|\mathcal D_j f_n(\omega^{\mathcal J})\|_2$.
The first-difference bound makes this $C(1+|\log\delta|)n^{-3/2}$, and the subset weights sum to one for each of the $n$ blocks.
Combining the Stein equation, \eqref{syn:resampling-variance}, and \eqref{syn:stein-resampling}, we obtain
\[
 |\E h(F_n)-\E h(Z_n^{\mathrm{syn}})|
 \le C\{\delta+n^{-\beta}+n^{-1/2}(1+|\log\delta|)\}.
\]
Taking $\delta=n^{-1/2}$ for $n\ge4$, then the supremum over $h$, and restoring the mean using \eqref{syn:mean-bound}, proves \eqref{syn:finite-bound} for all sufficiently large $n$.
Cube invariance and boundedness of $C_n^{\mathrm{syn}}$ cover the finitely many remaining horizons after increasing $C$.
\end{proof}

\subsection{Proof of Theorem~\ref{syn:lower}}\label{syn:lower-proof}

\synlowerthm*

\begin{proof}
There is one action, so $Q_t=L_t$ and $X_n=X_n^{\mathrm{lin}}$.
On the MDP in \eqref{lb:example}, direct calculation gives
\begin{equation}\label{syn:example-matrices}
 Q^\star=\begin{pmatrix}4/3\\3\end{pmatrix},\qquad
 A^{\mathrm{syn}}=\frac1{25}\begin{pmatrix}9&-4\\-6&11\end{pmatrix},\qquad
 S^{\mathrm{syn}}=\diag(64/225,28/75)\succ0.
\end{equation}
Thus Assumption~\ref{syn:assumption} holds.

First suppose $1/2<\beta<1$.
We apply the deterministic covariance expansion in the proof of Proposition~\ref{lb:prop-cov} after symmetrizing the drift.
Let
\begin{equation}\label{syn:symmetrization}
 \begin{aligned}
 T&=\diag(\sqrt{3/5},\sqrt{2/5}),\qquad
 \widetilde S=TS^{\mathrm{syn}}T^\top=\diag(64/375,56/375),\\
 \widetilde A&=TA^{\mathrm{syn}}T^{-1}
 =\frac1{25}\begin{pmatrix}9&-2\sqrt6\\-2\sqrt6&11\end{pmatrix}.
 \end{aligned}
\end{equation}
The matrix $\widetilde A$ is symmetric positive definite with eigenvalues $1/5$ and $3/5$.
The stepsize assumptions give
\[
 \alpha_0\lambda_{\max}(\widetilde A)\le1,\qquad
 \frac\beta{c_0}b^{\beta-1}
 \le\frac1{20}=\frac{\lambda_{\min}(\widetilde A)}4.
\]
These are the stability and stepsize comparisons used in Section~\ref{lb:cov-proof}.
That calculation uses only a fixed symmetric positive-definite drift, a positive-definite noise covariance, and these comparisons; it does not use the particular entries of the asynchronous matrices.
The weights are equivariant under similarity, so the weights formed with $\widetilde A$ are $TW_{n,t}^{\mathrm{syn}}T^{-1}$, and their covariance is $TC_n^{\mathrm{syn}}T^\top$.
Applying the expansion of Section~\ref{lb:cov-proof} to $(\widetilde A,\widetilde S)$ and transforming back yields
\begin{equation}\label{syn:covariance-expansion}
 C_n^{\mathrm{syn}}=\Sigma_\infty^{\mathrm{syn}}
       +n^{\beta-1}G^{\mathrm{syn}}+o(n^{\beta-1}),\qquad
 G^{\mathrm{syn}}=\frac1{c_0}T^{-1}\widetilde A^{-1}
          \widetilde{\mathcal X}\widetilde A^{-\top}T^{-\top}\succ0,
\end{equation}
where $\widetilde{\mathcal X}$ is the unique solution of
$\widetilde A\widetilde{\mathcal X}
+\widetilde{\mathcal X}\widetilde A^\top=\widetilde S$.
Its positive definiteness follows from Lemma~\ref{lb:lem-lyapunov}.
In particular, $(G^{\mathrm{syn}})_{11}>0$.
The one-dimensional Gaussian comparison in Lemma~\ref{lb:lem-gaussian}, applied after projection onto the first coordinate, implies
\[
 W_1(Z_n^{\mathrm{syn}},Z_\infty^{\mathrm{syn}})\ge c n^{\beta-1}
\]
for all sufficiently large $n$.
The reverse triangle inequality and \eqref{syn:finite-bound} now give
\begin{align*}
 W_1(X_n,Z_\infty^{\mathrm{syn}})
 &\ge W_1(Z_n^{\mathrm{syn}},Z_\infty^{\mathrm{syn}})
       -W_1(X_n^{\mathrm{lin}},Z_n^{\mathrm{syn}})\\
 &\ge c n^{\beta-1}-C\{n^{-1/2}\log(en)+n^{-\beta}\}
 \ge c' n^{\beta-1},
\end{align*}
since the subtracted terms are $o(n^{\beta-1})$ for every fixed $\beta>1/2$.

At $\beta=1/2$, use the mean instead.
By \eqref{syn:martingale-recursion} and $Q_0=0$,
\[
 \E X_n=-\frac1{\sqrt n}\sum_{k=1}^n
                       \Phi^{\mathrm{syn}}_{0:k-1}Q^\star
       =-\frac v{\sqrt n}+o(n^{-1/2}),\qquad
 v=\sum_{k=1}^\infty\Phi^{\mathrm{syn}}_{0:k-1}Q^\star.
\]
The series converges by product contraction.
Every factor $I-\alpha_tA^{\mathrm{syn}}$ is nonnegative with strictly positive row sums, and every coordinate of $Q^\star$ is positive.
Consequently $v$ has strictly positive coordinates and is nonzero.
The difference-of-means bound in Lemma~\ref{lb:lem-gaussian} gives
\[
 W_1(X_n,Z_\infty^{\mathrm{syn}})\ge\|\E X_n\|_2\ge c n^{-1/2}
\]
for all sufficiently large $n$.
The MDP and $Q_0$ are the same for every admissible choice of stepsize parameters.
\end{proof}

\subsection{Supporting Lemmas}\label{syn:support}

\begin{lemma}[Synchronous deterministic weights]\label{syn:weight-lemma}
The weights in \eqref{syn:weights} satisfy
\begin{align}
 \sup_{n,t<n}\|W_{n,t}^{\mathrm{syn}}\|_\infty&\le C,
 &\|C_n^{\mathrm{syn}}-\Sigma_\infty^{\mathrm{syn}}\|_F&\le Cn^{\beta-1},
 \label{syn:deterministic-bounds}\\
 \sup_{i\ge0}\alpha_i\sum_{k>i}\rho_{i+1:k-1}&\le C,
 &\sum_{j<i}\alpha_j\rho_{j+1:i-1}
 &=\frac{1-\rho_{0:i-1}}{a_{\mathrm{syn}}}
 \le a_{\mathrm{syn}}^{-1}.
 \label{syn:response-bounds}
\end{align}
Moreover,
\begin{equation}\label{syn:deterministic-products}
 \|\Phi^{\mathrm{syn}}_{i:j}\|_\infty
 \le\rho_{i:j}\le\exp\!\left(-a_{\mathrm{syn}}\sum_{u=i}^j\alpha_u\right),
 \qquad \|(A^{\mathrm{syn}})^{-1}\|_\infty\le a_{\mathrm{syn}}^{-1}.
\end{equation}
\end{lemma}

\begin{proof}
The mean update matrices are nonnegative with row sums $1-a_{\mathrm{syn}}\alpha_t$, giving the product bound.
The inverse bound follows from the Neumann series for $I-\gamma P^{\pi^\star}$.
The proof of Lemma~\ref{lem:weights}, with $D=I$ and $a_\star=a_{\mathrm{syn}}$, gives
\begin{equation}\label{syn:weight-pointwise}
 \|W_{n,t}^{\mathrm{syn}}-(A^{\mathrm{syn}})^{-1}\|_\infty
 \le C\left\{(t+b)^{\beta-1}
 +\exp\!\left(-\frac{a_{\mathrm{syn}}}2\sum_{u=t+1}^n\alpha_u\right)\right\},
\end{equation}
as well as uniform boundedness of the weights.
Those arguments require $0<\beta<1$ and the shift comparison, and do not use $\sum_t\alpha_t^2<\infty$; hence they also cover $\beta=1/2$.
For completeness, averaging the first term in \eqref{syn:weight-pointwise} gives $O(n^{\beta-1})$ by the integral test, and the exponential term satisfies
\[
 \sum_{t<n}\exp\!\left(-\frac{a_{\mathrm{syn}}}2
                                      \sum_{u=t+1}^n\alpha_u\right)
 \le\sum_{k\ge1}e^{-a_{\mathrm{syn}}k\alpha_n/2}
 \le C\alpha_n^{-1}\le Cn^\beta.
\]
Since the weights and $(A^{\mathrm{syn}})^{-1}$ are bounded, expanding the two covariance factors and averaging proves the covariance bound in \eqref{syn:deterministic-bounds}.

To prove the response bound directly, set
$s_{i,k}=\sum_{u=i+1}^{k-1}\alpha_u$.
The stepsize comparison used in Lemma~\ref{lem:weights} gives
\[
 \alpha_i/\alpha_k
 \le C\exp\{a_{\mathrm{syn}}s_{i,k}/4\},\qquad k>i.
\]
Thus
\[
 \alpha_i\sum_{k>i}\rho_{i+1:k-1}
 \le C\sum_{k>i}\alpha_k e^{-3a_{\mathrm{syn}}s_{i,k}/4}\le C.
\]
The last sum is bounded by an exponential integral, since the intervals
$[s_{i,k},s_{i,k+1}]$ partition $[0,\infty)$ and have lengths $\alpha_k\le1$.
Finally,
$a_{\mathrm{syn}}\alpha_j\rho_{j+1:i-1}
=\rho_{j+1:i-1}-\rho_{j:i-1}$.
Summing over $j<i$ proves the second identity in \eqref{syn:response-bounds}.
\end{proof}

\begin{lemma}[Synchronous iterate moments]\label{syn:moment-lemma}
Both recursions \eqref{syn:algorithm} and \eqref{syn:linear-recursion} satisfy
\begin{equation}\label{syn:moment-bounds}
 \E\|Q_t-Q^\star\|_\infty^p
 +\E\|L_t-Q^\star\|_\infty^p\le C\alpha_t^{p/2},
 \qquad p\in\{2,4,6\},\quad t\ge0.
\end{equation}
The action-gap and covariance assumptions are not needed for this estimate.
\end{lemma}

\begin{proof}
Choose a fixed $q\ge2$ sufficiently large that $\theta=\gamma d^{1/q}<1$.
Both mean Bellman maps, optimal and fixed-policy, are $\theta$-contractions in $\ell_q$, since
\[
 \|\mathcal T x-\mathcal T y\|_q
 \le d^{1/q}\gamma\|x-y\|_\infty
 \le\theta\|x-y\|_q,
\]
and the same bound holds for $r+\gamma P^{\pi^\star}x$.
For either error sequence $e_t$, write its update as
\[
 e_{t+1}=\bar e_t+\alpha_t\xi_t,\qquad
 \|\bar e_t\|_q\le[1-(1-\theta)\alpha_t]\|e_t\|_q,
 \qquad \E[\xi_t\mid\cF_{t-1}^{\mathrm{syn}}]=0.
\]
The noise is uniformly bounded by cube invariance, and $\bar e_t$ is measurable with respect to the past.
For $p\in\{2,4,6\}$, the first-order Taylor remainder for $x\mapsto\|x\|_q^p$ is at most
$C(\|x\|_q^{p-2}\|y\|_q^2+\|y\|_q^p)$.
This follows by integrating its second directional derivative along $x+uy$; its absolute value is bounded by $C\|x+uy\|_q^{p-2}\|y\|_q^2$.
Taking conditional expectations removes the linear term.
Consequently, with $M_{p,t}=\E\|e_t\|_q^p$ and $M_{0,t}=1$,
\[
 M_{p,t+1}\le(1-c\alpha_t)M_{p,t}
                  +C\alpha_t^2M_{p-2,t}+C\alpha_t^p.
\]
Inductively in $p$, the forcing terms are $O(\alpha_t^{p/2+1})$.
Also,
$\alpha_{t+1}^{p/2}/\alpha_t^{p/2}=1-O((t+b)^{-1})$
and $(t+b)^{-1}=o(\alpha_t)$ because $\beta<1$.
Comparing with a sufficiently large multiple of $\alpha_t^{p/2}$ proves
$M_{p,t}\le C\alpha_t^{p/2}$ from some fixed time onward.
Increasing $C$ covers the finite prefix and both recursions.
The maximum norm is bounded by the $\ell_q$ norm, proving \eqref{syn:moment-bounds}.
The argument applies without change at $\beta=1/2$.
\end{proof}

\begin{lemma}[Resampling covariance identity]\label{syn:resampling-lemma}
Let $\omega=(\omega_0,\ldots,\omega_{n-1})$ have independent coordinates and let $\omega'$ be an independent copy.
Use the resampling notation and weights of \eqref{syn:resampling-matrix}.
For square-integrable scalar functions $g,f$,
\begin{equation}\label{syn:resampling-identity}
 \Cov(g(\omega),f(\omega))
 =\frac12\sum_{\mathcal J\subsetneq\mathcal I_n}\vartheta_{\mathcal J}
    \sum_{j\notin\mathcal J}
    \E[\mathcal D_jg(\omega)\mathcal D_jf(\omega^{\mathcal J})].
\end{equation}
For each $j$, $\sum_{\mathcal J:j\notin\mathcal J}\vartheta_{\mathcal J}=1$.
The identity is Lemma~2.3 in \cite{chatterjee2008new}.
We also use the variance inequality: if $G$ is a square-integrable matrix-valued function of independent inputs $U_1,\ldots,U_m$, and $G^{[i]}$ replaces input $i$ by an independent copy, then
\begin{equation}\label{syn:variance-tensorization}
 \E\|G-\E G\|_F^2\le\frac12\sum_{i=1}^m\E\|G-G^{[i]}\|_F^2.
\end{equation}
\end{lemma}

\begin{proof}
Average the telescoping replacement of all coordinates over a uniform random permutation.
The probability that the coordinates before $j$ form exactly $\mathcal J$ is
$|\mathcal J|!(n-|\mathcal J|-1)!/n!=\vartheta_{\mathcal J}$.
Thus
\[
 f(\omega)-f(\omega')
 =\sum_{\mathcal J\subsetneq\mathcal I_n}\vartheta_{\mathcal J}
                       \sum_{j\notin\mathcal J}\mathcal D_jf(\omega^{\mathcal J}).
\]
Multiply by $g(\omega)$ and take expectations.
In each summand, interchanging $\omega_j$ and $\omega'_j$ preserves the joint law and reverses the sign of $\mathcal D_jf(\omega^{\mathcal J})$.
Averaging before and after this interchange gives \eqref{syn:resampling-identity}.
The subset-weight identity follows because exactly one predecessor set occurs for each $j$ in every permutation.
Finally, the Doob martingale for $G$ in the order $U_1,\ldots,U_m$ has orthogonal differences.
For each entry, the squared norm of its $i$th difference is at most
$\E\Var(G\mid U_1,\ldots,U_{i-1},U_{i+1},\ldots,U_m)$ by conditional Jensen.
This conditional variance equals half the expected squared resampling difference.
Summing over $i$ and matrix entries proves \eqref{syn:variance-tensorization}.
\end{proof}

\begin{lemma}[Gaussian smoothing and Stein equation]\label{syn:stein-lemma}
Let $V\succ0$ have eigenvalues in a fixed interval $[\lambda,\Lambda]\subset(0,\infty)$, and let $Z_V\sim\N(0,V)$.
For every Euclidean $1$-Lipschitz function $h$ and $0<\delta\le1/2$, there are smooth functions $h_\delta,\psi_\delta$ such that
\begin{align}
 \tr(VD^2\psi_\delta(x))-x^\top\nabla\psi_\delta(x)
 &=h_\delta(x)-\E h_\delta(Z_V),
 \label{syn:stein-equation}\\
 \|h_\delta-h\|_{\sup}\le C\delta,\qquad
 \|D^2\psi_\delta\|_{\sup}&\le C,\qquad
 \|D^3\psi_\delta\|_{\sup}\le C(1+|\log\delta|).
 \label{syn:stein-bounds}
\end{align}
The derivative norms are Euclidean operator norms, and $C$ depends only on $d,\lambda,\Lambda$.
\end{lemma}

\begin{proof}
Set $h_\delta(x)=\E h(x+\delta Z_V)$.
The smoothing error is at most $\delta\E\|Z_V\|_2\le\delta\sqrt{d\Lambda}$.
For the Gaussian semigroup
\[
 (P_sh_\delta)(x)=\E h_\delta(e^{-s}x+\sqrt{1-e^{-2s}}Z_V),
\]
define
\[
 \psi_\delta(x)=-\int_0^\infty
                 \{P_sh_\delta(x)-\E h_\delta(Z_V)\}\,ds.
\]
The integral converges by the Lipschitz property and Gaussian coupling.
Its generator is $\tr(VD^2)-x^\top\nabla$, and integration of
$\partial_sP_sh_\delta$ proves \eqref{syn:stein-equation}.
To bound derivatives, combine the two Gaussian convolutions:
\[
 P_sh_\delta(x)=\E h(e^{-s}x+\sqrt{1-e^{-2s}+\delta^2}Z_V).
\]
Use the bounded first derivative of a Lipschitz function and integrate the Gaussian density by parts once and twice, respectively.
This gives
\[
 \|D^2P_sh_\delta\|_{\sup}
 \le\frac{Ce^{-2s}}{\sqrt{1-e^{-2s}+\delta^2}},\qquad
 \|D^3P_sh_\delta\|_{\sup}
 \le\frac{Ce^{-3s}}{1-e^{-2s}+\delta^2}.
\]
The first bound has a uniformly finite integral over $s\ge0$.
The integral of the second is at most $C(1+|\log\delta|)$, since its denominator is comparable to $s+\delta^2$ on $0\le s\le1$ and its numerator decays exponentially for $s\ge1$.
Integrating these bounds proves \eqref{syn:stein-bounds}.
For nonsmooth $h$, the integrations by parts use its bounded weak gradient; equivalently, first smooth $h$ further and pass to the limit.
\end{proof}

\section{Details of Experiments}

\paragraph{The setup of the MDP and Simulation.}
We consider a fixed discounted MDP with state space
$\mathcal S=\{0,1,2,3\}$, action space $\mathcal A=\{0,1,2\}$,
and discount factor $\gamma=0.8$.
Let $\nu=(0.1,0.2,0.3,0.4)^\top$.
The transition matrix associated with action $a$ is
\[
P_a=p_a I+(1-p_a)\mathbf 1\nu^\top,
\qquad
(p_0,p_1,p_2)=(0.8,0.6,0.4),
\]
and the deterministic reward matrix, with rows indexed by states
and columns by actions, is
\[
r=
\begin{pmatrix}
0&0.4&0.1\\
1&0&0.2\\
0.1&0.2&1\\
0.3&0.9&0
\end{pmatrix}.
\]
The behavior policy selects each action with probability $1/3$
at every state. Its state transition matrix is
$P_b=0.6I+0.4\mathbf 1\nu^\top$, whose stationary distribution is
$\nu$. The optimal policy is $\pi^\star=(1,0,2,1)$. We use the stepsize
$
\alpha_t=c_0(t+b)^{-\beta}, \beta=\frac34, b=512, c_0=512^{3/4}.
$
These choices give $\alpha_0=1$ and satisfy
$
\frac{\beta}{c_0}b^{\beta-1} = \frac{\beta}{b} \le \frac{\rho(1-\gamma)}{4}.
$
We simulate $R=32{,}768$ independent trajectories through
$n_{\max}=10^8$.

\paragraph{Measuring Gaussian approximation.}
We represent Q-functions as twelve-dimensional vectors. Recall that $D=\operatorname{diag}(\mu)$,
$
A=D(I-\gamma P^{\pi^\star}),
P^{\pi^\star}_{(s,a),(s',a')} = P(s'\mid s,a)\mathbf 1\{a'=\pi^\star(s')\},
$
$
S_{(s,a),(s,a)}
=
\mu(s,a)\gamma^2
\operatorname{Var}_{S'\sim P(\cdot\mid s,a)}
\!\left[V^\star(S')\right], V^\star(s)=\max_{a\in\mathcal A}Q^\star(s,a),
$
and
$\Sigma_\infty=A^{-1}SA^{-\top}$.
To measure Gaussian approximation, we consider a fixed
one-dimensional projection $u=v/\|v\|_2$, where $v$ is the
first-order stationary mean-bias coefficient of the locally
linearized Markov recursion. We obtain $v$ by solving a finite system of 48 scalar equations
for the joint first moments of the state and the twelve-dimensional
error. The projection is therefore determined entirely by the
MDP and is fixed before simulation.
At each horizon $n$, we estimate the projected Wasserstein
distance by
\[
\hat{d}_n
=
W_1\!\left(
\frac1R\sum_{j=1}^{R}\delta_{u^\top X_n^{(j)}},
\mathcal N(0,u^\top\Sigma_\infty u)
\right),
\]
where the target Gaussian has standard deviation approximately
$5.22327$. 
Figure~\ref{fig:example}(a) compares $\hat{d}_n$ with the benchmark
$n^{-1/4}B$, where
\[
B
=
\frac{c_0}{1-\beta}|u^\top v|
\approx 73.13670.
\]
This coefficient is computed from the MDP and stepsize
parameters before simulation.

\paragraph{Covariance comparison.}
To examine the covariance error in our Gaussian approximation
analysis, we consider the linear noise recursion
\[
\ell_0=0,
\qquad
\ell_{t+1}
=
(I-\alpha_tA)\ell_t+\alpha_t\epsilon_t,
\qquad
M_n=\frac1{\sqrt n}\sum_{k=1}^{n}\ell_k,
\]
where $\epsilon_t$ is the Bellman innovation at $Q^\star$
along the stationary behavior trajectory. These innovations
form a martingale difference sequence with unconditional
covariance $S$. We compute \(C_n=\operatorname{Cov}(M_n)\) using a deterministic finite-step covariance recursion, without Monte Carlo approximation. This allows us to directly examine the covariance error \(\|C_n-\Sigma_\infty\|_{\mathrm F}\) appearing in the Gaussian approximation decomposition in Section \ref{sec_ub}. To obtain the first-order covariance benchmark, we solve the Lyapunov equation $AH+HA^\top=S$
and define
\[
G=c_0^{-1}A^{-1}HA^{-\top},
\qquad
\|G\|_{\mathrm F}\approx24.44716.
\]
Figure~\ref{fig:example}(b) compares the covariance error
$\|C_n-\Sigma_\infty\|_{\mathrm F}$ with the reference curve
$n^{-1/4}\|G\|_{\mathrm F}$, illustrating the covariance
error due to finite-sample approximation.

\end{document}